\documentclass{report}

\newcommand{\mychapter}[2]{
    \setcounter{chapter}{#1}
    \setcounter{section}{0}
    \chapter*{#2}
    \addcontentsline{toc}{chapter}{#2}
}

\usepackage{Lifu}
\usepackage[english]{babel}
\usepackage[utf8x]{inputenc}

\usepackage{booktabs}
\usepackage{tabu}
\usepackage[T1]{fontenc}

\usepackage[a4paper,top=3cm,bottom=2cm,left=3cm,right=3cm,marginparwidth=2cm]{geometry}

\usepackage{booktabs}
\usepackage{adjustbox}
\usepackage{supertabular}

\usepackage[ruled,vlined]{algorithm2e}

\usepackage{makecell}
\usepackage{subcaption}
\usepackage{amsfonts}
\usepackage{amsmath}
\usepackage{graphicx}
\usepackage{times}
\usepackage{natbib}
\usepackage[colorinlistoftodos]{todonotes}
\usepackage[colorlinks=true, allcolors=blue]{hyperref}

\DeclareMathOperator*{\argmax}{argmax}
\DeclareMathOperator*{\argmin}{argmin}
\DeclareMathOperator*{\LM}{TLM}
\usepackage{multirow}

\DeclareMathOperator*{\BLSTM}{BiLSTM}

\DeclareMathOperator*{\softmax}{softmax}
\newcommand{\Expect}{{\rm I\kern-.3em E}}
\newcommand{\x}{\boldsymbol{x}}
\newcommand{\y}{\boldsymbol{y}}
\newcommand{\z}{\boldsymbol{z}}
\newcommand{\q}{\boldsymbol{q}}
\newcommand{\yv}{\mathbf{y}}
\newcommand{\Xspace}{\mathcal{X}}
\newcommand{\yspace}{\mathcal{Y}}
\newcommand{\relyspace}{\mathcal{Y}_{R}}
\newcommand{\cost}{\bigtriangleup}
\newcommand{\relyspacer}{\mathcal{Y}_{R'}}
\DeclareMathOperator{\E}{\mathbb{E}}

\newcommand{\vv}{\mathbf{v}}

\newcommand{\onehot}{\textit{onehot}}
\newcommand{\score}{\textit{score}}
\newcommand{\NN}{\textit{NN}}
\newcommand{\name}{ENGINE\xspace}

\newcommand{\infnetnoparams}{\mathbf{A}}

\newcommand{\infnet}{\mathbf{A}_{\Psi}}
\newcommand{\canet}{\mathbf{F}_{\Phi}}
\newcommand{\mlocal}{\mathit{loc}}
\newcommand{\mlabel}{\mathit{lab}}

\newcommand{\ltok}{\ell_{\mathrm{token}}}

\newcommand{\loss}{\ell_{\mathrm{loss}}}

\usepackage{xspace}
\newcommand{\blstmcrf}{BLSTM-CRF\xspace}
\newcommand{\blstmcnncrf}{BLSTM-CRF+\xspace}

\newcommand{\info}{\mathit{inf}}
\newcommand{\vito}{\mathit{vit}}

\newcommand{\infnetBase}{\textrm{infnet}\xspace}
\newcommand{\infnetLarge}{\textrm{infnet+}\xspace}

\DeclareMathOperator*{\ce}{CE}

\newcommand{\given}{\mid}
\newcommand{\oone}{\mathbf{O_1}}
\newcommand{\otwo}{\mathbf{O_2}}
\newcommand{\ogen}{\mathbf{O}}

\newcommand{\pvect}{\mathbf{v}}

\DeclareMathOperator*{\attention}{attention}

\DeclareMathOperator*{\MLP}{MLP}

\DeclareMathOperator*{\product}{VKP}

\begin{document}

\title{Learning Energy-Based Approximate Inference Networks\\
            for  Structured Applications in NLP}
\author{Lifu Tu}
\principaladviser{Kevin Gimpel}
\firstreader{Karen Livescu}
\secondreader{Sam Wiseman}
\thirdreader{Kyunghyun Cho} 
\fourthreader{Marc’Aurelio Ranzato} 
 
\beforepreface
\prefacesection{Abstract}
 Structured prediction in natural language processing (NLP) has a long history. The complex models of structured application come at the difficulty of learning and inference. These difficulties lead researchers to focus more on models with simple structure components (e.g., local classifier). Deep representation learning has become increasingly popular in recent years. The structure components of their method, on the other hand, are usually relatively simple. We concentrate on complex structured models in this dissertation. We provide a learning framework for complicated structured models as well as an inference method with a better speed/accuracy/search error trade-off.

 The dissertation begins with a general introduction to energy-based models. In NLP and other applications, an energy function is comparable to the concept of a scoring function. In this dissertation, we discuss the concept of the energy function and structured models with different energy functions. Then, we propose a method in which we train a neural network to do argmax inference under a structured energy function, referring to the trained networks as "inference networks" or "energy-based inference networks". 
 We then develop ways of jointly learning energy functions and inference networks using an adversarial learning framework. 
 Despite the inference and learning difficulties of energy-based models, we present approaches in this thesis that enable energy-based models more easily to be applied in structured NLP applications.

\prefacesection{Acknowledgments}
First and foremost, I'd like to express my gratitude to Kevin Gimpel, my advisor. Before working with him, I didn't know much about NLP. I've learnt a lot from him, not only in terms of knowledge, but also in terms of how to dig deeply into research problems. It's good to be able to formalize a research problem and enjoy research process, even if there's a little pressure or no obvious path sometimes. Kevin give me a lot of freedom for my research directions and provide lots of help kindly. I'm honored to have been his first graduated Ph.D. student at Toyota Technological Institute in Chicago (TTIC). Thank you for being a mentor to me.

Next, I'd like to express my gratitude to Kyunghyun Cho, Karen Livescu, Marc'Aurelio Ranzato, and Sam Wiseman, other members of my committee. It's fantastic to have so many accomplished researchers on my committee. Even though their schedules are really, they are still glad to provide help. Some of them, I did not know them before. I really appreciate their time and great help. Thanks for your input, which has helped me improve the presentation of my research work, and rethink my research. I've learned a lot, especially when it comes to relating my study to earlier work.

During my Ph.D. path, I benefited immensely from my internship experience. I'd want to express my gratitude to Dong Yu for hosting me at the Tencent AI lab. It's great to be able to take in the sights of Seattle while working on my research internship project. In my second internship, I was lucky to do some research work at AWS AI. Interaction with He He, Spandana Gella, Garima Lalwani, Alex Smola, and others in the AWS AI Lex and Comprehend groups has been beneficial. The internships listed above allowed me to learn more about industry. 

I'd want to thank everyone at TTIC for making my Ph.D. path so enjoyable. Jinbo Xu, my temporary advisor, who is assisting me with my application for the TTIC Ph.D. program. I'm grateful to David McAllester for assisting me in seeing my work from numerous angles. I think the fellow students, including Heejin Choi, Mingda Chen, Zewei Chu, Falcon Dai, Xiaoan Ding, Lingyu Gao, Ruotian Luo, Jianzhu Ma, Mohammadreza Mostajabi, Takeshi Onishi, Freda Shi, Siqi Sun, Hao Tang, Qingming Tang, Shubham Toshniwal, Hai Wang, Zhiyong Wang, John wieting, Davis Yoshida. Thanks to Aynaz Taheri and Xiang Li. We work on my first NLP project together. I thank visiting students, Jon Cai, Yuanzhe (Richard) Pang, Tianyu Liu, and Manasvi Sagarkar, for our wonderful collaborations. I would also like to thank friends in Chicago.

Finally, thanks to my family for their support over the years, unconditionally! Thank you for everything!  
\afterpreface


\tableofcontents
\newpage

\mychapter{1}{Introduction}
\section{Structured Prediction in NLP}




\paragraph{Structured Prediction (or Structure Prediction):} In NLP applications, there exists strong complex dependency between the structured outputs. We call them as \textbf{structured applications} here. Structured prediction is a machine learning term that refers to predict the structured output in structured applications. Such applications also appear in computer vision (e.g., image segmentation that interpreting an image of different objects ), computational biology (e.g., protein folding that translates a protein sequence into a three-dimensional structure). In NLP, there are lots of linguistic structure~\citep{smith:2011:synthesis}, for example, phonology, morphology, semantic etc.

Two structured applications in NLP, Part-of-Speech (POS) Tagging in Table~\ref{tab:pos} and machine translation in Table~\ref{tab:mt} are shown below. In both of these two tasks, there are strong dependency between structured output. For example, in POS tagging, the tag ``poss.'' is highly followed by tag ``noun'', and ``adj.'' is highly followed by ``noun''. In machine translation, translations need to have similar meanings with given source language sequence, and keep the syntactic property of target languages.  


\vspace{1cm}

\begin{table}[ht]
\centering
\begin{tabular}{llllllllll}
John                                                                                    & Verret                                                                                  & ,                      & the                         & agency                      & 's                     & president              & and                   & chief                  & executive              \\
\begin{tabular}[c]{@{}l@{}}\textcolor{red}{proper }\\\textcolor{red}{noun}\end{tabular} & \begin{tabular}[c]{@{}l@{}}\textcolor{red}{proper }\\\textcolor{red}{noun}\end{tabular} & \textcolor{red}{comma} & \textcolor{red}{determiner} & \textcolor{red}{noun}       & \textcolor{red}{poss.} & \textcolor{red}{noun}  & \textcolor{red}{cc.}  & \textcolor{red}{adj.}  & \textcolor{red}{noun}  \\
                                                                                        &                                                                                         &                        &                             &                             &                        &                        &                       &                        &                        \\
                                                                                        & ,                                                                                       & will                   & retain                      & the                         & title                  & of                     & president             & .                      &                        \\
                                                                                        & \textcolor{red}{comma}                                                                  & \textcolor{red}{modal} & \textcolor{red}{verb}       & \textcolor{red}{determiner} & \textcolor{red}{noun}  & \textcolor{red}{prep.} & \textcolor{red}{noun} & \textcolor{red}{punc.} &                       
\end{tabular}
\caption{Here we show one example from POS Tagging, which is a sequence labeling task. The above example is from PTB~\citep{ptb}. For a sequence label task, every token (shown with black text) in the sequence has a label (shown with red text in the above example)  . 
The output space is all the possible label sequence with the \textbf{same} length as input sequence. So the size of the space is usually \textbf{exponentially large}.}
\label{tab:pos}
\end{table}

\vspace{1cm} 
\begin{table}[ht]
\centering
\begin{tabular}{llllllllllll}
German:                    &  & aber                 & warten                & sie                & ,                     & dies                & hier                      & ist                 & wirklich                  & meine                    & .  \\
                           &  &                      &                       &                    &                       &                     &                           &                     &                           &                          &    \\
\textcolor{blue}{English:} &  & \textcolor{red}{but} & \textcolor{red}{wait} & \textcolor{red}{,} & \textcolor{red}{this} & \textcolor{red}{is} & \textcolor{red}{actually} & \textcolor{red}{my} & \textcolor{red}{favorite} & \textcolor{red}{project} &  \textcolor{red}{.}  
\end{tabular}
\caption{One translation pair from IWSLT14 German (DE) $\rightarrow$ English (EN) is shown above. Machine translation is a hard task. The output space of a machine translation system is all possible translations given a source language sequence. The output space size is \textbf{infinite}.}
\label{tab:mt}
\end{table}

\begin{figure}[ht]
\centering
\includegraphics[width=1.0\textwidth]{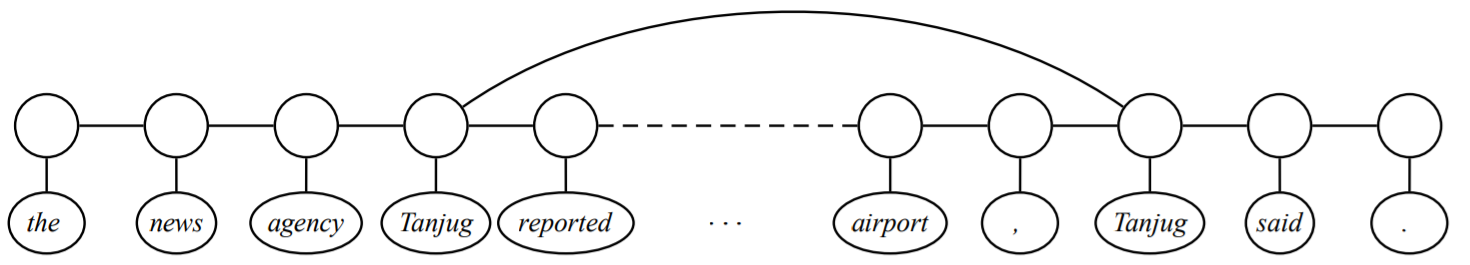}
\caption{An example from CoNLL 2003 Named Entity Recognition~\citep{tjong-kim-sang-de-meulder-2003-introduction}. The second occurrence of the token ``Tanjug'' is unclear whether it is a person or organization. The first occurrence of ``Tanjug'' provides evidence that it is an organization. In order to enforce label consistency for the two occurrences, high-order energies are needed. The example is from ~\citet{finkel-etal-2005-incorporating}.\label{fig:complexDependency}}
\end{figure}

 In natural language processing, many tasks(e.g., sequence labeling, semantic role labeling, parsing, machine translation) involve predicting structured outputs. structured outputs can be a Part-of-Speech (POS) sequence, a parser tree for parsing, an English translation, etc. There are dependencies among the labels. It is crucial to model the dependencies between the structured output. And complex structures can exist in NLP tasks. Figure~\ref{fig:complexDependency} shows one example from CoNLL Named Entity Recognition dataset~\citep{tjong-kim-sang-de-meulder-2003-introduction}, which is one important structured application in NLP. The set of entity type is {none, person, location, organization, location, miscellaneous entity}. Tag ``O'' means the token is outside of entities. If the entity type is one of set {person, location, organization, location, miscellaneous entity}, we add special symbols for the entity. ``B'' stands for ``begin'', ``I'' stands for ``inside''. It is called BIO tagging. We can see there is a long-range dependence between the labels of two occurrences of ``Tanjug''. If there is a strong assumption: we can get perfect representations for the two occurrences, maybe strong output structure can be ignored. However, this is a very strong assumption, especially for noisy inputs in the real world.

Recently, deep representation models ~\citep{peters-etal-2018-deep,gpt,devlin-etal-2019-bert} obtain amazing performance for a wide range of tasks in NLP. However, they usually assume that the structured outputs are independent. During the decoding process, the structured output are generated ignoring previous predicted output, for example local classifiers. The local classifier can be fed into strong deep representation, however, has independent assumption over the structured output given these representations. 

Large models are popular because with these pretrained models~\citep{peters-etal-2018-deep,gpt,devlin-etal-2019-bert,radford2019language}, researchers get strong performance on lots of downstream NLP tasks: GLUE~\citep{wang-etal-2018-glue}, SQuAD~\citep{rajpurkar-etal-2016-squad}, LAMBADA~\citep{paperno-etal-2016-lambada},  SWAG~\citep{zellers-etal-2018-swag}, Children’s Book Test~\citep{cbt}, CoQA~\citep{reddy-etal-2019-coqa}, machine translation, and question answering etc.

\paragraph{Our Focus:} Researchers are increasingly applying deep representation learning to these problems, but the structured component of these approaches is usually quite \textbf{simplistic}\footnote{The size of structured components will be discussed in the next chapter. There is a quick look about at Figure~\ref{fig:parts}, which shows structured models with different part sizes.}. In this thesis, we focus more on how to learn complex structured components for structured tasks, and how to the do inference for complex structured models. 




\section{The Benefits of Energy-Based Modeling for Structured Prediction}

For previous structured models, the dependence of their expressivity on the structured output is limited. Here, we present the concept of "energy-based modeling"~\citep{lecun-06, belanger2016structured} to model complex dependencies between structured outputs.

Give an input sequence $\x$ and a output sequence $\y$ pair, energy-based modeling \citep{lecun-06, belanger2016structured} associates a scalar measure $E(\x,\y)$ of compatibility to each configuration of input $\x$ and output variables $\y$. \citet{belanger2016structured} formulated deep energy-based models for structured prediction, which they called structured prediction energy networks (SPENs). SPENs use arbitrary neural networks to define the scoring function over input/output pairs. Compared with other structured models, they are much more powerful. Energy-based models do not place any limits on the size of the structured parts. 


The potential benefits of Energy-Based modeling is to model complex structured components. For example, sequence labeling tasks usually learn a linear-chain CRFs that only learn the weight between successive labels and neural machine translation systems use unstructured training of local factors. For the energy model, it could capture the arbitrary dependence, especially the long-range dependency. For the generation, energy-based models could be used to generate outputs that favor fewer repetitions, higher BLEU scores, or high semantic similarity with golden outputs with complex energy terms.    



\section{The Difficulties of Energy-Based Models}

The energy captures dependencies between labels with flexible neural networks. However, this flexibility of the deep energy-based models leads to challenges for \textbf{learning} and \textbf{inference}.

For inference, given the input $\x$, we need to find a sequence $\y$ in the output space with lowest energy:
\begin{align}
    \min_{\y}  E_{\Theta}(\x,\y) \nonumber
\end{align}
The output space is exponentially-large output space. This step is hard to jointly predict the label sequence for a task with complex structured components because there are no strong independent assumptions. The process can be intractable for general energy functions. Other inference problems (e.g., cost-augmented inference and marginal inference) also require calculations over an exponentially-large output space.

The original work on SPENs used \textbf{gradient descent} for structured inference~\citep{belanger2016structured,End-to-EndSPEN}. In order to apply gradient descent for training and inference, they relax the output space from \textbf{discrete} to \textbf{continuous}. However, it is hard to guarantee the convergence for gradient descent inference. Furthermore, a lot of iterations could be needed for the convergence. Both of these could slow down the inference step and decrease the performance. 

In our work, we replace this use of gradient descent with a neural network trained to approximate structured inference. The neural network is called "\textbf{energy-based inference network}". It outputs continuous values that we treat as the output structure. 

In summary, the \textbf{contributions of this thesis} are as follows:

\begin{itemize}
    \item Developing a novel inference method called "inference networks" or "energy-based inference network" for structured tasks;
    \item Demonstrating our proposed method achieves a better speed/accuracy/search error trade-off than gradient descent, while also being faster than exact inference at similar accuracy levels;
    \item Applying our method on lots of structured NLP tasks, such as multi-label classification, part-of-speech tagging, named entity recognition, semantic role labeling, and non-autoregressive machine translation. Especially, we achieve state-of-the-art \textbf{purely} non-autoregressive machine translation on the IWSLT 2014 DE-EN and WMT 2016 RO-EN datasets;
    
    \item Developing a new margin-based framework that jointly learns energy functions and inference networks. The proposed framework enables us to explore rich energy functions for sequence labeling tasks. 
    
    \end{itemize}




\section{Overview and Contributions}

The thesis is organized as follows.

\begin{itemize}
    \item In chapter 2, we summarize the history of energy-based models and some connections with previous structured models in natural language processing. Some previous wildly used learning and inference approaches are also discussed.
    
    \item In chapter 3, we replace this use of gradient descent with a neural network trained to approximate structured argmax inference. The "inference network" outputs continuous values that we treat as the output structure. According to our experiments, ``Inference networks'' achieves a better speed/accuracy/search error trade-off than gradient descent, while also being faster than exact inference at similar accuracy levels.
    
    \item In chapter 4, inference networks are used for non-autoregressive machine translation model training with pretrained autoregrssive energies. We achieve state-of-the-art purely non-autoregressive results on the IWSLT 2014 DE-EN and WMT 2016 RO-EN datasets, approaching the performance of autoregressive models.
    
    \item In chapter 5, we design large-margin training objectives to jointly train deep energy functions and inference networks adversarially. As we know, it is the first that adversarial training approach is used in structured prediction. Our training objectives resemble the alternating optimization framework of generative adversarial
networks~\citep{goodfellow2014generative}.

    \item We find that alternating optimization is a little unstable. In chapter 6, we contribute several strategies to stabilize and improve this joint training of energy functions and inference networks for structured prediction. We design a
compound objective to jointly train both cost-augmented and test-time inference networks along with the energy function. It also simpifies our learning pipline.

    \item In chapter 7,  we apply our framework to learn high-order models in structured applications. Neural parameterizations of linear chain CRFs or high-order CRFs are learned with the framework proposed in chapter 6. We empirically demonstrate that this approach achieves substantial improvement using a variety of high-order energy terms. We also find high-order energies to help
in noisy data conditions.
   
    \item Chapter 8 summarizes the contributions of the thesis and discuss some future research directions. Our hope is that energy-based models to be applied to a larger set of natural language processing applications, especially text generation tasks in the future.

\begin{figure}[h]
\includegraphics[width=1.0\textwidth]{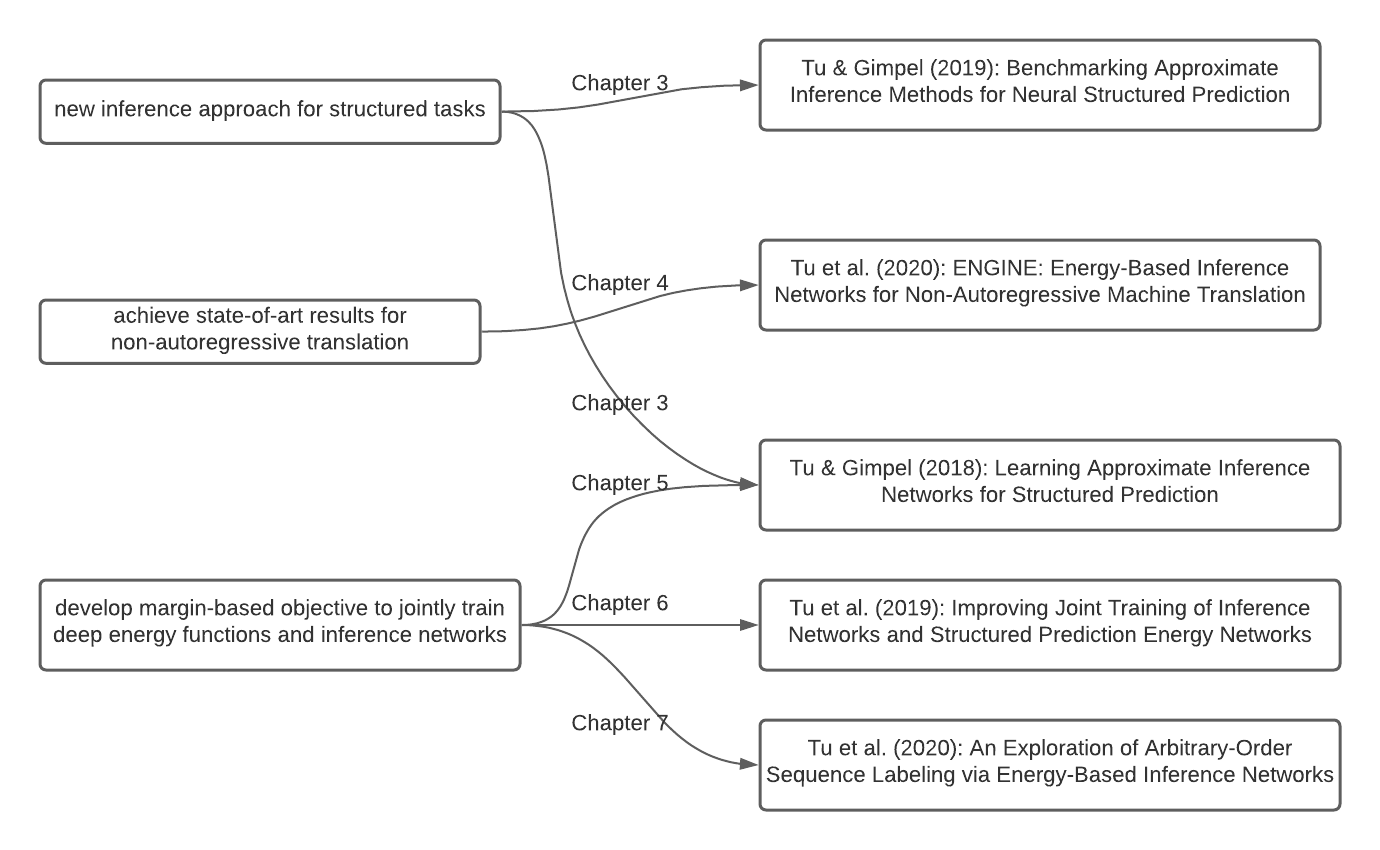}
\caption{Contributions of this thesis.\label{fig:contributions}}
\end{figure}

In a summary (see also Figure~\ref{fig:contributions}), we propose a method called ``energy-based inference network'' (or called ``an inference network''), which outputs continuous values that we treat as the output structure. The method could be easily applied for inference in the complex models with arbitrary energy functions. The time complexity of this method is also linear with the label set size. According to our experiments, “energy-based Inference networks” achieve a better speed/accuracy/search error trade-off than gradient descent, while also being faster than exact inference at similar accuracy levels. We also design a margin-based method that jointly learns energy function and inference networks.  We have applied the method on several NLP tasks, including multi-label classification, part-of-speech tagging, named entity recognition, semantic role labeling, and non-autoregressive machine translation .
\end{itemize}

\newpage

\mychapter{2}{Background}
\label{sec:background}

In this chapter, we introduce the energy-based models approach to structure prediction in NLP. The connections between energy-based models and previous approaches are discussed in particular. We then go over some related learning and inference methods for energy-based models. We will discuss our approaches to learning and inference for energy-based models in NLP structured applications in the following chapters.

\section{What are Energy-Based Models}

Energy-based models \citep{Contrastive,lecun-06, pmlr-v2-ranzato07a,belanger2016structured} associate a function that maps each point of a space to a scalar, which is called ``energy''. The map is called ``energy function''. It is a general framework. The point of the space could be  a sequence of acoustic signals, an image, or a sequence of tokens, etc. We can treat these models as part of them: language model~\citep{JelMer80, neuralLM, peters-etal-2018-deep, devlin-etal-2019-bert}, Autoencoder~\citep{Vincent08,vaeScoreMatch,ZhaoML16,xiao2021vaebm}, etc. 

For \textbf{structured applications in NLP}, the energy input space is input-output pairs $\Xspace \times \yspace$. We denote $\Xspace$ as the set of all possible inputs, and $\yspace$ as the set of all possible outputs. For a given input $\x\in\Xspace$, we denote the space of legal structured outputs by $\yspace(\x)$. We denote the entire space of structured outputs by $\yspace = \cup_{\x \in \Xspace} \yspace(\x)$. Here we use $\yspace(\x)$ to filter ill-formed outputs~\citep{smith:2011:synthesis}. Typically, $|\yspace(\x)|$ is exponential in the size of $\x$. The output space size is infinity in some cases (e.g., machine translation task).

The concept of an \textbf{energy function} $E_{\theta}$ used in my thesis:
\begin{align}
E_{\Theta}: \Xspace \times \yspace \rightarrow \mathbb{R} \nonumber
\end{align}
is parameterized by $\Theta$ that uses a functional architecture to compute a scalar energy for an input/output pair. The energy function can be an arbitrary function of the entire input/output pair, such as a deep neural network.

Given an energy function, the inference step is to find the output with lowest energy:
\begin{align}
\hat{\y} = \argmin_{\y\in\yspace(\x)}E_{\Theta}(\x, \y)
\end{align}
However, solving the above search problem requires combinatorial algorithms because $\yspace$ is a discrete structured space. It could become intractable when $E_{\Theta}$ does not decompose into a sum over small ``parts'' of $\y$.



\subsection{Connection with NLP}

In the NLP community, the concept ``score function'' is wildly used. The book by Smith~\citep{smith:2011:synthesis}, shows that many examples of linguistic structure are considered as output to be predicted from the text. They also demonstrate the standard approach in the NLP task is to define a score function:
\begin{align}
\score: \Xspace \times \yspace \rightarrow \mathbb{R} 
\end{align}

The \textbf{scoring function} is generally defined as a linear model:
\begin{align}
    \score(\x,\y) = W^\top F(\x,\y)
\end{align}
\noindent Where $F(\x, \y)$ is a feature extraction function and $W$ is a weight vector.

Search-based structured prediction is formulated over possible structure:
\begin{align}
    \textit{predict}(\x) = \argmax_{\hat{\y} \in \yspace(\x)} \score(\x,\y)
\end{align}
\noindent Where $\yspace(\x)$ is the set of all valid structures over $\x$.

In recent years, 
$\score$ replace the linear scoring function over parts with a neural network. 
\begin{align}
    \score(\x,\y) = \sum_{\textit{part} \in \y}\NN(\x, part)
\end{align}
\noindent Where $\textit{part}$ is a small part in $\y$. 

We can see that the concepts of \textbf{scoring function} and \textbf{energy function} are similar. Both of them define a function that map any point in one space to a scalar. Given an input $\x$, the goal of learning is to make the sample with ground truth label $\y$ have highest score. 

\subsection{Energy-Based Models for Structured Applications in NLP}

In this section, we show several widely used models in NLP. Table~\ref{tab:dsmodels} lists four different structured prediction methods, which are widely used before. 
 
All of them can be treated as special cases of energy-based models.

\begin{table}[h]
    \centering
    \small
    \begin{tabular}{c|c|l|}
    \cline{2-3}
      & modeling & learning \\ \hline
      
 \multicolumn{1}{|l|}{\multirow{2}{*}{transition-based}} & $P(\y \mid \x)=\Pi_t P(y_t \mid x, y_{t-1})$ & $\max_{\Theta} \sum_{\langle \x_i, \y_i\rangle\in\mathcal{D}} \log P_{\Theta}(\y_i \mid \x_i )$ \\ 
\multicolumn{1}{|l|}{} & locally normalized & Previous gold label is used during training. \\
\hline       
     
 \multicolumn{1}{|l|}{\multirow{2}{*}{CRF}} & A linear model; $P(\y \mid \x)$ is usually defined by & $\max_{\Theta} \sum_{\langle \x_i, \y_i\rangle\in\mathcal{D}} \log P_{\Theta}(\y_i \mid \x_i )$ \\ 
 \multicolumn{1}{|l|}{} & uniary potential and pair-wise potential. &  \\ 
 \hline

\multicolumn{1}{|l|}{\multirow{2}{*}{perceptron}} & The score function $S$ is usually linear weighted & $\min_{\Theta} \sum_{\langle \x_i, \y_i\rangle\in\mathcal{D}} [ \, \max_{\y} ( S_{\Theta}(\x_i,\y) - $ \\ 
\multicolumn{1}{|l|}{} & sum of the features, $S(x, y ) = W^\top f(\x,\y)$ & \qquad \qquad \quad $-S_{\Theta}(\x_i, \y_i) )]_{+}$ \\
\hline

\multicolumn{1}{|l|}{\multirow{2}{*}{large margin}}  & The score function $S$ is usually linear weighted  & $\min_{\Theta} \sum_{\langle \x_i, \y_i\rangle\in\mathcal{D}} [ \, \max_{\y} (\cost(\y, \y_i) + $ \\ 
\multicolumn{1}{|l|}{}  & sum of the features, $S(x, y ) = W^\top f(\x,\y)$ & \qquad \qquad \quad $S_{\Theta}(\x_i,\y) - S_{\Theta}(\x_i, \y_i) ) ]_{+}$ \\\hline
    \end{tabular}
    \caption{Comparisons of different structured models. $\mathcal{D}$ is the set of training pairs, $\langle \x_i, \y_i\rangle$ is one pair in the set, $[f]_+ = \max(0,f)$, and $\cost(\y,\y')$ is a structured \textbf{cost} function that returns a non-negative value indicating the difference between $\y$ and $\y'$.}
    \label{tab:dsmodels}
\end{table}

\begin{figure}[h]
\centering
\includegraphics[width=1.0\textwidth]{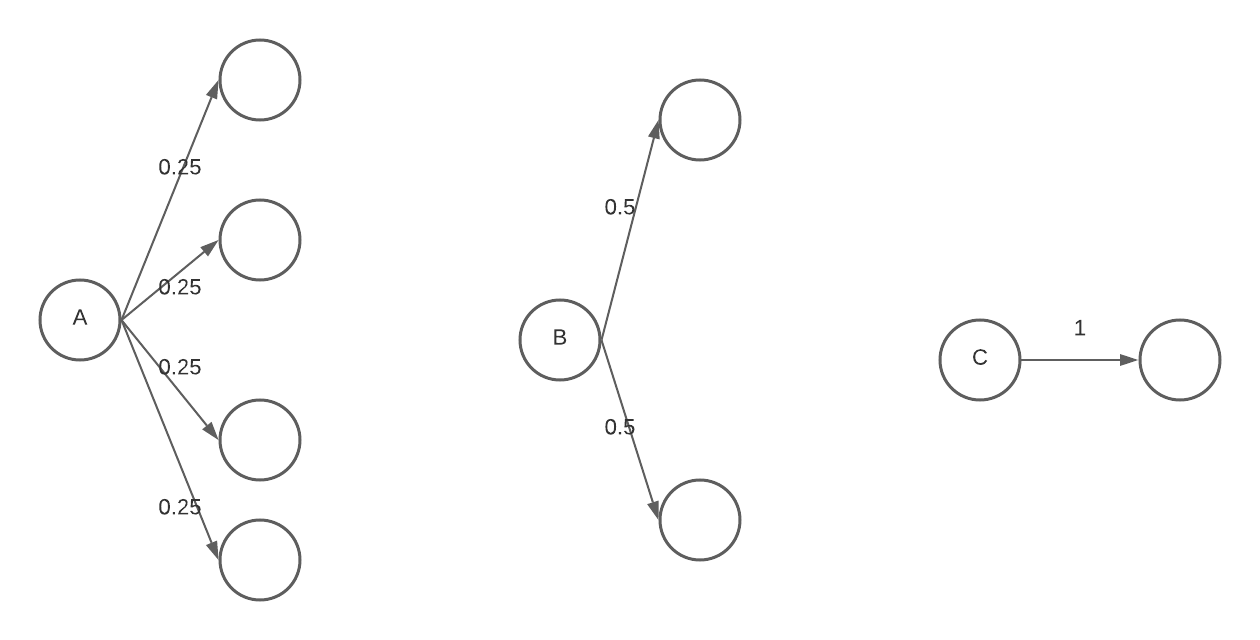}
\caption{This figure shows one label bias example. It shows $p(y_t \mid x_t, y_{t-1})$. Although at position $t-1$, there are three states ( A, B, and C) that have uniform conditional probability given current state. It means the threes states do not doing anything useful. However, the inference algorithm which maximizes $p(y_{1:t} \mid x_{1:t})$ will choose the path $y_{1:t}$ go through state $C$. The inference algorithm prefer to set $y_{t-1}=C$. \label{fig:labelBias}} 
\end{figure}


\paragraph{Local classifiers:} This is a widely used framework. Assume we have the features for a given sequence $\x$:
\begin{align}
    F(\x) = (F_1(\x), F_2(\x), \dots, F_{|\x|}(\x)) \nonumber
\end{align}
These could be a hand-engineered set of feature functions or by the way of a learned deep neural network, such as Long Short-Term Memory Networks (LSTMs)~\citep{Hochreiter:1997:LSM:1246443.1246450}.
For the local classifiers, the outputs are conditionally independent
given the features:
\begin{align}
    \log p(\y \mid \x) = \sum_i \log p(y_i \mid F_i(\x)) \nonumber
\end{align}
It is natural to use the $p(y_i \mid F_i(\x))$ to predict the tag at the position $i$. It is done with a trivial operations that computes the argmax of a vector. According to the above, we could see that the local classifiers are easy to train and do inference with. However, because of the independence assumptions , the expressive power of models could be limited. And it is hard to guarantee that the decoded output is a valid sequence, for example, a valid B-I-O tag sequence in named entity recognition task. This task contains sentences annotated with named entities and their types. There are four named entity types: PERSON, LOCATION, ORGANIZATION, and MISC. The English data from the CoNLL 2003 shared task~\citep{tjong-kim-sang-de-meulder-2003-introduction} is one popular dataset.

We can observe that the local classifiers completely ignore the current label when predicting the next label. For the predictions at position $i+1$ and $i$ can be done simultaneously   

In this case, the energy can be decomposed as a sum of energies for each tag:
\begin{align}
    E_\Theta(\x, \y) = \sum_i E_{\theta}(y_i \mid F_i(\x))
\end{align}

And,
\begin{align}
    E_{\theta}(y_i \mid F_i(\x)) = - \log p(y_i \mid F_i(\x)) \nonumber
\end{align}

According to recent work, the model can still achieve pretty good performance on some sequence labeling tasks with strong deep representations~\citep{peters-etal-2018-deep, devlin-etal-2019-bert}.

\begin{figure}[h]
\centering
\includegraphics[width=1.0\textwidth]{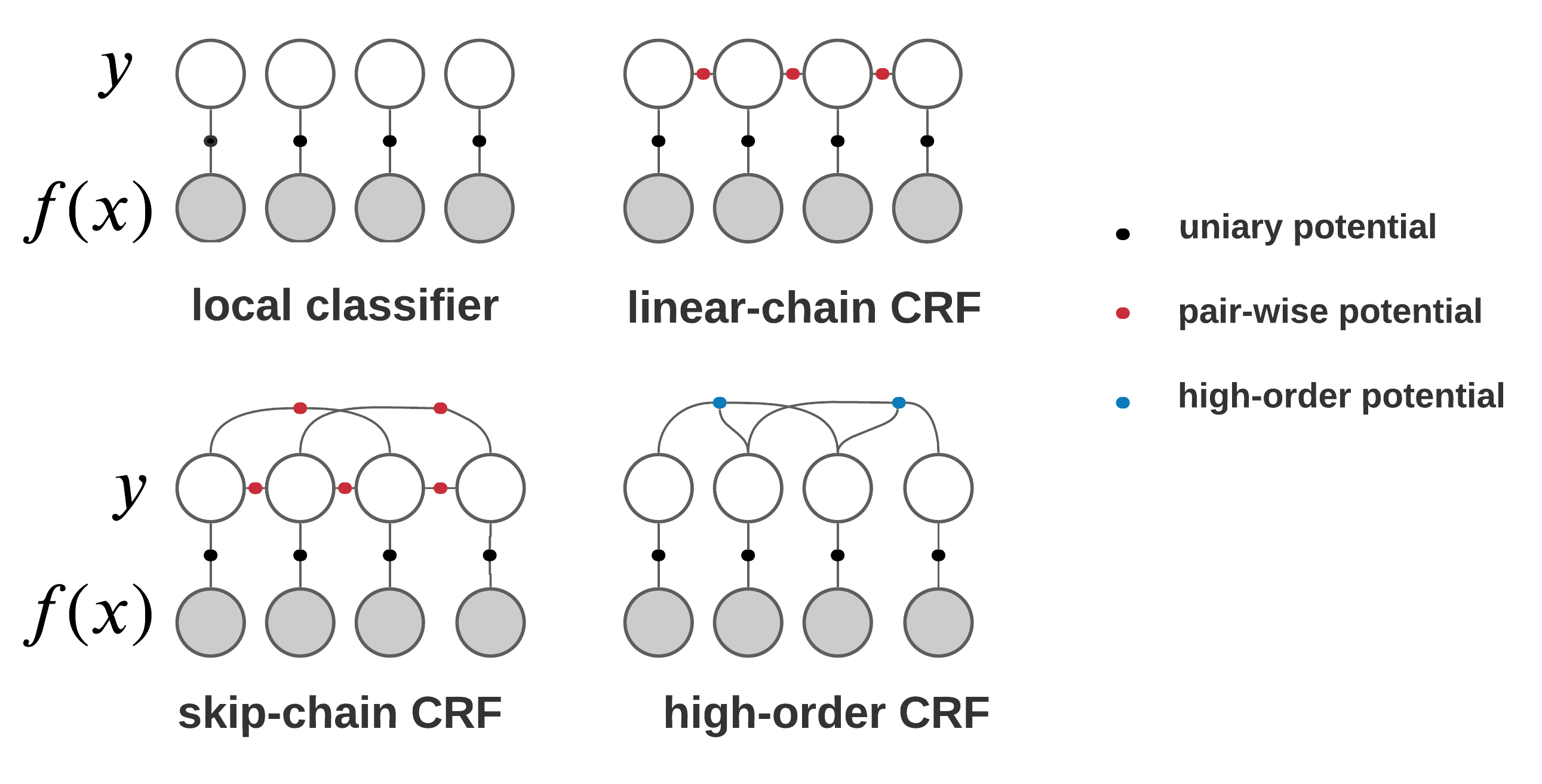}
\caption{Visualization of several discriminative structure models with different part sizes. $f(\x) = \langle f_1(\x), \dots, f_n(\x) \rangle $ is the representation of a given input $\x$. The decomposed parts for different discriminative structure models: local classifier, $\{\langle f_i(x), y_i> : 1 \leq i \leq n \}$; linear-chain CRF,  $\{\langle f_i(x), y_i \rangle : 1 \leq i \leq n \} \cup \{ \langle y_i, y_{i+1} \rangle : 1 \leq i \leq n-1 \} $; skip-chain CRF,  $\{\langle f_i(x), y_i \rangle : 1 \leq i \leq n \} \cup \{ \langle y_i, y_{i+1} \rangle : 1 \leq i \leq n-1 \} \cup \{ \langle y_i, y_{i+M} \rangle : 1 \leq i \leq n-M \} $;  high-order CRF: $\{ \langle f_i(x), y_i \rangle : 1 \leq i \leq n \} \cup \{\langle y_i, y_{i+1}, y_{i+2} \rangle : \langle i_1, i_2 \rangle \in  \mathcal{C} \} $. $\mathcal{C}$ is the set of long-range pair-wise potential. We did not consider sequence start symbol and end symbol here. \label{fig:parts}} 
\end{figure}

The energy function (score function) decomposes additively across parts. Each \textbf{part} is a sub-component of input/output pair. In chapter 2.2 of ~\citet{smith:2011:synthesis}, five views of linguistic structure prediction are shown. In the graphical model, each part is \textbf{clique}. Figure~\ref{fig:parts} shows the graphic model for different discriminative structured models. However, people typically uses small potential functions in order to enable tractable learning and inference. 
The top left figure shows the visualization of local classifier, which only include the uniary potentials. $\{\langle f_i(x), y_i> : 1 \leq i \leq n \}$. Linear-chain Conditional Random Field (CRFs)~\citep{Lafferty:2001:CRF} have a little large part size $\{\langle f_i(x), y_i \rangle : 1 \leq i \leq n \} \cup \{ \langle y_i, y_{i+1} \rangle : 1 \leq i \leq n-1 \} $ .
The complexity
of training and inference with CRFs, which are quadratic in
the number of output labels for first order models
and grow exponentially when higher order dependencies are considered. 

\paragraph{Conditional Log-Linear Models:} Linear chain CRFs~\citep{Lafferty:2001:CRF} and other conditional log-liner models, achieve strong performance on many structured NLP tasks. The scoring functions or energy functions have the following form:
\begin{align}
    E_{\Theta}(\x, \y) = w^\top f(\x,\y) \nonumber
\end{align}
\noindent where $f(\x, \y)$ is a feature vector of $\x$ and $\y$, which is called feature function. $w$ is a parameter vector. 

Particularly, linear-chain CRF has this following form:
\begin{align}
    E_{\Theta}(\x, \y) = -\left(\sum_{t} U_{y_t}^\top f(\x,t)
+ \sum_{t}W_{y_{t-1},y_t}\right) \nonumber
\end{align}
\noindent where $f(\x,t)$ is the input feature vector at position $t$,  $U_{i}\in\mathbb{R}^d$ is a parameter vector for label $i$  
and the parameter matrix $W\in\mathbb{R}^{L \times L}$ contains label pair parameters. 
The full set of parameters $\Theta$ includes 
the $U_i$ vectors,  
$W$, 
and the parameters of the input feature function. 
It solves the label bias problem. It has the efficient training and decoding based on dynamic programming for linear-chain CRF. However, it could be computationally expensive given a large label space. And the inference could be challenging for a general CRF framework.

\paragraph{Transition-Based Model:}
We can rewrite the conditional probability $p(\y \mid \x))$ as follows:
\begin{align}
    \log p(\y \mid \x) = \sum_i \log p(y_i \mid y_{1:i-1}, \x) \nonumber
\end{align}
In particular, we can rewrite the above equation
\begin{equation}
E_\Theta(\x, \y) = \sum_{t=1}^{|\y|} e_t (\x,\y) \noindent
\end{equation}
where
\begin{equation}
e_t(\x, \y) = - \y_t^\top \log p_{\Theta} (\cdot \mid \y_{0}, \y_{1}, \dots, \y_{t-1}, \x) \noindent
\end{equation}
\noindent Where $\y_t$ is the relaxed continuous representation of $y_t$. In the discrete case, it is a one-hot vector. In the continuous case, it can be probability of the $t$th position\footnote{We will use the formulation in chapter 4.}. $E(\x,\y)$ can be used to score a given language pair. $p(y_i \mid y_{1:i-1} \x)$ can be parameterized by Recurrent Neural Networks
(RNNs) or Long Short-Term Memory Networks (LSTMs). The whole energy function $E_{\theta}(\x,\y)$ can be represented by Sequence-to-sequence (seq2seq; \citealt{sutskever2014sequence}) models. It is common to augment models with an attention mechanism that focuses on particular positions of the input sequence while generating the output sequence~\citep{BahdanauCB14}. Recently, transformer-based models~\citep{NIPS2017_7181} are commonly used in machine translation, summarization, question answer, or other text-based generation tasks.

The joint conditional is modeled as the product of locally normalized probability distribution over all positions. During training, the true previous label is always used. This could cause mismatch between training and test time, which is exposure bias~\citep{RanzatoCAZ15}. It could also lead \textbf{label bias} issue~\citep{bottou-91a}: non-generative finite-state models based on next-state classifiers (e.g., discriminative markov models, maximum entropy Markov models~\citep{memm}), which are locally normalized, could ignore the current observation when predicting the next label. Figure~\ref{fig:labelBias} shows one example. In the work of ~\citep{wiseman-rush-2016-sequence}, they use beam-search training scheme to learn global sequence scores.

\paragraph{General Complex Energy} There has been a lot of work on using neural networks to define the potential functions in the discriminative structure models, e.g., neural CRF~\citep{passos-etal-2014-lexicon},  RNN-CRF~\citep{DBLP:journals/corr/HuangXY15,lample-etal-2016-neural},  CNN-CRF~\citep{journals/jmlr/CollobertWBKKK11} etc. However the potential functions are still limited in size. \citet{belanger2016structured} formulated deep energy-based models for structured prediction, which they called structured prediction energy networks (SPENs). 
SPENs use \textbf{arbitrary neural networks} to define the scoring function over input/output pairs. For example, they define the energy function for multi-label classification (MLC) as the sum of two terms: 

\begin{align*}
    E_{\Theta}(\x,\y) = E^{\mlocal}(\x,\y) + E^{\mlabel}(\y)
\end{align*}

$E^{\mlocal}(\x,\y)$ is the sum of linear models:
\begin{align}
E^{\mlocal}(\x,\y)= \sum_{i=1}^L y_{i}b_{i}^\top F(\boldsymbol{x})
\end{align}
where $b_{i}$ is a parameter vector for label $i$ and $F(\x)$ is a multi-layer perceptron computing a feature representation for the input $\x$. 
$E^{\mlabel}(\y)$ scores $\y$ independent of $\x$:
\begin{align}
E^{\mlabel}(\y) = c_{2}^\top g(C_{1}\y)
\end{align}
where $c_2$ is a parameter vector, $g$ is an elementwise non-linearity function, and $C_1$ is a parameter matrix.


Recently, structured models have been combined with deep nets~\citep{passos-etal-2014-lexicon,DBLP:journals/corr/HuangXY15,lample-etal-2016-neural,journals/jmlr/CollobertWBKKK11, 8682728, MMS:CVPR:2018,asm2019, GraberNIPS2018,zhang-etal-2019-empirical}. However the potential functions are still limited.  To address the shortcoming, energy-based models are proposed, for instance, SPENs~\citep{belanger2016structured} and GSPEN~\citep{GraberNeurIPS2019}. They do not allow for the explicit specification of output structure. Recently, ~\citet{Grathwohl2020Your} also demonstrate that energy based training of the joint distribution improves calibration and robustness.

Although energy-based models have the strong ability to model complex structured components, they have had limited application in NLP due to the computational challenges involved in learning and inference in extremely large search spaces. In the next two subsections, we describe background on learning and inference. It is mainly from the perspective in NLP community.

\section{Learning of Energy-Based Models}

At first, we discuss several ways for energy-based learning. There are two different approaches: probabilistic and non-probabilistic learning.

\subsection{Log loss}

\paragraph{Probabilistic}

We can learn the model parameters $\theta$, by maximizing the probability of a training set $\mathcal{D}$ of data:

\begin{align}
    \mathcal{L} = \frac{1}{N} \sum_{\y \in \mathcal{D} } \log p_{\theta}(\y) = \frac{1}{N} \sum_{\y \in \mathcal{D} } \log \frac{\exp(-E_{\theta}(\y))}{Z(\theta)} = -\log Z(\theta) -\frac{1}{N} \sum_{\y \in \mathcal{D}} E_{\theta}(\y) \nonumber
\end{align}
\noindent $N$ is the number of examples in training set $\mathcal{D}$.
\begin{align}
    Z(\theta) = \int_{\y} \exp(-E_{\theta}(\y)) \nonumber
\end{align}
And,
\begin{align}
    p(\theta) = \frac{\exp(-E_{\theta}(\y))}{Z(\theta)} \nonumber
\end{align}

We will derive the gradient equation by firstly writing down the
partial derivative:
\begin{align}
\frac{\partial \mathcal{L}}{\partial \theta} = -\frac{\partial \log Z(\theta)}{\partial \theta} - \frac{1}{N} \sum_{\y \in \mathcal{D}} \frac{\partial E_{\theta}(\y)}{\partial \theta}
\label{eq:mlegd}
\end{align}
The first term compute the gradient from the partition function $Z(\theta)$, which involves an integration over $\y$. Then we have:
\begin{align*}
    \frac{\partial \log Z(\theta)}{\partial \theta} = &\frac{1}{Z(\theta)} \frac{\partial Z(\theta)}{\partial \theta} \\
             =& \frac{1}{Z(\theta)} \frac{\partial \int_{\y} \exp(-E_{\theta}(\y))}{\partial \theta} \\
             = & \frac{1}{Z(\theta)} \int_{\y} \frac{\partial  \exp(-E_{\theta}(\y))}{\partial \theta} \\
             = & \frac{1}{Z(\theta)} \int_{\y} \exp(-E_{\theta}(\x)) \frac{\partial E_{\theta}(\y)}{\partial \theta} \\
             = & - \frac{\exp(-E_{\theta}(\y)) }{Z(\theta)} \int_{\y} \frac{\partial E_{\theta}(\y)}{\partial \theta} \\
             = & - \int_{\y} p_{\theta}(\y) \frac{\partial E_{\theta}(\y)}{\partial \theta}
\end{align*}

By putting above results into Equation~\ref{eq:mlegd}:
\begin{align}
\frac{\partial \mathcal{L}}{\partial \theta} = \int_{\y} p_{\theta}(\y)\frac{\partial E_{\theta}(\y)}{\partial \theta} - \frac{1}{N} \sum_{\y \in \mathcal{D}} \frac{\partial E_{\theta}(\y)}{\partial \theta}
\end{align}
The first term could be hard and intractable. The expectation is over the model distribution. 

For conditional models, we parameterize the conditional probability $p_{\theta}( \y \mid \x)$, similarly we can get:

\begin{align*}
    \frac{\partial \mathcal{L}}{\partial \theta} = & \frac{\partial -\log p_{\theta}(\y \mid \x) }{\partial \theta}  \\
     = & \frac{\partial E_{\theta}(\x, \y)}{\partial \theta} - \int_{\y'} p_{\theta}(\y' \mid \x)\frac{\partial E_{\theta}(\x, \y')}{\partial \theta}
\end{align*}

Typically, it is not easy to do the sampling from the model distribution. It leads in interesting research question how to approximate the gradient. Following are several previous methods.

\paragraph{Contrastive Divergence:} To avoid the computation difficulty of log-likelihood gradient, ~\citet{Contrastive} uses contrastive divergence to approximate the gradient.

\begin{align}
    \frac{\partial \mathcal{L}}{\partial \theta} = \E_{\y \in \bar{p}} \frac{\partial E_{\theta}(\y)}{\partial \theta} - \E_{\y \in p_d} \frac{\partial E_{\theta}(\y)}{\partial \theta}
\end{align}
\noindent where $\bar{p}$ is the Markov Chain Monte Carlo sampling distribution from data distribution $p_d$. In the work, they run the chain for a small number of steps (e.g. 1). However, this technique relies on the particular form of the energy function in the case of products of experiments, which is naturally fit to Gibbs sampling. The intuition behind is that after a few iterations, the data moves towards the proposed distribution.

\paragraph{Importance Sampling:}
It is hard to sample from model distribution in the above equation especially if vocabulary size is large. The idea of importance sampling is to generate $k$ samples $\bar{\y_1}$, $\bar{\y_2}$, $\dots$, $\bar{\y_k}$ from an easy-to-sample-from distribution $Q$. This can be a n-gram language model.
If $\y$ is a token or sequence of tokens, The first term in Equation~\ref{eq:mlegd} can be approximate as following:
\begin{align}
\int_{\y} p_{\theta}(\y)\frac{\partial E_{\theta}(\y)}{\partial \theta} \approx  \sum_{j=1}^{k}\frac{v(\bar{\y_j})}{V} \frac{\partial E_{\theta}(\y_j)}{\partial \theta}
\label{eq:ISgd}
\end{align}
\noindent where $V = \sum_{k} v(\bar{\y_j})$ and $v(\bar{\y}) = \frac{\exp(-E_{\theta})}{Q(w=\y)}$. The normalization by $V$ is computed with unnormalized model distribution $E_{\theta}(\y)$. However, the weight term $v(\bar{\y}) = \frac{\exp(-E_{\theta})}{Q(w=\y)}$ can make learn unstable because value is with high variance. In order to reduce the variance, one way is to increase the number of samples during training. In the work of ~\citet{Bengio03quicktraining}, a few sampled negative example words are used for language model training. A very significant speed-up is obtained. 

\paragraph{Score Matching~\citep{hyvarinen05a} and Langevin dynamics~\citep{Neal93probabilisticinference,pmlr-v2-ranzato07a}:}
These two method are \textbf{not applicable when input is discrete}. Both of the two method need to calculate the gradient w.r.t. the random variable $\y$. 
For score matching~\citep{hyvarinen05a}, the object bypass the intractable unnormalized constant term $Z$ as the following objective:
\begin{align*}
    \mathcal{L} = 0.5* \E_{\y \in p_d } || \frac{\partial \log p_d(\y)}{\partial \y} - \frac{\partial E_{\theta}(\y)}{\partial \y} ||^2 
\end{align*}
\noindent where $\textit{const}$ is a constant number and $p_d$ is the data distribution.

For Langevin dynamics, it iterative update from initial sample $x_0$ to draw sample from model distribution as following:
\begin{align*}
    \y_{t+1} = \y_t - 0.5*\eta* \frac{\partial E_{\theta}(\y_t)}{\partial \y_t} +  \omega
\end{align*}
\noindent $\eta$ is the step size and $\omega \in \mathcal{N}(0, \eta)$ is Gaussian noise.
With these samples $\y_0, \y_1,\dots$, the gradient from normalization term $Z$ is approximated.

\paragraph{Noise-Contrastive Estimation (NCE)~\citep{pmlr-v9-gutmann10a}} NCE is a more stable method for effective training. It uses logistic regression to distinguish between the data samples from the distribution $p_{\theta}$ and noise samples that are generated from a noise distribution $p_{n}$. If we assume the noise samples are $k$ times more frequent than data samples, then the posterior probability that sample $w$ came from the data distribution is :
\begin{align}
    P (D =1 \mid w) = \frac{p_{d}(w)}{p_{d}(w) + k * p_{n}(w)} \nonumber
\end{align}
\noindent where $p_{d}$ is the data distribution. We use $p_{\theta}$ in place of $p_d$ in abve equation, then
\begin{align}
    P (D =1 \mid w) = \frac{p_{\theta}(w)}{p_{\theta}(w) + k * p_{n}(w)} \nonumber
\end{align}

With this posterior probability, the training objective is to maximized the following:

\begin{align}
    \mathcal{L} = \E_{w \in p_d } \log P (D =1 \mid w) + k*\E_{\hat{w} \in p_n }\log P (D =0 \mid \hat{w}) \nonumber
\end{align}

And the gradient can be expressed as:
\begin{align*}
    \frac{\partial \mathcal{L}}{\partial \theta} 
     = & \E_{w \in p_{d} } \log \frac{p_{d}(w)}{p_{\theta}(w) + k * p_{n}(w)} + k*\E_{\hat{w} \in p_n } \log \frac{k * p_{n}(w)}{p_{\theta}(w) + k * p_{n}(w)} + k*\E_{\hat{w} \in p_n }    \\
     = & \E_{w \in p_{d} }  \frac{k*p_n(w)}{p_{\theta}(w) + k * p_{n}(w)} \frac{\partial \log p_{\theta}(w)}{\partial \theta} - k*\E_{\hat{w} \in p_n } \frac{p_{\theta} (\hat{w})}{p_{\theta}(\hat{w}) + k * p_{n}(\hat{w})}\frac{\partial \log p_{\theta}(\hat{w})}{\partial \theta}   \\
    =& \sum_{w} \frac{k*p_{n} (w)}{p_{\theta}(w) + k * p_{n}(w)} (p_{d} - p_{\theta}) \frac{\partial \log p_{\theta}(w)}{\partial \theta}
\end{align*}

We can see that $k \to \infty$ 
then:

\begin{align}
    \frac{\partial \mathcal{L}}{\partial \theta} \to \sum_{w} (p_{d} - p_{\theta}) \frac{\partial \log p_{\theta}(w)}{\partial \theta}
\end{align}
\noindent The gradient is 0 when the model distribution $p_{\theta}$ match the empirical distribution $p_d$

The good property is that the weight $\frac{p_{d}(w)}{p_{\theta}(w) + k * p_{n}(w)}$ are always between 0 and 1. This leads NCE training more stable than importance sampling.

Chris's note~\citep{ncenote} shows some analysis on NCE and negative sampling.   Negative sampling method is used in the paper~\citep{mikolov2013distributed}. It is similar to a special case for NCE. If there is self-normalized assumption for the learned model distribution $P_d$, and the noise distribution $p_n = \frac{1}{V}$ and $k = V$. The objective is not to optimize the likelihood of the language model. It is appropriate for representation learning, which is not consistent with language model probabilities.

\subsection{Margin Loss}
The one wildly used objective for binary classification is the support vector machine (SVM; ~\citealt{cortes1995support}). Instead of a probabilistic view that transform $\score(\x, \y)$ or $E(\x,\y)$ into a probability, it takes a geometric view~\citep{smith:2011:synthesis}. Hinge loss with multiclass setting attempts to score the correct class above all other classes with a margin. The margin is generally set as 1. In some tasks, the margin is set as hamming loss, L1, or L2 loss.

\paragraph{Ranking Loss:}

In some settings, there is no any supervision (with labels). However, there are a pair of correct and incorrect one $\y$ and $\y'$. We can use pairwise ranking approach~\citep{rankingloss}. It is a popular loss in NLP applications. 
\begin{align}
    \mathcal{L} (\y, \y') = [
    \cost + E(\y) - E(\y') ] 
\end{align}
In the work of ~\citet{journals/jmlr/CollobertWBKKK11}, $\y$ is one possible text windows, $\y'$ is the text window that the central word of text $\y$ by another word. They use the ranking loss for learning word embeddings. In the next part, we will talk about hinge loss used in strucutred application in NLP.

\paragraph{Margin-based loss:}
Structured Perceptron~\citep{collins2002discriminative} describe an algorithm for training discriminative models, for example CRF. Usually Viterbi algorithm or other algorithms are used rather than an exhausive search in the exponentially large label space.
\begin{align}
    \mathcal{L} = \sum_{\langle\x, \y\rangle \in \mathcal{D}}\max_{\bar{\y}} [ E(\x, \y) - E(\x, \bar{\y}) ]_{+} \nonumber
\end{align}
where $\mathcal{D}$ is the set of training pairs, $[f]_+ = \max(0,f)$. 
As argued in (~\citealt{lecun-06}, Section 5), the perceptron loss may not be a good loss function when training structured prediction neural networks as it does not have a margin. 

Max-margin structured learning~\citep{ssvm,m3} uses the following loss:
\begin{align}
    \mathcal{L} = \sum_{\langle\x, \y\rangle \in \mathcal{D}}\max_{\bar{\y}} [ \cost(\y, \bar{\y}) - ( E(\x, \bar{\y})- E(\x, \y) ) ]_{+}  \nonumber
\end{align}
\noindent where $\cost$ is an non-negative term, which could be a constant number. It is to measure the difference between the candidate output $\bar{\y}$ and ground-truth output $\y$. 

In the previous work, this loss is used to learning a linear model $E(\x,\y) = -S(\x, \y ) = -W^\top f(\x,\y)$. Recently, ~\citet{belanger2016structured} use above objective to learn Structured Prediction Energy Networks. ``cost-augmented inference step'' $\max_{\bar{\y}} (\cost(\y, \bar{\y}) - E(\x, \bar{\y}))$ is done with gradient descent based inference. We describe gradient descent based inference in the next subsection.

There are some theory analysis and learning bounds in the work~\citep{m3, ssvm}. However, in the neural-network framework, the objectives are no longer convex, and so lack the formal guarantees and bounds associated with convex optimization problems. Similarly, the theory, learning bounds, and guarantees associated with the algorithms do not automatically transfer to the neural versions.

A model trained with this objective is often called a structure SVM. It enforces the model to learn good scoring functions when incorporating cost function $\cost$. 

There are also several other losses mentioned in Section 2 in the tutorial~\citep{lecun-06}.

\subsection{Some Discussion on Different Losses}

\begin{table*}[h]
    \centering
    \begin{tabular}{c|c|c|}
    \cline{2-3}
      & learning objective & gradient or sub-gradient \\ \hline
      
 \multicolumn{1}{|l|}{\multirow{1}{*}{log}} & $\mathcal{L}= - \log p_{\theta}(\y \mid \x) = \log \frac{\exp{E_{\theta}(\x,\y)}}{\sum_{y'}\exp(E_{\theta}(\x, \y'))} $   & $\frac{\partial E_{\theta}(\x, \y)}{\partial \theta} - \int_{\y'} p_{\theta}(\y' \mid \x)\frac{\partial E_{\theta}(\x, \y')}{\partial \theta} $ \\ 
 \hline

 \multicolumn{1}{|l|}{\multirow{2}{*}{perceptron}} & $\mathcal{L}= [\max_{\y'} E_{\theta}(\x, \y) - E_{\theta}(\x, \y')]_{+}  $   & $\frac{\partial E_{\theta}(\x, \y)}{\partial \theta} - \frac{\partial E_{\theta}(\x, \bar{\y})}{\partial \theta} $ or 0 \\ 
 \multicolumn{1}{|l|}{} & & where $\bar{\y} = \argmin_{\y'} E_{\theta}(\x, \y')$ \\
 
 \hline
 
  \multicolumn{1}{|l|}{\multirow{1}{*}{margin}} & $\mathcal{L}= [\max_{\y'} \cost(\y, \y') + E_{\theta}(\x, \y) - E_{\theta}(\x, \y')]_{+}  $   & $\frac{\partial E_{\theta}(\x, \y)}{\partial \theta} - \frac{\partial E_{\theta}(\x, \bar{\y})}{\partial \theta}$ or 0 \\ 
   \multicolumn{1}{|l|}{} & &  where $\bar{\y} = \argmin_{\y'} E_{\theta}(\x, \y')- \cost(\y, \y')$ \\
 \hline
    \end{tabular}
    \caption{Comparisons of different learning objectives. $[f]_+ = \max(0,f)$, and $\cost(\y,\y')$ is a structured \textbf{cost} function that returns a nonnegative value indicating the difference between $\y$ and $\y'$.}
    \label{tab:gradient}
\end{table*}

\paragraph{Generalization} Table~\ref{tab:gradient} shows the gradient of subgradient of different objectives. For log loss, given the input $\x$,  the optimizer will push down the energy of data with ground truth label $\y$, and push up the energies of the other labels. It continues this process without stopping. However, for perceptron or margin-based loss, the gradient can be zero when the energy of ground truth label $\y$ is smaller than others with a margin. Maximum likelihood training can easily lead to overfitting models on the training data without any regularizer. On the other hand, perceptron or margin-based loss will have zero gradients when the optimization is done well.

\paragraph{Probabilistic VS Non-Probabilistic Learning}

With log loss, we usually learn data distribution with likelihood training. However, a margin-based loss does not have a probabilistic interpretation. They can only answer the decoding question. It does not provide joint or conditional likelihood. The good thing is that the margin-based learning use cost function $\cost$, which is defined by the task and is related to goal or performance metric. This provides an opportunity to learn models. 

For probabilistic learning, as mentioned in (~\citealt{lecun-06}, Section 1.3),  it constrains $\int_{\y}\exp(-E(\x,\y))$ converges and domain $\yspace$ that can be used. Hence probabilistic learning comes with a higher price. LeCun stated that probabilistic modeling should be avoided when the application does not require it. More discussions or experiments could be done in the future.

\paragraph{Negative Examples}

In the log loss, the gradient term from partition function:
\begin{align*}
    \int_{\y'} p_{\theta}(\y' \mid \x)\frac{\partial E_{\theta}(\x, \y')}{\partial \theta} 
\end{align*}
All the structured output space is considered during training. They are all ``negative examples''. The computation could be intractable. So approximation is done: contrastive divergence and importance sampling are used. 

In the SSVM loss, there is one step called ``cost-augmented inference step'':
\begin{align*}
    \bar{\y} = \argmin_{\y'} E_{\theta}(\x, \y')- \cost(\y, \y')
\end{align*}
Only one negative example is used during training. However, this step could be hard and intractable. 

We can see the learning signal of different objectives depend on the negative examples used.  

~\citet{smith-eisner-2005-contrastive} use contrastive criterion which estimates the likelihood of the data conditioned to a ``negative neighborhood'': all sequences generated by deleting a single symbol,  transposing any pair of adjacent words, deleting any contiguous subsequence of words.   
~\citet{journals/jmlr/CollobertWBKKK11} uses ranking loss to learn word embedding. The negative examples are the text window that the central word of text $\x$ by another word. So hinge loss can ``inject domain
knowledge``: not only the observed positive examples, but also a set of similar but deprecated negative examples. 

And ``cost-augmented inference step'' can be intractable and/or exact maximization has some undesirable quality (e.g., it’s an
alternative viable prediction). In this case, maximization is replaced by sampling
~\citet{wieting-16-full} select the negative samples from the current minibatch.

Noise-Contrastive Estimation~\citep{pmlr-v9-gutmann10a} are used for energy-based models training in some recently work~\citep{ouSLT2018, bakhtin2020energybased}. The noise samples that are generated from the noise distribution can be understood as ``negative examples''. The negative examples are sampled from pre-trained language models.
Importance of negative examples also been shown in multimodel learning~\citep{Kiros2014UnifyingVE}, open-domain question answering~\citep{karpukhin-etal-2020-dense}, model robustness~\citep{tu20tacl} etc.

\paragraph{Directly Optimizing Task Metrics}
It is a popular approach to use maximum likelihood estimation (MLE) for learning models. However, the performance of these models is typically evaluated with task metrics, e.g., accuracy, F1, BLEU~\citep{papineni-etal-2002-bleu}, ROUGE~\citep{lin-2004-rouge}. In the previous work, reinforcement learning (RL) objective~\citep{RanzatoCAZ15,RAML}, which is to maximize the expect reward (task metrics) over trajectories by the policy, is used. In particular, the actor-critic approach~\citep{ActorCritic} train the actor by policy gradient with advantages of the critic. AlphaGo~\citep{alphago} use the actor-critic method for self-learning in the game of Go: a value network (critic) is to evaluate positions, and a policy network (actor) is to sample actions. However, there are still many challenges in RL for sparse rewards. 

~\citet{DVNGygliNA17} proposes a deep value network (DVN) to estimate task metrics on different structured outputs. In their work, the deep value network is trained on tuples comprises an input, an output, and a corresponding oracle value (task metrics). Gradient descent\footnote{We will discuss this inference method in the next subsection.} is used for inference to iteratively find better output, which is with lower value. It would be interesting to explore other ways of learning energy functions, which can estimate task metrics on structured output. 








\section{Inference}

In the structured applications, we need to search of $\y$ with the lowest energy over the structured output space $\yspace(\x)$, which is generally exponentially large. The search space size could be even infinity if the target sequence length is unknown. The inference problem is challenging.

\begin{align*}
    \argmin_{\y\in\yspace(\x)}E_{\Theta}(\x, \y) \label{eq:inf}
\end{align*}

In this section, several popular inference methods in NLP are summarized here.

\paragraph{Greedy Decoding}

One simple decoding method used in the structured applications is greed decoding. Once we know probability $p(y_i \mid .)$, we can do the argmax operation for position $i$ over distribution vector. 
\begin{align}
    \bar{y} = \argmax_{y_i}{p(y_i \mid .) } \nonumber
\end{align}

We can do the heuristic operations over the whole inference process for each position. It is a faster decoding method. However, there are some constraints.

If the model is a local classifier, greed decoding is a natural choice. However, a local classifier have a strong conditional independent assumption, which can limit model performance.
    
For other models, like a transition-based model, the greedy approach suffers from error propagation. The mistakes in early decisions influence later decisions. For autoregressive models,
\begin{align*}
    \min_{\y} E_{\theta}(\x,\y) =&  \min_{\y}-\log p_{\theta}(\y \mid \x) = \min_{\y} - \sum_i \log p_{\theta}(\y_i \mid \y_{<i} , \x) \\
                      =& \min_{\y} - \sum_i \log p_{\theta}(\y_i \mid \y_{<i} , \x)
\end{align*}

To solve above optimization problem, one easy solution is to do argmin operation for each term $\y'_i =\min_{\y_i} -\log p_{\theta}(\y_i \mid \y_{<i} , \x)$, which is called greedy decoding. However, the greed decoding output $\y'$ is usually sub-optimal, because 

\begin{align*}
    \min_{\y} - \sum_i \log p_{\theta}(\y_i \mid \y_{<i} , \x) \leq  \sum_i \min_{\y'} -\log p_{\theta}(\y'_i \mid \y'_{<i} , \x)
\end{align*}

\paragraph{Dynamic Programming}
Viterbi algorithm~\citep{Viterbi67} is one of the popular dynamic programming algorithms for finding the most likely sequence in NLP.  
In CRF or HMM, the conditional probability $\log p(\y \mid \x)$ could be decomposed similarly. 
\begin{align}
   \log p(\y \mid \x) = \sum_{i=1}^{|x|} \score_1(y_i, y_{i-1}) + \score_2(y_i, \x) \nonumber
\end{align}


\noindent here $\score_1(y_i, y_{i-1})$ is a bigram score between the label $y_i$ and $y_{i-1}$, $\score_2(y_i, \x)$ is a uniary score at position $i$ with label $y_i$. Particularly, in HMM, $\score_1(y_i, y_{i-1}) = \log p_{\eta}(y_i \mid y_{i-1})$, and $\score_2(y_i, \x)=  \log p_{\tau}(x_i \mid y_i)$.
The inference in HMMs or CRF is done with the following optimization:
\begin{align}
    \argmax_{\y} \sum_{i=1}^{|x|} \score_1(y_i, y_{i-1}) + \score_2(y_i, \x)
\end{align}
The above optimization problem could be solved with the dynamic programming algorithm. We set a variable $V(m, y')$, which means the probability of sequence starting with label $y'$ at the position $m$. Then we have:

\begin{align}
    V(1,\hat{y}) = &  \score_1(\hat{y}, \langle s \rangle) + \score_2(\hat{y}, \x)   
    \nonumber \\
    V(m,\hat{y}) =&  max_{y'} ( \score_1(\hat{y}, y') + \score_2(\hat{y}, \x) +  V(m-1, y') ) \nonumber
\end{align}
$\langle s \rangle$ is the start sequence symbol.The second equation could be done recursively. If we consider that the last symbol is the end symbol $\langle /s \rangle$, then the output sequence $\y_{|\x|}$ is:
\begin{align}
    \argmax_{y'}\score_1(</s>, y') + V(|\x|, y') \nonumber
\end{align}
$\y_{|\x|-1}$, $\y_{|\x|-2}$,..., $\y_{2}$, $\y_{1}$ are computed recursively. The time complexity is $\mathcal{O}(nL^2)$, where $n$ is the sequence length and $L$ is the size of the label space. 

For energy function has the similar form:
\begin{align*}
    E_{\theta} (\x,\y) = \sum_{i=1}^{|x|} \score_1(y_i, y_{i-1}) + \score_2(y_i, \x)
\end{align*}
Then, Viterbi algorithm can be used for decoding. However, the time complexity is $\mathcal{O}(nL^2)$. If the label set size $L$ is larger, e.g., large word vocabulary size, it is not doable.  

\paragraph{Coordinate Descent}
Coordinate descent algorithms~\citep{coordinate} solve optimization problems by successively performing approximate minimization along coordinate directions or coordinate hyperplanes. 

When the number of coordinates is large, it is computationally expensive to solve the optimization problem. To find the optimal solution, it makes sense to search each coordinate direction,  decreasing the objective. One potential benefit is that it is computationally cheap to search along each coordinate. Algorithm~\ref{ag:coor} is shown below 

\hspace{1cm}

\begin{algorithm}[H]
\SetAlgoLined
 \KwIn{Given Energy Function:$E_{\Theta}$, Max Iteration Number $T_{max}$}
 \KwOut{$\y$}
 initialization $\y^{(0)}$ \;
 \While{$t < T_{max}$}{
  choose index $i \in \{1,2,\dots, n \}$\; 
  $y^{(t+1)}_{i} \leftarrow  \argmin_{y_i} E_\Theta(\x, y_i, \y^{(t)}_{-i})$ \; 
 
 }
 \caption{Coordinate Descent for finding $\argmin_{\y\in\yspace(\x)}E_{\Theta}(\x, \y)$}\label{ag:coor}
\end{algorithm}
$\y_{-i}$ represent all other coordinates except $i$.

There are mainly two ways to choose $\y_{-i}$ and many ways to choose the coordinate:

\begin{itemize}
    \item Gauss-Seidel style 
    \begin{align*}
        \y^{(t)}_{-i} = (y^{(t+1)}_1, \dots,y^{(t+1)}_i, y^{(t)}_{i+1}, \dots, y^{(t)}_{n}  )
    \end{align*}
    when updating each coordinate, the Gauss-Seidel style fixes the rest coordinates to be most up-to-date solution. It generally converges faster.
    \item Jacobi style
    \begin{align*}
        \y^{(t)}_{-i} = (y^{(t)}_1, \dots,y^{(t)}_i, y^{(t)}_{i+1}, \dots, y^{(t)}_{n}  )
    \end{align*}
    When updating each coordinate, the Jacobi style fixes the rest coordinates to the solution from previous circle. So the Jacobi style can update coordinate in parallel for each circle. 
\end{itemize}

Rules for selecting coordinates:
\begin{itemize}
    \item Cyclic Order: choose coordinate in cyclic order, i.e. $1 \to 2 \dots \to n$
    \item Randomly Sampling: randomly select coordinates
    \item Easy-First (Gauss−Southwell): pick coordinate $i$ so that $i = \argmax_{1 \leq i \leq n} \bigtriangledown E_{\theta}(\x, y_1, \dots,  y_n) $ 
\end{itemize}

\paragraph{Beam Search}

As mentioned in the above paragraphs, the greedy decoding approach likely does not find the optimal solutions for autoregressive models. Algorithm~\ref{ag:bs} is shown below. 

\hspace{1cm}

\begin{figure}[h]
\centering
\includegraphics[width=0.8\textwidth]{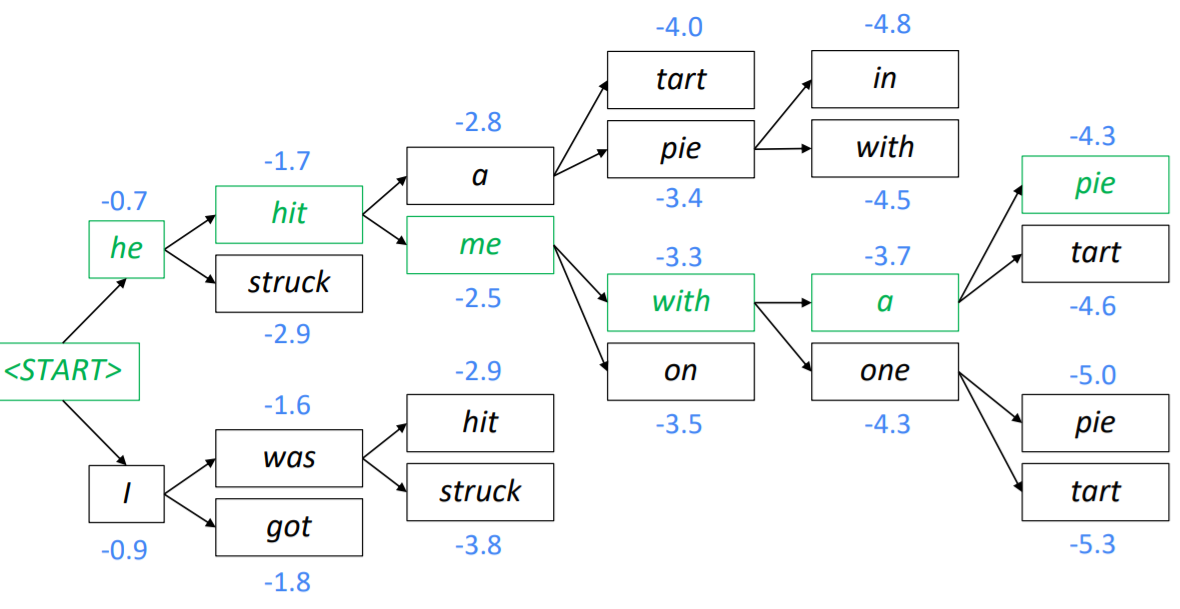}
\caption{A beam search example with beam size $=$ 2. The top score hypothesis is shown in green. The blue numbers are $\score(\x, \y) = - E(\x,\y)$. So the top score hypothesis is the hypothesis with larger score in the beam. \label{fig:bs}.  }
\end{figure}

\begin{algorithm}[H]
\SetAlgoLined
 \KwIn{Given Score Function:$E(\x, \y )$, Beam Size $K$, Max Iteration Number $T_{max}$}
 \KwOut{$\bar{\y}$}
 set $\bar{\y} \longleftarrow null$ \;
 set $\bar{\y}^{1:K} $ with $K$ copies\;
 \While{$t < T_{max}$}{
  \# walk over each stop\;
  \# the succession of a competed hypothesis is itself \;
  $\bar{\y}^{1:K} \longleftarrow \textit{TopK}(\cup_{k=1}^{K} \textit{succ}(\x, \bar{\y}^{k})) $\; 
  \For{$k=1, \dots, K$}
  {\If{$\bar{\y}^k$ is competed and $E(\x, \bar{\y}^k) < E(\x, \bar{\y})  $}{
  
       $\bar{\y} \longleftarrow \bar{\y}^k$
  } 
 
  } 
 }
 \caption{Beam Search for Solving $\argmin_{\y}E_{\Theta}(\x, \y)$}\label{ag:bs}
\end{algorithm}
Where $\textit{succ}(\x, \bar{\y}^{k})$ is the set where additional token is added in $\bar{\y}^{k}$ and $\textit{TopK}(\cup_{k=1}^{K} \textit{succ}(\x, \bar{\y}^{k})$ are selected $K$ hypothesis with lowest energy. Beam size $K=1$ gives greedy decoding output. Figure~\ref{fig:bs} shows a beam search example with beam size 2\footnote{The Figure is from Stanford University lecture at \url{https://web.stanford.edu/class/cs224n/slides/cs224n-2021-lecture07-nmt.pdf}}.

The beam search algorithm is wildly used in machine translation~\citep{BahdanauCB14,googlenmt}. Researchers also find that considering length, coverage~\citep{googlenmt}, and an additional language model~\citep{2015arXiv150303535G} can lead to better decoding output in neural machine translation.

Although beam search algorithm can find fluent output, however, it often generally finds a sub-optimal solution of $\argmin_{\y} E(\x, \y)$. For linear-chain CRF, even if beam size is equal to label set size, the beam search algorithm is not guaranteed to find the optimal solution.

\paragraph{Gradient Descent}

Gradient from back-propagation is usually used to update neural network parameters. Several popular optimizers are used, such as stochastic gradient descent with momentum, Adagrad~\citep{adagrad}, RMSprop~\citep{rmsprop}, adam~\citep{adam}.  However gradient descent inference has been used in a variety of deep learning applications. Algorithm~\ref{ag:gd} which is used structure inference is shown below:

\hspace{1cm}

\begin{algorithm}[H]
\SetAlgoLined
 \KwIn{Given Energy Function:$E_{\Theta}$, Max Iteration Number $T_{max}$}
 \KwOut{$\y$}
 initialization $\y^{(0)}$, $t \leftarrow 0$ \;
 \While{$t < T_{max}$}{
  $\y \longleftarrow \y - \eta \frac{\partial E_\Theta(\x, \y}{\partial \y}$ \; 
  $t \longleftarrow t +1 $\; 
 }
 \caption{Gradient Descent for Solving $\argmin_{\y\in\yspace(\x)}E_{\Theta}(\x, \y)$}\label{ag:gd}
\end{algorithm}

\begin{figure}[h]
\centering
\includegraphics[width=0.8\textwidth]{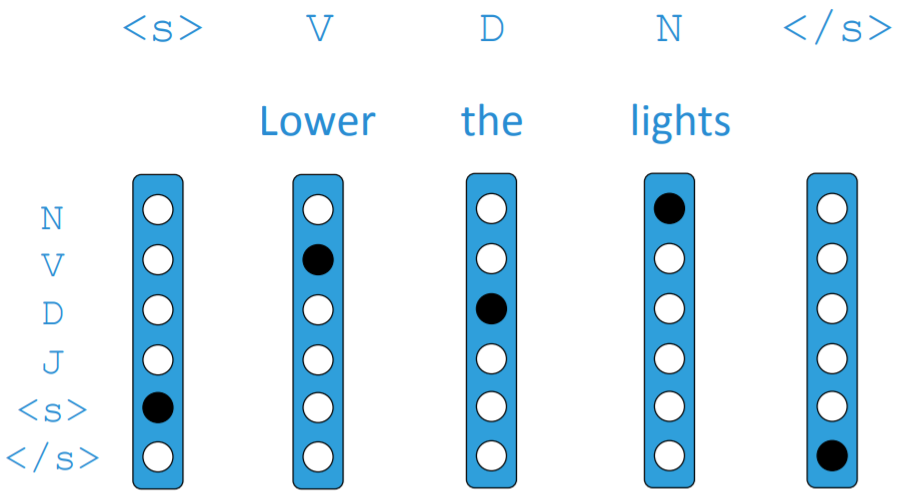}
\caption{Discrete structured output can be represented using	
one-hot	vectors. \label{fig:onehot} }
\end{figure}

\begin{figure}[h]
\centering
\includegraphics[width=0.8\textwidth]{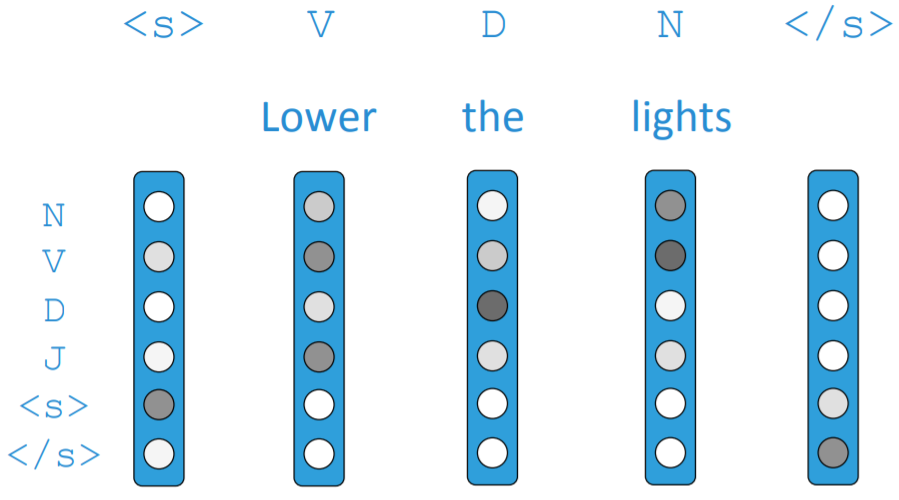}
\caption{In the relaxed continuous output space, each tag output can be treat as a distribution vector over tags\label{fig:relaxed}.}

\end{figure}

To use gradient descent (GD) for structured inference, researchers typically relax the output space from a discrete, combinatorial space to a continuous one and then use gradient descent to solve the following optimization problem: 
\begin{align}
\argmin_{\y\in\relyspace(\x)}E_\Theta(\x, \y) \nonumber 
\end{align}
where $\relyspace$ is the relaxed continuous output space. For sequence labeling, $\relyspace(\x)$ consists of length-$|\x|$ sequences of probability distributions over output labels. Figure~\ref{fig:onehot} and Figure~\ref{fig:relaxed} that are from the lecture~\citep{relaxed} shows the example how to relax discrete output space.
To obtain a discrete labeling for evaluation, the most probable label at each position is returned. 

\textbf{Gradient descent} is used for inference, e.g., image generation applications like DeepDream~\citep{deedream} and neural style transfer~\citep{DBLP:GatysEB15a}, structured prediction energy networks \citep{belanger2016structured}, as well as machine translation~\citep{DBLP:conf/emnlp/HoangHC17}.

\newpage

\mychapter{3}{Inference Networks}
\label{sec:infnet}

This chapter describes our contributions to approximate structure inference for structured tasks. It is computational challenging for structured inference with complex score functions. Previous work~\citep{belanger2016structured} relaxed $\y$ from a discrete to a continuous vector and used gradient descent for inference. 
We also relax $\y$  
but we use a different strategy to approximate inference. We demonstrate that our method achieves a better speed/accuracy/search error trade-off than gradient descent, while also being faster than exact inference at similar accuracy levels. We find further benefit by combining inference networks and gradient descent, using the former to provide a warm start for the latter.\footnote{Code is available at 
\texttt{github.com/lifu-tu/} \texttt{Benchmarking}\texttt{ApproximateInference}}

This chapter includes some material originally presented in ~\citet{tu-18,tu-gimpel-2019-benchmarking}.

\section{Inference Networks}

In chapter 2, we presented energy-based models, learning and inference difficulties of energy-based models. The complex energy functions with neural networks is commonly intractable~\citep{Cooper:1990:CCP:77754.77762}. There are generally two ways to address this difficulty. One is to restrict the model family to those for which inference is feasible. For example, state-of-the-art methods for sequence labeling use structured energies that decompose into label-pair potentials and then use rich neural network architectures to define the potentials~\citep[\emph{inter alia}]{journals/jmlr/CollobertWBKKK11,lample-etal-2016-neural}.
Exact dynamic programming algorithms like the Viterbi algorithm can be used for inference.   

The second approach is to retain computationally-intractable scoring functions but then use approximate methods for inference. 
For example, some researchers relax the structured output space from a discrete space to a continuous one and then use gradient descent to maximize the score function with respect to the output \citep{belanger2016structured}. 

We define an \textbf{inference network} $\infnet(\boldsymbol{x})$ (also called ``energy-based inference network'' in this thesis) parameterized by $\Psi$ 
and train it with the goal that
\begin{align}
\infnet(\x) \approx \argmin_{\y\in\relyspace(\x)}E_\Theta(\x, \y)\label{eq:infnet}
\end{align}

Given an energy function $E_\Theta$ and a dataset $X$ of inputs, we solve the following optimization problem:
\begin{align}
\hat{\Psi} \gets \argmin_{\Psi}\sum_{\x\in X}E_\Theta(\x, \infnet(\x))
\label{eq:finalnet}
\end{align}

\begin{figure}[h]
\centering
\includegraphics[width=0.95\textwidth]{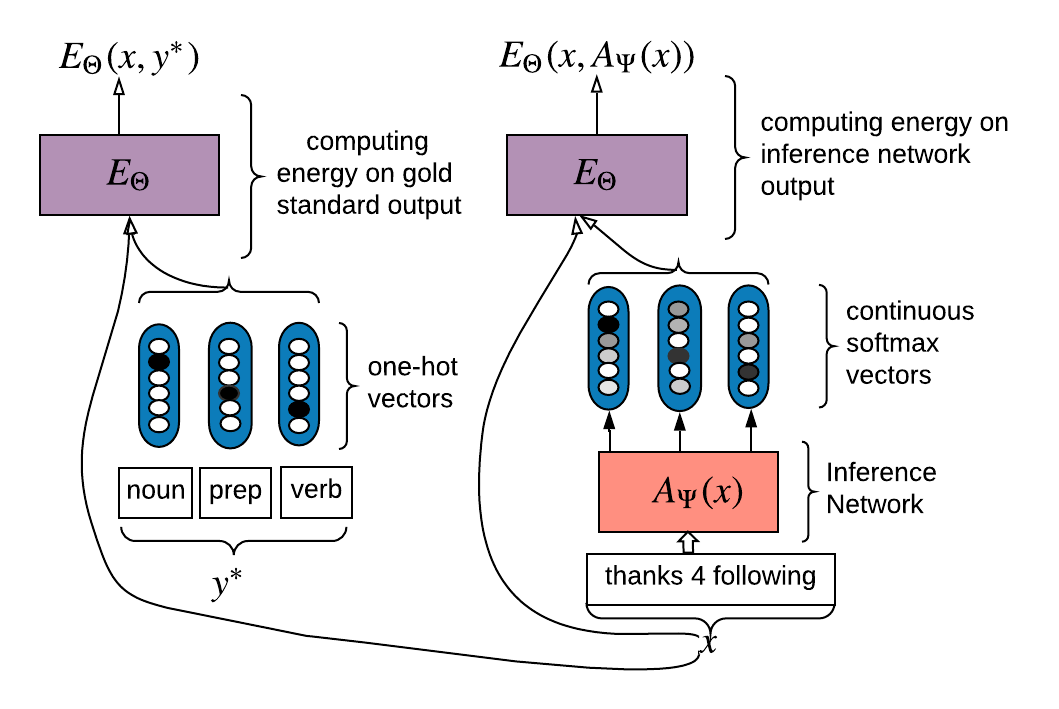}
\captionsetup{labelformat=empty}
\caption{
The architectures of inference network $\infnet$ and energy network $E_\Theta$. 
}
\end{figure}

The above Figure shows how to compute energy on the inference network output. The architecture of $\infnet$ will depend on the task. For Multiple Label Classification(MLC), the same set of labels is applicable to every input, so $\y$ has the same length for all inputs. So, we can use a feed-forward network for $\infnet$ with a vector output, treating each dimension 
as the prediction for a single label. 
For sequence labeling, each $\x$ (and therefore each $\y$) can have a different length, so we must use a network architecture for $\infnet$ that permits different lengths of predictions. We use an RNN that returns a vector at each position of $\x$. We interpret this vector as a probability distribution over output labels at that position.

\begin{figure}[h]
\centering
\includegraphics[width=0.9\textwidth]{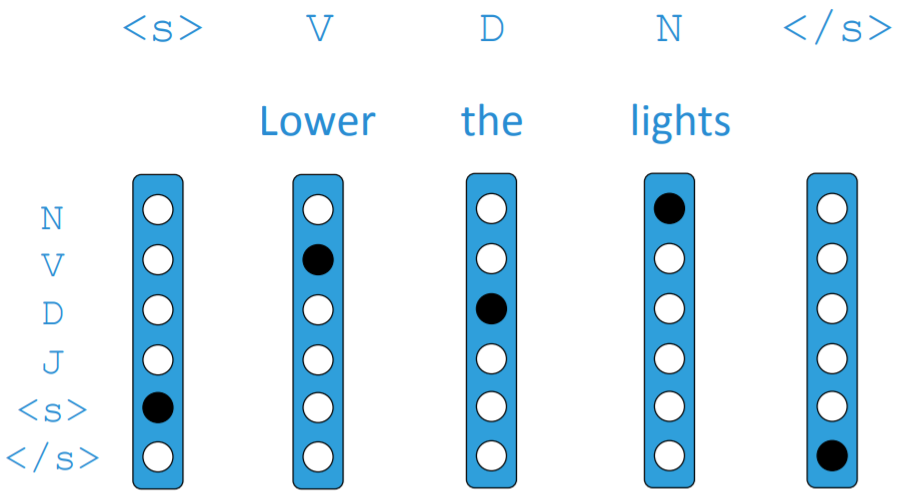}
\captionsetup{labelformat=empty}
\caption{Discrete structured output can be represented using	
one-hot	vectors.}
\end{figure}

\begin{figure}[h]
\centering
\includegraphics[width=0.9\textwidth]{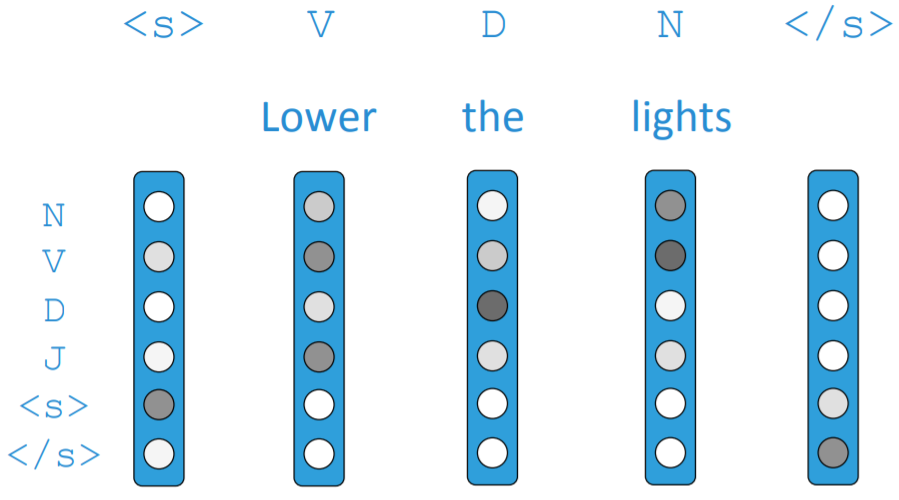}
\captionsetup{labelformat=empty}
\caption{In the relaxed continuous output space, each tag output can be treat as a distribution vector over tags.}
\end{figure}


We note that the output of $\infnet$ must be compatible with the energy function, which is typically defined in terms of the original discrete output space $\yspace$. This may require generalizing the energy function to be able to operate both on elements of $\yspace$ and $\relyspace$. The above figure show the example how to relax discrete output space so that inference network can be optimized with gradient methods.

\section{Improving Training for Inference Networks}

Below we describe several techniques we found to help stabilize training inference networks, which are optional terms added to the objective in Equation~\ref{eq:finalnet}.

\paragraph{$L_2$ Regularization:} 
We use $L_2$ regularization, adding the penalty term $\| \Psi \|_2^2$ with coefficient $\lambda_1$. It is a commonly used regularizer in deep neural network training.

\paragraph{Entropy Regularization:} 
We add an entropy-based regularizer $\mathrm{loss}_{\mathrm{H}}(\infnet(\x))$ defined for the problem under consideration. For MLC, the output of $\infnet(\x)$ is a vector of scalars in $[0,1]$, one for each label, where the scalar is interpreted as a label probability. The entropy regularizer $\mathrm{loss}_{\mathrm{H}}$ is the sum of the entropies over these label binary distributions. For sequence labeling, where the length of $\x$ is $N$ and where there are $L$ unique labels, the output of $\canet(\x)$ is a length-$N$ sequence of length-$L$ vectors, each of which represents the distribution over the $L$ labels at that position in $\x$. Then, $\mathrm{loss}_{\mathrm{H}}$ is the sum of entropies of these label distributions across positions in the sequence. 

When tuning the coefficient $\lambda_2$ for this regularizer, we consider both positive and negative values, permitting us to favor either low- or high-entropy distributions as the task prefers. For MLC, encouraging lower entropy distributions worked better, while for sequence labeling, higher entropy was better, similar to the effect found by \citet{DBLP:journals/corr/PereyraTCKH17}. Further research is required to gain understanding of the role of entropy regularization in such alternating optimization settings.

\paragraph{Local Cross Entropy Loss:} 
We add a local (non-structured) cross entropy $\mathrm{loss}_{\mathrm{CE}}(\infnet(\x_i), \y_i)$ defined for the problem under consideration. We have done the experiments with this loss for sequence labeling. 
It is the sum of the label cross entropy losses over all positions in the sequence. This loss provides more explicit feedback to the inference network, helping the optimization procedure to find a solution that minimizes the energy function while also correctly classifying individual labels. It can also be viewed as a multi-task loss for the inference network.

\paragraph{Regularization Toward Pretrained Inference Network:} 
We add the penalty $\| \Phi-\Phi_{0} \|_2^2$ 
where $\Phi_{0}$ is a pretrained network, e.g., a local classifier trained to independently predict each part of $\y$. 

\section{Connections with Previous Work}

\paragraph{Comparison to knowledge distillation:}~\citep{NIPS2014_5484,hinton2015distilling}, which refers to strategies in which one model (a ``student'') is trained to mimic another (a ``teacher''). Typically, the teacher is a larger, more accurate model but which is too computationally expensive to use at test time. \citet{urban2016deep} train shallow networks using image classification data labeled by an ensemble of deep teacher nets. \citet{lstms_to_cnns} train a convolutional network to mimic an LSTM for speech recognition. Others have explored knowledge distillation for sequence-to-sequence learning~\citep{DBLP:conf/emnlp/KimR16} and parsing~\citep{DBLP:conf/emnlp/KuncoroBKDS16}. It has been empirically observed that distillation can improve generalization, ~\citet{SelfDistillation} provides a theoretical analysis of distillation when the teacher and student architectures are identical. In our methods, there is no limitation for model size of ``student'' and ``teacher''.

\paragraph{Connection to amortized inference:} Since we train a single inference network for an entire dataset, our approach is also related to ``\textbf{amortized inference}''~\citep{Srikumar:2012:AIC:2390948.2391073,gershman2014amortized,paige2016inference,CUKR15}. Such methods precompute or save solutions to subproblems for faster overall computation. Our inference networks likely devote more modeling capacity to the most frequent substructures in the data. 
A kind of inference network is used in variational autoencoders~\citep{KingmaW13} to approximate posterior inference in generative models. 

Our methods are also related to work in structured prediction that seeks to approximate structured models with factorized ones, e.g., mean-field approximations in graphical models~\citep{Koller:2009:PGM:1795555,NIPS2011_4296}. 
Like our use of inference networks, there have been efforts in designing differentiable approximations of combinatorial search procedures~\citep{martins-kreutzer:2017:EMNLP2017,goyal18aaai} and structured losses for training with them~\citep{wiseman-rush-2016-sequence}. 
Since we relax discrete output variables to be continuous, there is also a connection to recent work that focuses on structured prediction with continuous valued output variables~\citep{NIPS2016_6074}. They also propose a formulation that yields an alternating optimization problem, but it is based on proximal methods.

\paragraph{Actor-Critic:} The actor-critic method is a popular reinforcement learning method, which trains a ``critic'' network to provide an estimation of value given the policy of an actor network. It avoids sampling from the policy's (actor's) action space, which can be expensive. The method have been applied to structure prediction~\citep{Bahdanau2017AnAA,Zhang2017ActorCriticST}. Comparing to the actor-critic method, In our work, the energy function behaves as a critic network, and the inference network is similar to an actor.

\paragraph{Gradient descent:} There are other settings in which \textbf{gradient descent} is used for inference, e.g., image generation applications like DeepDream~\citep{deedream} and neural style transfer~\citep{DBLP:GatysEB15a}, as well as machine translation~\citep{DBLP:conf/emnlp/HoangHC17}. In these and related settings, gradient descent has started to be replaced by inference networks, especially for image transformation tasks~\citep{Johnson2016Perceptual,DBLP:journals/corr/LiW16b}. Our results below provide more evidence for making this transition. An alternative to what we pursue here would be to obtain an easier convex optimization problem for inference via input convex neural networks~\citep{DBLP:journals/corr/AmosXK16}.

\section{General Energy Function}
\label{sec:benchmark}

The input space $\Xspace$ is now the set of all sequences of symbols drawn from a vocabulary. For an input sequence $\x$ of length $N$, where there are $L$ possible output labels for each position in $\x$, the output space $\yspace(\x)$ is $[L]^N$, where the notation $[q]$ represents the set containing the first $q$ positive integers. 
We define $\y = \langle y_{1},y_{2}, .., y_{N}\rangle$ where each $y_{i}$ ranges over possible output labels, i.e., $y_{i}\in [L]$. 

When defining our energy for sequence labeling, we take inspiration from bidirectional LSTMs (BLSTMs; \citealt{Hochreiter:1997:LSM:1246443.1246450}) and conditional random fields (CRFs; \citealt{Lafferty:2001:CRF}). 
A ``linear chain'' CRF uses two types of features: one capturing the connection between an output label and $\x$ and the other capturing the dependence between neighboring output labels. 
We use a BLSTM to compute feature representations for $\x$. We use $f(\x,t)\in\mathbb{R}^d$ to denote the ``input feature vector'' for position $t$, defining it to be the $d$-dimensional BLSTM hidden vector at $t$. 

The CRF energy function is the following:
\begin{align}
E_{\Theta}(\x, \y) = -\left(\sum_{t} U_{y_t}^\top f(\x,t)
+ \sum_{t}W_{y_{t-1},y_t}\right) 
\label{eq:crf-energy}
\end{align}
where $U_{i}\in\mathbb{R}^d$ is a parameter vector for label $i$  
and the parameter matrix $W\in\mathbb{R}^{L \times L}$ contains label pair parameters. 
The full set of parameters $\Theta$ includes 
the $U_i$ vectors,  
$W$, 
and the parameters of the BLSTM. 
The above energy only permits \textbf{discrete} $\y$. However, the general energy which permits \textbf{continuous} $\y$ is needed. Now, I will discuss the continuous version of the above energy.

For sequence labeling tasks, given an input sequence $\x = \langle x_1, x_2,..., x_{|\x|}\rangle$, we wish to output a sequence $\y = \langle \yv_1, \yv_2,..., \yv_{|\x|}\rangle \in \yspace(\x)$. Here $\yspace(\x)$ is the structured output space for $\x$. Each label $\yv_t$ is represented as an $L$-dimensional one-hot vector where $L$ is the number of labels. 

For the general case that permits relaxing $\y$ to be \textbf{continuous}, we treat each $y_t$ as a vector. It will be one-hot for the ground truth $\y$ and will be a vector of label probabilities for relaxed $\y$'s. Then the general energy function is:
\begin{align}
\label{eq:crf}
E_{\Theta}(\x, \y) = -\left(\sum_{t} \sum_{i=1}^L y_{t,i} \left(U_i^\top f(\x,t)\right) + \sum_{t}\y_{t-1}^\top W \y_{t}\right)
\end{align}
where $y_{t,i}$ is the $i$th entry of the vector $y_t$. In the discrete case, this entry is 1 for a single $i$ and 0 for all others, so this energy reduces to Eq.~\eqref{eq:crf-energy} in that case. In the continuous case, this scalar indicates the probability of the $t$th position being labeled with label $i$. 

For the label pair terms in this general energy function, we use a bilinear product between the vectors $\y_{t-1}$ and $\y_t$ using parameter matrix $W$, which also reduces to Eq.~\eqref{eq:crf-energy} when they are one-hot vectors.

\section{Experimental Setup}

In this section, we introduce how to apply our method on several tasks and compare with several other inference method: Viterbi, Gradient descent inference. We perform experiments on three tasks: Twitter part-of-speech tagging (POS)~\citep{gimpel-etal-2011-part,owoputi-etal-2013-improved} and, named entity recognition (NER)~\citep{tjong-kim-sang-de-meulder-2003-introduction}, and CCG supersense tagging (CCG)~\citep{hockenmaier-steedman-2002-acquiring}.

For our experimental comparison, we consider two CRF variants. The first is the basic model described above, which we refer to as \blstmcrf. We refer to the CRF with the following three techniques (word embedding fine-tuning, character-based embeddings, dropout) as \blstmcnncrf:
\paragraph{Word Embedding Fine-Tuning.} 
We used pretrained, fixed word embeddings when using the \blstmcrf model, but for the more complex \blstmcnncrf model, we fine-tune the pretrained word embeddings during training. 

\paragraph{Character-Based Embeddings.}
Character-based word embeddings provide consistent improvements in sequence labeling  \citep{lample-etal-2016-neural,ma-hovy-2016-end}. In addition to pretrained word embeddings, we produce a character-based embedding for each word using a character convolutional network like that of \citet{ma-hovy-2016-end}. The filter size is 3 characters and the character embedding dimensionality is 30. We use max pooling over the character sequence in the word and the resulting embedding is concatenated with the word embedding before being passed to the BLSTM. 

\paragraph{Dropout.}
We also add dropout during training~\citep{dropout}. 
Dropout is applied before the character embeddings are fed into the CNNs, at the final word embedding layer before the input to the BLSTM, and after the BLSTM. The dropout rate is 0.5 for all experiments.

\begin{figure}[h]

\centering
\includegraphics[width=1.0\textwidth]{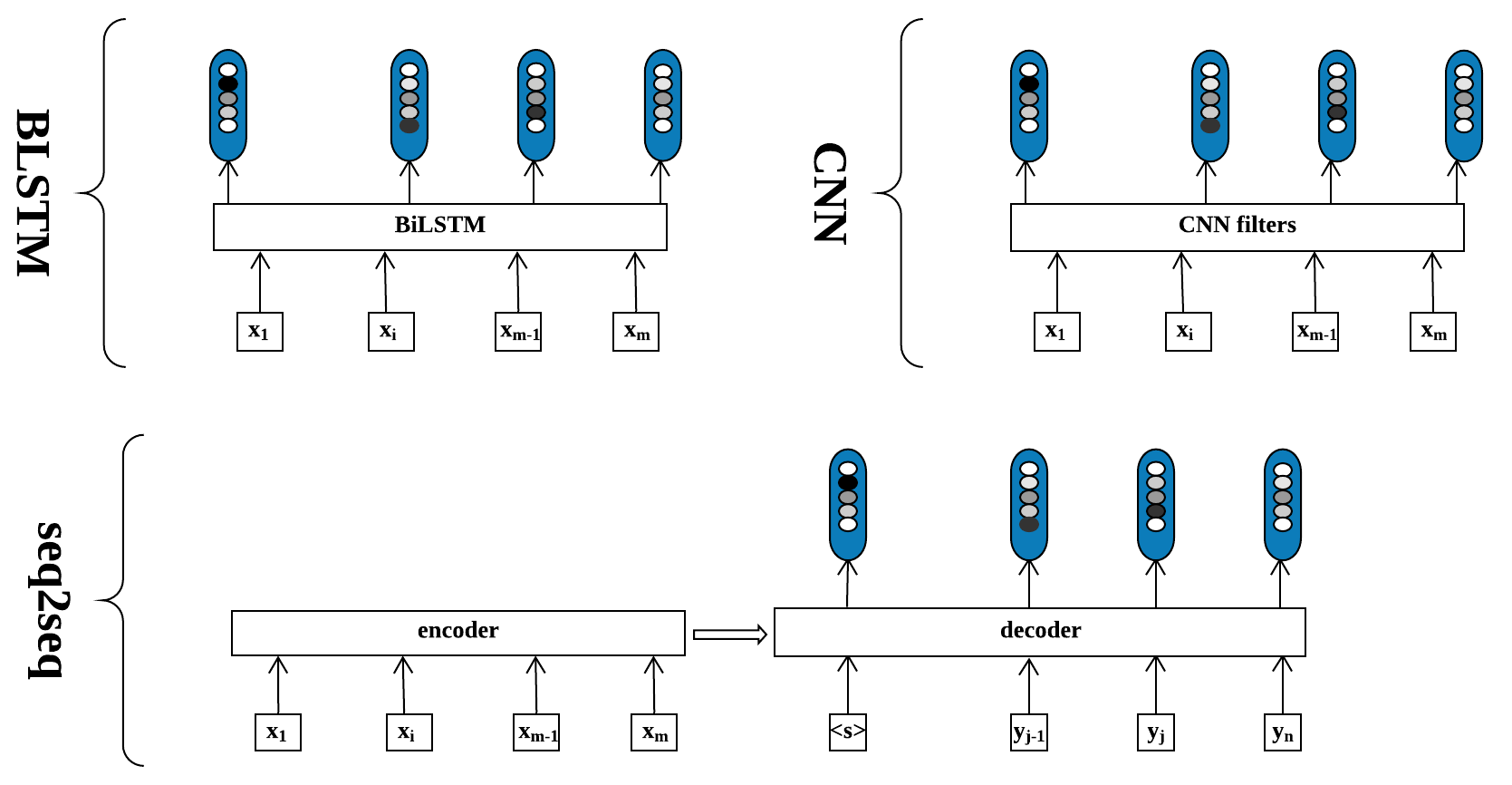}
\caption{Several inference network architectures.}
\label{fig:infArch}
\end{figure}

\paragraph{Inference Network Architectures.}
In our experiments, we use three options for the inference network architectures: convolutional neural networks (CNN), recurrent neural networks, sequence-to-sequence (seq2seq, \citealt{sutskever2014sequence}) models as shown in Figure~\ref{fig:infArch}. For seq2seq inference network, since sequence labeling tasks have equal input and output sequence lengths and a strong connection between corresponding entries in the sequences, \citet{goyal18aaai} used fixed attention that deterministically attends to the $i$th input when decoding the $i$th output, and hence does not learn any attention parameters.
For each, we optionally include the modeling improvements (word embedding fine-tuning, character-based embeddings, dropout) described in the above. 
When doing so, we append ``+'' to the setting's  name to indicate this (e.g., $\infnetLarge$).

\paragraph{Gradient Descent for Inference Details}
\label{sec:gd}
To use gradient descent (GD) for structured inference, we need to solve the following optimization problem: 

\begin{align}
\argmin_{\y\in\relyspace(\x)}E_\Theta(\x, \y) \nonumber  
\end{align}
where $\relyspace$ is the relaxed continuous output space. For sequence labeling, $\relyspace(\x)$ consists of length-$|\x|$ sequences of probability distributions over output labels. 
To obtain a discrete labeling for evaluation, the most probable label at each position is returned.

Gradient descent has the advantage of simplicity. Standard autodifferentiation toolkits can be used to compute gradients of the energy with respect to the output once the output space has been relaxed. 
However, one challenge is maintaining constraints on the variables being optimized. 

Therefore, we actually perform gradient descent in an relaxed output space $\relyspacer(\x)$ which consists of length-$|\x|$ sequences of 
vectors, where each vector $\yv_t\in\mathbb{R}^L$. 
When computing the energy, we use a softmax transformation on each $\yv_t$, solving the following optimization problem with gradient descent:

\begin{align}
\argmin_{{\y}\in\relyspacer(\x)}E_\Theta(\x, \softmax({\y})) 
\label{eq:gdinf}
\end{align}
where the softmax operation above is applied independently to each vector $\yv_t$ in the output structure $\y$.

For the number of epochs $N$, we consider values in the set $\{5, 10, 20, 30, 40, 50, 100, 500, 1000\}$. For each $N$, we tune the learning rate over the set  $\{1\mathrm{e}^4,5\mathrm{e}^3,1\mathrm{e}^3, 500, 100, 50, 10, 5, 1\}$). These learning rates may appear extremely large when we are accustomed to choosing rates for empirical risk minimization, but we generally found that the most effective learning rates for structured inference are orders of magnitude larger than those effective for learning. 
To provide as strong performance as possible for the gradient descent method, we tune $N$ and the learning rate via oracle tuning, i.e., we choose them separately for each input to maximize performance (accuracy or F1 score) on that input.

\section{Training Objective}

For training the inference network parameters $\Psi$, we find that a local cross entropy loss consistently worked well for sequence labeling. We use this local cross entropy loss in this proposal, so we perform learning by solving the following:
\begin{align}
\argmin_{\Psi}\!\!\sum_{\langle\x,\y\rangle} \!\! E_\Theta(\x, \infnet(\x))\! +\! \lambda\ltok(\y, \infnet(\x))\nonumber
\end{align}
\noindent where 
the sum is over $\langle \x, \y\rangle$ pairs in the training set.
The token-level loss is defined:  
\begin{align}
\ltok(\y, \infnetnoparams(\x)) = \sum_{t=1}^{|\y|} \ce (\yv_t, \infnetnoparams(\x)_t)  
\label{eq:token_loss}
\end{align}
where $\yv_t$ is the $L$-dimensional one-hot label vector at position $t$ in $\y$, $\infnetnoparams(\x)_t$ is the inference network's output distribution at position $t$,
and $\ce$ stands for cross entropy. $\ltok$ is the loss used in our non-structured baseline models.


\section{\blstmcrf Results.}

Table~\ref{table:loss} shows shows test results for all tasks and architectures. The results use the simpler \blstmcrf modeling configuration: word embedding are fixed, no character embeddings and no dropout technique during training. The inference networks use the same architectures as the corresponding local baselines, but their parameters are trained with both the local loss and the \blstmcrf energy, leading to consistent improvements. 
CNN inference networks work well for POS, but struggle on NER and CCG compared to other architectures. 
BLSTMs work well, but are outperformed slightly by seq2seq models across all three tasks. 
Using the Viterbi algorithm for exact inference yields the best performance for NER but is not best for the other two tasks. 

It may be surprising that an inference network trained to mimic Viterbi would outperform Viterbi in terms of accuracy, which we find for the CNN for POS tagging and the seq2seq inference network for CCG. We suspect this occurs for two reasons. One is due to the addition of the local loss in the inference network objective; the inference networks may be benefiting from this multi-task training. 
\citet{edunov-etal-2018-classical} similarly found benefit from a combination of token-level and sequence-level losses. 
The other potential reason is beneficial inductive bias with the inference network architecture. For POS tagging, the CNN architecture is clearly well-suited to this task given the strong performance of the local CNN baseline. Nonetheless, the CNN inference network is able to improve upon both the CNN baseline and Viterbi.   

\begin{table*}[h]
\centering
\begin{tabular}{l|c|c|c||c|c|c||c|c|c|}
\cline{2-10}
& \multicolumn{3}{c||}{Twitter POS Tagging}
& \multicolumn{3}{c||}{NER}
& \multicolumn{3}{c|}{CCG Supertagging}
\\ \cline{2-10} 
& CNN & BLSTM & seq2seq & CNN & BLSTM & seq2seq &CNN & BLSTM & seq2seq \\\hline
\multicolumn{1}{|l|}{local baseline} & 89.6 & 88.0 & 88.9 & 79.9 & 85.0 & 85.3 & 90.6& 92.2  &  92.7\\
\hline
\multicolumn{1}{|l|}{$\infnetBase$} & \bf 89.9 & 89.5  & 89.7   & 82.2 & 85.4  & 86.1  & 91.3 & 92.8  & \bf 92.9  \\
\hline
\multicolumn{1}{|l|}{gradient descent} & \multicolumn{3}{c||}{89.1} & \multicolumn{3}{c||}{84.4} & \multicolumn{3}{c|}{89.0} \\
\hline
\multicolumn{1}{|l|}{Viterbi } & \multicolumn{3}{c||}{89.2} & \multicolumn{3}{c||}{\bf 87.2} & \multicolumn{3}{c|}{92.4} \\
\hline
\end{tabular}
\caption{Test results for all tasks. Inference networks, gradient descent, and Viterbi are all optimizing the \blstmcrf energy. Best result per task is in bold.
}
\label{table:loss}
\end{table*}

\begin{figure}[t]
  \centering
  \begin{subfigure}[b]{0.3\linewidth}
    \includegraphics[width=\linewidth]{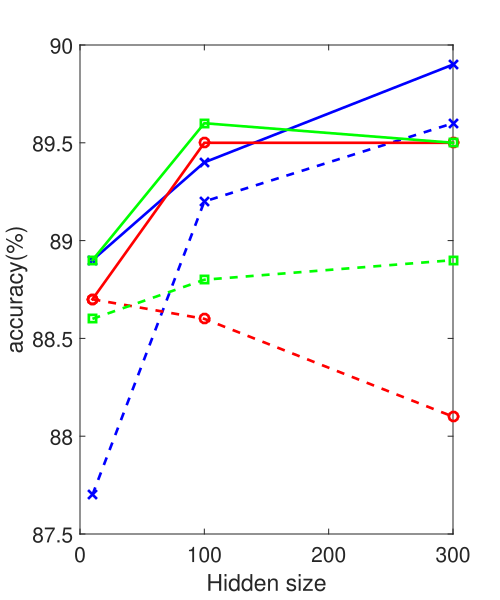}
    \caption{POS}
  \end{subfigure}
  \begin{subfigure}[b]{0.3\linewidth}
    \includegraphics[width=\linewidth]{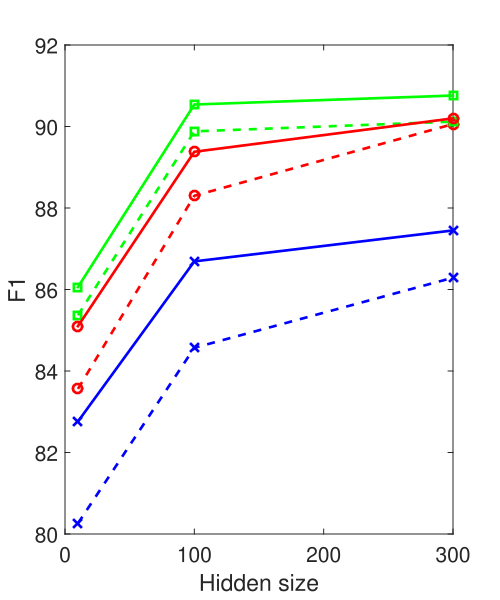}
    \caption{NER}
  \end{subfigure}
  \begin{subfigure}[b]{0.3\linewidth}
    \includegraphics[width=\linewidth]{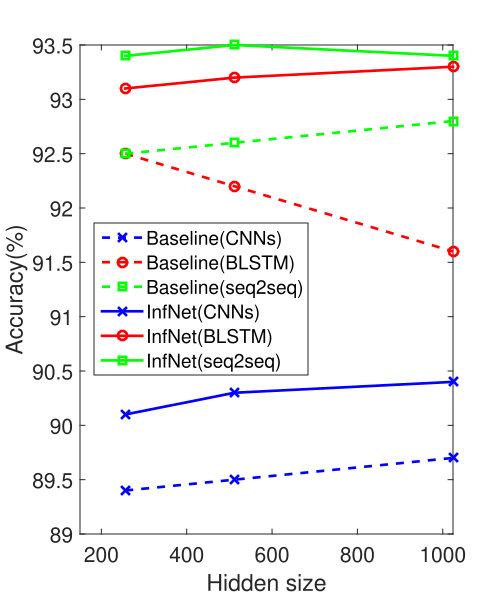}
    \caption{CCG Supertagging}
  \end{subfigure}
  \caption{Development results for inference networks with different architectures and hidden sizes ($H$).  
  }
  \label{fig:hiddensize}
\end{figure}

\paragraph{Hidden Size.} For the test results in Table~\ref{table:loss}, we did limited tuning of $H$ for the inference networks based on the development sets. Figure~\ref{fig:hiddensize} shows the impact of $H$ on performance. Across $H$ values, the inference networks outperform the baselines. For NER and CCG, seq2seq outperforms the BLSTM which in turn outperforms the CNN.

\paragraph{Tasks and Window Size.} 
Table~\ref{table:cnnresults} shows that CNNs with smaller windows are better for POS, while larger windows are better for NER and CCG. This suggests that POS has more local dependencies among labels than NER and CCG. 
\begin{table}[h]
\centering
\large
\begin{tabular}{|ll|c|c|}
\cline{3-4}
\multicolumn{2}{l|}{} & $\{$1,3$\}$-gram & $\{$1,5$\}$-gram\\
\hline
\multirow{2}{*}{POS} & local baseline & 89.2&  88.7\\ 
 & $\infnetBase$ & \bf 89.6 & 89.0\\ 
\hline
\multirow{2}{*}{NER} & local baseline &84.6   &  85.4\\ 
 & $\infnetBase$  & 86.7  & \bf 86.8\\
\hline
\multirow{2}{*}{CCG} & local baseline & 89.5  & 90.4\\
 & $\infnetBase$  &  90.3 & \bf 91.4\\
\hline
\end{tabular}
\caption{Development results for CNNs with two filter sets ($H=100$).}
\label{table:cnnresults}
\end{table}

\paragraph{Speed Comparison}
Asymptotically, Viterbi takes $\mathcal{O}(nL^2)$ time, where 
$n$ is the sequence length. The BLSTM and our deterministic-attention seq2seq models have time complexity $\mathcal{O}(nL)$. 
CNNs 
also have complexity $\mathcal{O}(nL)$ but are more easily parallelizable. 
Table~\ref{table:speed} shows test-time inference speeds for inference networks, gradient descent, and Viterbi for the \blstmcrf model. We use GPUs and a minibatch size of 10 for all methods. 
CNNs are 1-2 orders of magnitude faster than the others. BLSTMs work almost as well as seq2seq models and are 2-4 times faster in our experiments. 
Viterbi is actually faster than seq2seq when $L$ is small, but for CCG, which has $L=400$, it is 4-5 times slower. Gradient descent is slower than the others because it generally needs many iterations (20-50) for competitive performance.

\begin{table}[h]
\setlength{\tabcolsep}{4pt}
\centering
\large
\begin{tabular}{l|c|c|c|c|c|}
\cline{2-4}
& \multicolumn{3}{c|}{Inference Networks}   
\\ \cline{2-6}
\multicolumn{1}{l|}{} & CNN & BLSTM  & seq2seq & Viterbi  & Gradient Descent
\\ \hline
POS & 12500 & 1250  & 357   & 500 & 20 \\  

NER &  10000 & 1000  & 294 & 360 & 23\\

CCG & 6666 & 1923 &  1000 & 232 & 16 \\

\hline
\end{tabular}
\caption{Speed comparison of inference networks across tasks and architectures (examples/sec). 
}
\label{table:speed}
\end{table}

\paragraph{Search Error} 
We can view inference networks as approximate search algorithms and assess characteristics that affect  search error.
To do so, we train two  
LSTM language models (one on word sequences and one on gold label sequences) on the Twitter POS data. 

We compute the difference in the \blstmcrf energies between the inference network output $\y_{\info}$ and the Viterbi output $\y_{\vito}$ as the search error:
\begin{align}
E_{\Theta}(\x, \y_{\info}) - E_{\Theta}(\x, \y_{\vito})   
\end{align}

We compute the same search error for gradient descent.
For the BLSTM inference network, Spearman's $\rho$ between the word sequence perplexity and search error is 0.282; for the label sequence perplexity, it is 0.195.
For gradient descent inference, Spearman's $\rho$ between the word sequence perplexity and search error is 0.122; for the label sequence perplexity, it is 0.064.
These positive correlations mean that for frequent sequences, inference networks and gradient descent exhibit less search error. We also note that the correlations are higher for the inference network than for gradient descent, showing the impact of amortization during learning of the inference network parameters. That is, since we are learning to do inference from a dataset, we would expect search error to be smaller for more frequent sequences, and we do indeed see this correlation.

\section{\blstmcnncrf Results}
We now compare inference methods when using the improved modeling techniques: word embedding fine-tuning, character-based embeddings and dropout. We use these improved techniques for all models, including the CRF, the local baselines, gradient descent, and the inference networks.

The results are shown in Table~\ref{table:advanceloss}. 
With a more powerful local architecture, structured prediction is less helpful overall, but inference networks still improve over the local baselines on 2 of 3 tasks.

\begin{table}[h]
\centering
\large
\begin{tabular}{|l|c|c|c|}
\cline{2-4}
\multicolumn{1}{l|}{} &POS
& NER
& CCG \\
\hline
local baseline & 91.3 & 90.5 & 94.1 \\ 
$\infnetLarge$  & 91.3 & 90.8 & 94.2 \\
gradient descent & 90.8 & 89.8 & 90.4 \\
Viterbi & 90.9 & 91.6 & 94.3 \\ \hline
\end{tabular}
\caption{Test results 
with \blstmcnncrf. For local baseline and inference network architectures, we use 
CNN for POS, seq2seq for NER, and BLSTM for CCG.
}
\label{table:advanceloss}
\end{table}

\paragraph{POS.}
As in the \blstmcrf setting, the local CNN baseline and the CNN inference network outperform Viterbi. This is likely because the CRFs use BLSTMs as feature networks, but our results show that CNN baselines are consistently better than BLSTM baselines on this task.
As in the \blstmcrf setting, gradient descent works quite well on this task, comparable to Viterbi, though it is still much slower. 

\paragraph{NER.} 
We see slightly higher \blstmcnncrf results than several previous state-of-the-art results (cf.~90.94;~\citealp{lample-etal-2016-neural} and 91.37;~\citealp{ma-hovy-2016-end}). 
The stronger \blstmcnncrf configuration also helps the inference networks, 
improving performance from 90.5 to 90.8 for the seq2seq architecture over the local baseline. 
Though gradient descent reached high accuracies for POS tagging, it does not perform well on NER, possibly due to the greater amount of non-local information in the task. 

While we see strong performance with \infnetLarge, it still lags behind Viterbi in F1. 
We consider additional experiments in which we increase the number of layers in the inference networks. We use a 2-layer BLSTM as the inference network and also use weight annealing of the local loss hyperparameter $\lambda$, setting it to $\lambda=e^{-0.01t}$ where $t$ is the epoch number. Without this annealing, the 2-layer inference network was difficult to train. 

The weight annealing was helpful for encouraging the inference network to focus more on the non-local information in the energy function rather than the token-level loss. 
As shown in Table~\ref{table:morelayer_ner_result}, these changes yield an improvement of 0.4 in F1. 

\begin{table}[h]
\centering
\large
\begin{tabular}{|l|c|}
\cline{2-2}
\multicolumn{1}{l|}{}  & F1 
\\ \hline
local baseline (BLSTM)  & 90.3 \\ 
$\infnetLarge$ (1-layer BLSTM) & 90.7 \\
$\infnetLarge$ (2-layer BLSTM) & 91.1 \\
Viterbi &  91.6 \\
\hline
\end{tabular}
\caption{NER test results (for \blstmcnncrf)  with more layers 
in the 
BLSTM inference network.} 
\label{table:morelayer_ner_result}
\end{table}

\paragraph{CCG.} 
Our \blstmcnncrf reaches an accuracy of 94.3\%, which is comparable to several recent results (93.53, \citealp{xu-etal-2016-expected}; 94.3, \citealp{lewis-etal-2016-lstm}; and 94.50,  \citealp{vaswani-etal-2016-supertagging}). 
The local baseline, the BLSTM inference network, and Viterbi are all extremely close in accuracy. 
Gradient descent struggles here, likely due to the large number of candidate output labels. 

\subsection{Methods to Improve Inference Networks} 


To further improve the performance of an inference network for a particular
test instance $\x$, we propose two novel approaches that leverage the strengths of inference networks to provide effective starting points and then use instance-level fine-tuning in two different ways. 

\paragraph{Instance-Tailored Inference Networks}

For each test example $\x$, we initialize an instance-specific inference network $\infnet(\x)$ using the trained inference network parameters, then run gradient descent on the following loss:
\begin{align}
\argmin_{\Psi} E_\Theta(\x, \infnet(\x))
\label{eq:finetuneloss}
\end{align}
This procedure fine-tunes the inference network parameters for a single test example to minimize the energy of its output. For each test example, the process is repeated, with a new instance-specific inference network being initialized from the trained inference network parameters. 

\paragraph{Warm-Starting Gradient Descent with Inference Networks}
Given a test example $\x$, we initialize ${\y}\in\relyspacer(\x)$ using the inference network and then use gradient descent by solving Eq.~\ref{eq:gdinf} described in Section~\ref{sec:gd} to update $\y$. 
However, the inference network output is in $\relyspace(\x)$ while gradient descent works with the more relaxed space $\relyspacer(\x)$. So we simply use the logits from the inference network, which are the score vectors before the softmax operations.

\subsection{Speed, Accuracy, and Search Error}
\begin{table*}[t]
\centering
\begin{tabular}{lc|c|c||c|c||c|c|}
\cline{3-8}
&& \multicolumn{2}{c||}{Twitter POS Tagging}
& \multicolumn{2}{c||}{NER}
& \multicolumn{2}{c|}{CCG Supertagging}
\\ \cline{3-8} 
& $N$ & Acc. $(\uparrow)$ & Energy $(\downarrow)$   & F1 $(\uparrow)$ & Energy $(\downarrow)$ & Acc. $(\uparrow)$ & Energy  $(\downarrow)$ \\
\hline
\multicolumn{1}{|l}{gold standard} &  & 100 & -159.65 &100    & -230.63 &  100 &  -480.07  \\
\hline \hline
\multicolumn{1}{|l}{\blstmcnncrf/Viterbi} &  & 90.9  & -163.20 & 91.6   & -231.53 & 94.3  & -483.09   \\
\hline \hline
\multicolumn{1}{|l}{} & 10 & 89.2 & -161.69 &81.9 & -227.92 & 65.1 &-412.81  \\
\multicolumn{1}{|l}{} & 20 &90.8  & -163.06 &  89.1  & -231.17 &  74.6 & -414.81   \\
\multicolumn{1}{|l}{} & 30 & 90.8 & -163.02 & 89.6 & -231.30 & 83.0 & -447.64\\
\multicolumn{1}{|l}{\multirow{2}{*}{gradient descent}} & 40 & 90.7 & -163.03 &  89.8  & -231.34 & 88.6  &-471.52    \\
\multicolumn{1}{|l}{} & 50 & 90.8 & -163.04 &  89.8  & -231.35 &  90.0 &-476.56    \\
\multicolumn{1}{|l}{} & 100 & - & - &  -  & - &  90.1 &-476.98    \\
\multicolumn{1}{|l}{} & 500 & - & - & -  & - &  90.1 &-476.99    \\
\multicolumn{1}{|l}{} & 1000 &- & - &  -  & - &  90.1 &-476.99    \\
\hline \hline
\multicolumn{2}{|l|}{$\infnetLarge$} & 91.3 & -162.07 &  90.8  & -231.19 & 94.2 & -481.32    \\
\multicolumn{2}{|l|}{discretized output from $\infnetLarge$} & 91.3 & -160.87 & 90.8   & -231.34 & 94.2  & -481.95    \\
\hline \hline
\multicolumn{1}{|l}{} & 3 &91.0  &-162.59  &91.3  &-231.32  & 94.3 & -481.91    \\
\multicolumn{1}{|l}{instance-tailored $\infnetLarge$} & 5 & 90.9 & -162.81 & 91.2 & -231.37 & 94.3 &  -482.23   \\
\multicolumn{1}{|l}{} & 10 & 91.3 & -162.85 & 91.5 & -231.39 & 94.3 & -482.56    \\
\hline \hline
\multicolumn{1}{|l}{\multirow{2}{*}{$\infnetLarge$ as warm start for}} & 3 &91.4  &-163.06  & 91.4 & -231.42 & 94.4 &-482.62     \\
\multicolumn{1}{|l}{\multirow{2}{*}{gradient descent}} & 5 & 91.2 & -163.12  & 91.4 & -231.45 & 94.4 & -482.64    \\
\multicolumn{1}{|l}{} & 10 &91.2  & -163.15 & 91.5 & -231.46 &  94.4& -482.78    \\
\hline
\end{tabular}
\caption{Test set results of approximate inference methods for three tasks, showing performance metrics (accuracy and F1) as well as average energy of the output of each method.
The inference network architectures in the above experiments are: CNN for POS, seq2seq for NER, and BLSTM for CCG. $N$ is the number of epochs for GD inference or instance-tailored fine-tuning. 
}
\label{table:sgd}
\end{table*}

\begin{figure}[h]
  \centering
\includegraphics[width=0.7\linewidth]{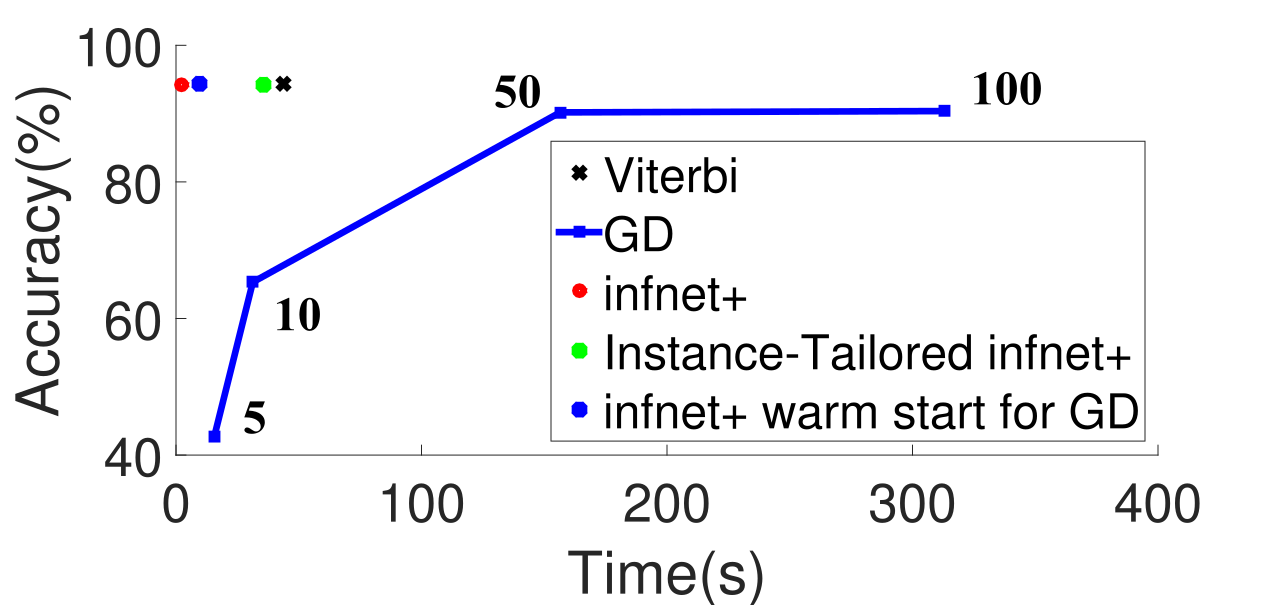}
  \caption{Speed and accuracy comparisons of three difference inference methods: Viterbi, gradient descent and inference network.
  }
  \label{fig:sgd1}
\end{figure}

\begin{figure}[h]
  \centering
\includegraphics[width=0.7\linewidth]{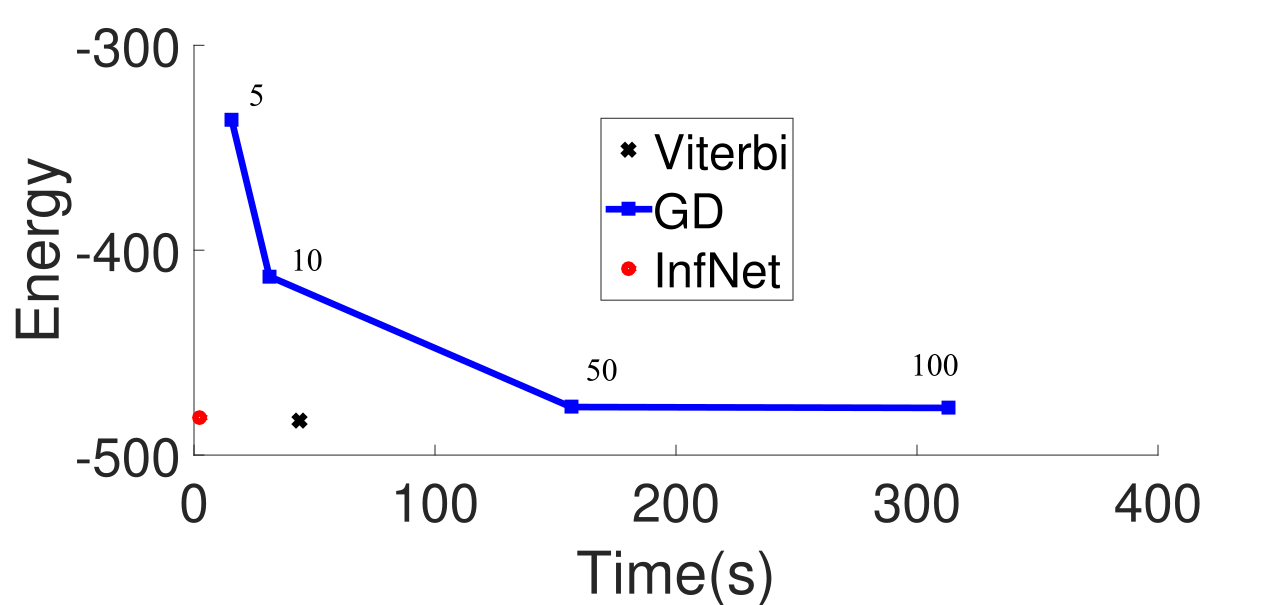}
  \caption{Speed and search error comparisons of three difference inference methods: Viterbi, gradient descent and inference network.
  }
  \label{fig:sgd2}
\end{figure}

Table~\ref{table:sgd}, Figure~\ref{fig:sgd1} and Figure~\ref{fig:sgd2} compares inference methods in terms of both accuracy and energies reached during inference. 
For each number $N$ of gradient descent iterations in the table, we tune the learning rate per-sentence and report the average accuracy/F1 with that fixed number of iterations. We also report the average energy reached. 
For inference networks, we report energies both for the output directly and when we discretize the output (i.e., choose the most probable label at each position).

\paragraph{Gradient Descent Across Tasks.} 
The number of gradient descent iterations required for competitive performance varies by task. For POS, 20 iterations are sufficient to reach accuracy and energy close to Viterbi. 
For NER, roughly 40 iterations are needed for gradient descent to reach its highest F1 score, and for its energy to become very close to that of the Viterbi outputs. However, its F1 score is much lower than Viterbi. 
For CCG, gradient descent requires far more iterations, presumably due to the larger number of labels in the task. Even with 1000 iterations, the accuracy is 4\% lower than Viterbi and the inference networks. Unlike POS and NER, the inference network reaches much lower energies than gradient descent on CCG, suggesting that the inference network may not suffer from the same challenges of searching high-dimensional label spaces as those faced by gradient descent.

\paragraph{Inference Networks Across Tasks.}
For POS, the inference network does not have lower energy than gradient descent with $\geq 20$ iterations, but it does have higher accuracy. 
This may be due in part to our use of multi-task learning for inference networks. 
The discretization of the inference network outputs increases the energy on average for this task, whereas it decreases the energy for the other two tasks. 
For NER, the inference network reaches a similar energy as gradient descent, especially when discretizing the output, but is considerably better in F1. 
The CCG tasks shows the largest difference between gradient descent and the inference network, as the latter is much better in both accuracy and energy.

\paragraph{Instance Tailoring and Warm Starting.}
Across tasks, instance tailoring and warm starting lead to lower energies than $\infnetLarge$. The improvements in energy are sometimes joined by improvements in accuracy, notably for NER where the gains range from 0.4 to 0.7 in F1. 
Warm starting gradient descent yields the lowest energies (other than Viterbi), showing promise for the use of gradient descent as a local search method starting from inference network output.

\paragraph{Wall Clock Time Comparison.} 

\begin{figure}[h]
  \centering
\includegraphics[width=0.7\linewidth]{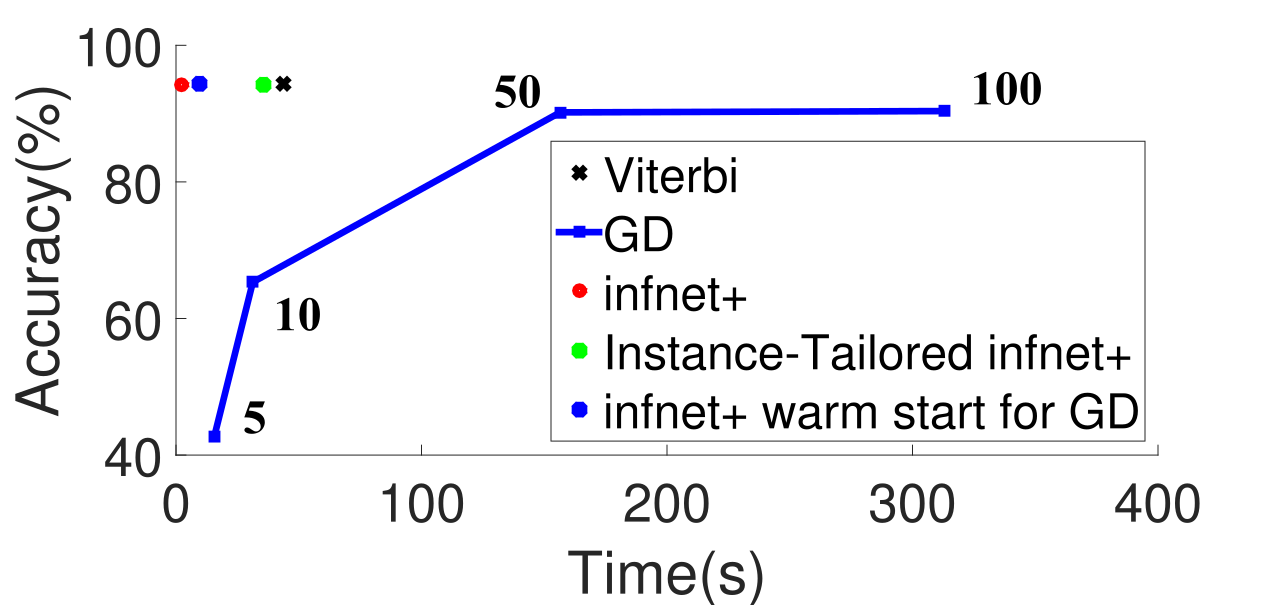}
  \caption{CCG test results for inference methods (GD = gradient descent). The x-axis is the total inference time for the test set. The numbers on the GD curve are the number of gradient descent iterations. 
  }
  \label{fig:inference_method}
\end{figure}
Figure \ref{fig:inference_method} shows the speed/accuracy trade-off for the inference methods, using wall clock time for test set inference as the speed metric. 
On this task, Viterbi is time-consuming because of the larger label set size. The inference network has comparable accuracy to Viterbi but is much faster. Gradient descent needs much more time to get close to the others but plateaus before actually reaching similar accuracy. Instance-tailoring and warm starting reside between $\infnetLarge$ and Viterbi, with warm starting being significantly faster because it does not require updating inference network parameters.

\section{Conclusion}
We compared several methods for approximate inference in neural structured prediction, finding that inference networks achieve a better speed/accuracy/search error trade-off than gradient descent. 
We also proposed instance-level inference network fine-tuning and using inference networks to initialize gradient descent, finding further reductions in search error and improvements in performance metrics for certain tasks.
\newpage

\mychapter{4}{Energy-Based Inference Networks for Non-Autoregressive Machine Translation}
\label{sec:ENGINE}

In this chapter, We use the proposed structure inference method in chapter 3 for non-autoregressive machine model. we propose to train a non-autoregressive machine translation model to minimize the energy defined by a pretrained autoregressive model. In particular, we view our non-autoregressive translation system as an inference network trained to minimize the autoregressive 
teacher energy. 

This chapter includes some material originally presented in ~\citet{tu-etal-2020-engine}. Code is available at \url{https://github.com/lifu-tu/ENGINE}.

\section{Background}

\subsection{Autoregressive Machine Translation}
A neural machine translation system is a neural network that directly models the conditional probability $p(\y \mid \x)$ of translating a source sequence $\x = \langle x_1, x_2,..., x_{|x|}\rangle$ to a target sequence $\y = \langle y_1, y_2,..., y_{|y|}\rangle$. $y_{|y|}$ is a special end of sentence token $<eos>$. The Seq2Seq framework relies on the encoder-decoder paradigm. The encoder encodes the input sequence, while the decoder produces the target sequence.
Then the conditional probability can be decomposed as the follows:

\begin{equation}
\log p(\y \mid \x)=\sum_i \log p(y_j \mid y_{<j},s) \nonumber
\end{equation}
Here $s$ is the source sequence representation which is computed by the encoder.
\\\\
\textbf{Encoder}
The encoder reads the source sequence $\langle x_1, x_2,..., x_{|x|}\rangle$ into another sequence $\langle s_1, s_2,..., s_{|x|}\rangle$ . The encoder could be realized such as a recurrent neural network such that

\begin{equation}
s_{i+1}=f_e(x_{i+1}, s_i) \nonumber
\end{equation}
where $s_{i} \in \mathbb{R}^d$ is the hidden state at time $i$, $f_e$ is a nonlinear function.
\\\\
\textbf{Decoder}
The decoder is trained to predict the next word $y_{j+1}$ given the encoder output $s$ and all the previously predicted words. The probability is parameter as the following
\begin{equation}
 p(y_j \mid y_{<j},s)=\softmax(g(h_j)) \nonumber
\end{equation}
with $g$ is a transformation function that outputs a vocabulary-sized vector. Here $h_j$ is also the RNN hidden unit of the decoder at time step $j$, which is could computed as:

\begin{equation}
h_j=f_d(h_{j-1},y_{j-1}) 
\end{equation}
Here $f_d$ is also a nonlinear function.

\noindent For the attention-based models, the attention hidden state $\bar{h}$ is difference from the above. It is computed as follows:
\begin{equation}\label{eq:dehidden}
\bar{h}_j=f(W_c [h_j,c_j]) \nonumber
\end{equation}
Here $c_j$ is the source-side context vector.

\begin{equation}\label{eq:deoutput}
p(y_j \mid y_{<j},s)=\softmax(W_d \bar{h}_j) \nonumber
\end{equation}

\noindent The aligned vector $a_t$, whose size is equal the number of time step on the source sentence.
\begin{equation}
a_j(i)=\frac{\exp(\score(h_j, \bar{s_i}))}{\sum_{i'}\exp(\score(h_j, \bar{s_{i'}}))} \nonumber
\end{equation}

There are three different alternatives to compute the score. In my experiment, I just use the following general one:
\begin{equation}
\score(h_j, s_i) = h_j^{T}W_a s_i \nonumber
\end{equation}

The context vector $c_j$ is then computed as the weighted sum of the source hidden vector $h_i$ as follow:
\begin{equation}
c_j= \sum_{i=1}^{|x|}a_j(i)s_i \nonumber
\end{equation}

The computation graph is simple, we can go from $h_j \to a_j(i) \to c_j \to \bar{h}_j \to y_j$, which follow ~\citet{luong-etal-2015-effective} 's step. ~\citet{BahdanauCB14} at each time $j$, go from $h_{j-1} \to a_j(i) \to c_j \to \bar{h}_j \to y_j$.

\subsection{Non-autoregresive Machine Translation System}
In the work~\citep{gu2018non}, they introduce non-autoregressive neural machine translation (NAT) systems based on transformer network~\citep{NIPS2017_7181} in order to remove the autoregressive connection and do parallel decoding. The naive solution is to have the following assumption:

\begin{align*}
    \log p_{\theta} (\y \mid \x) = \sum_{t=1}^{|\y|} \log p_{\theta}(y_{t} \mid \x ) 
\end{align*}
The target token is independent given the input. Unfortunately, the performance of non-autoregressive models fall far behind autoregressive models.

\begin{figure}
\centering
\includegraphics[width=0.9\textwidth]{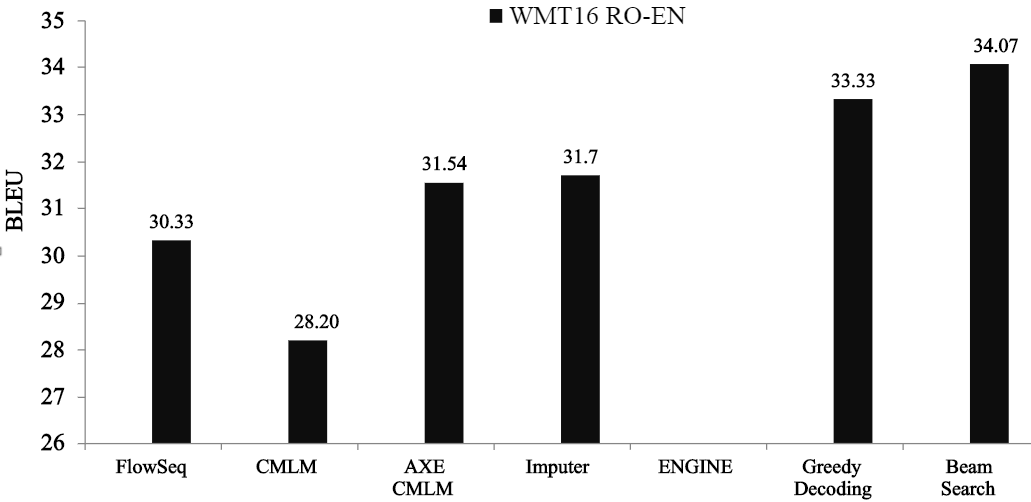}
\caption{The performance of autogressive models and non-autoregressive models on WMT16 RO-EN dataset.}
\label{fig:autoVsNonauto}
\end{figure}

The performance of non-autoregressive neural machine translation (NAT) systems, which predict tokens in the target language independently of each other conditioned on the source sentence, has been improving steadily in recent  years~\citep{lee-etal-2018-deterministic,ghazvininejad-etal-2019-mask,ma-etal-2019-flowseq}. The performance of several non-autoregressive modes are shown in Figure~\ref{fig:autoVsNonauto}.  One common ingredient in getting non-autoregressive systems to perform well is to train them on a corpus of distilled translations~\citep{DBLP:conf/emnlp/KimR16}. This distilled corpus consists of source sentences paired with the translations produced by a pretrained autoregressive ``teacher'' system. 

Non-autoregressive neural machine translation began with the work of \citet{gu2018non}, who found benefit from using knowledge distillation~\citep{hinton2015distilling}, and in particular sequence-level distilled outputs~\citep{DBLP:conf/emnlp/KimR16}. 
Subsequent work has narrowed the gap between non-autoregressive and autoregressive translation, including multi-iteration refinements~\citep{lee-etal-2018-deterministic, ghazvininejad-etal-2019-mask,saharia2020nonautoregressive,kasai2020parallel} and rescoring with autoregressive models~\citep{kaiser2018fast,wei-etal-2019-imitation,ma-etal-2019-flowseq,NIPS2019_8566}.  \citet{ghazvininejad2020aligned} and \citet{saharia2020nonautoregressive} proposed aligned cross entropy or latent alignment models and achieved the best results of all non-autoregressive models without refinement or rescoring. 
We propose training inference networks with autoregressive energies and outperform the best purely non-autoregressive methods.

Another related approach trains an ``actor'' network to manipulate the hidden state of an autoregressive neural MT system~\citep{gu-etal-2017-trainable,chen-etal-2018-stable,zhou20iclr} in order to bias it toward outputs with better BLEU scores. This work modifies the original pretrained network rather than using it to define an energy for training an inference network.

\section{Generalized Energy and Inference Network for NMT}

\begin{figure}
\centering
\includegraphics[width=0.6\textwidth]{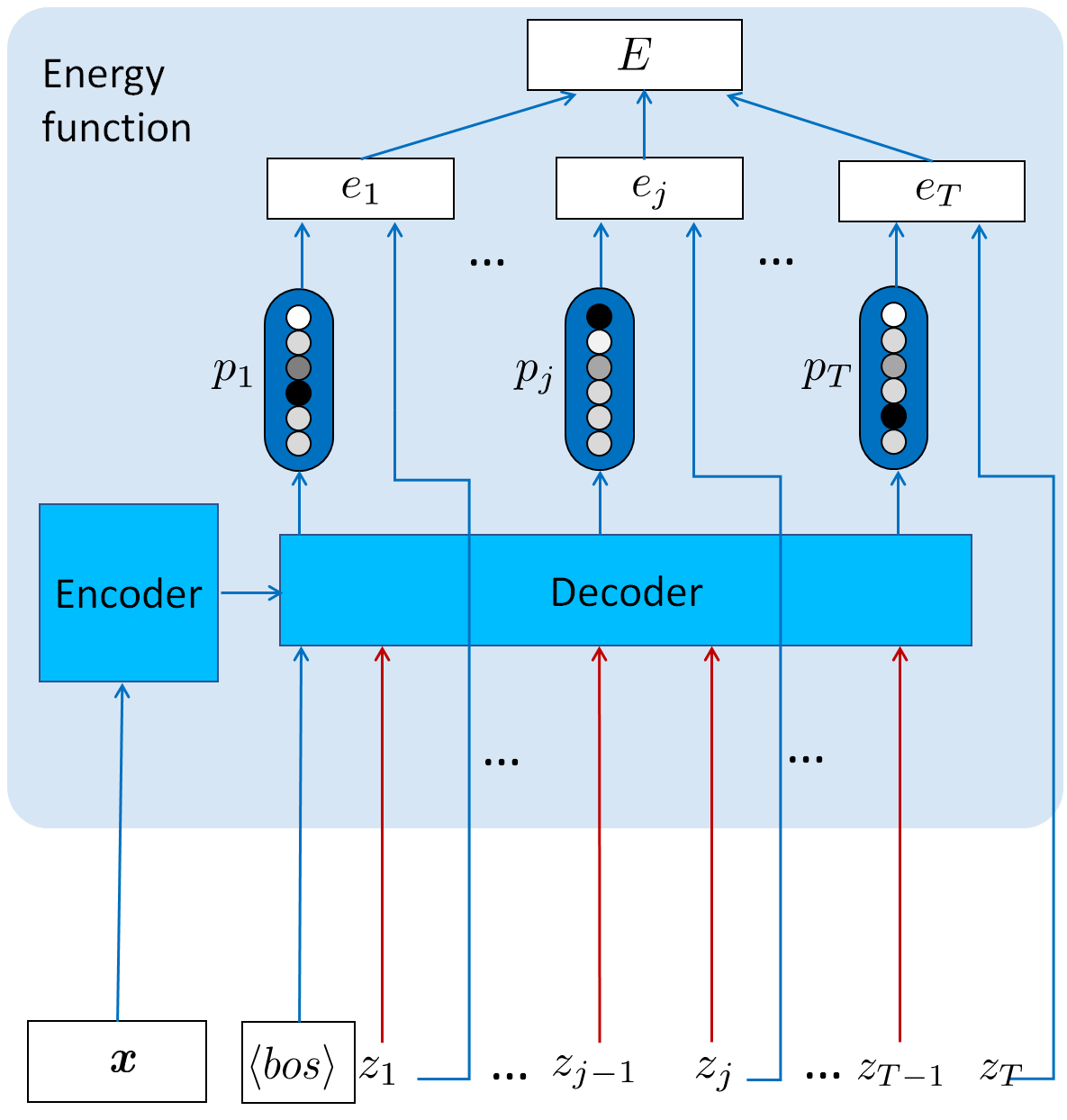}
\caption{The autogressive model can be used to score a sequence of words. The beam search algorithm is also to minimize the score (Energy)}
\end{figure}

\begin{figure}
\centering
\includegraphics[width=0.6\textwidth]{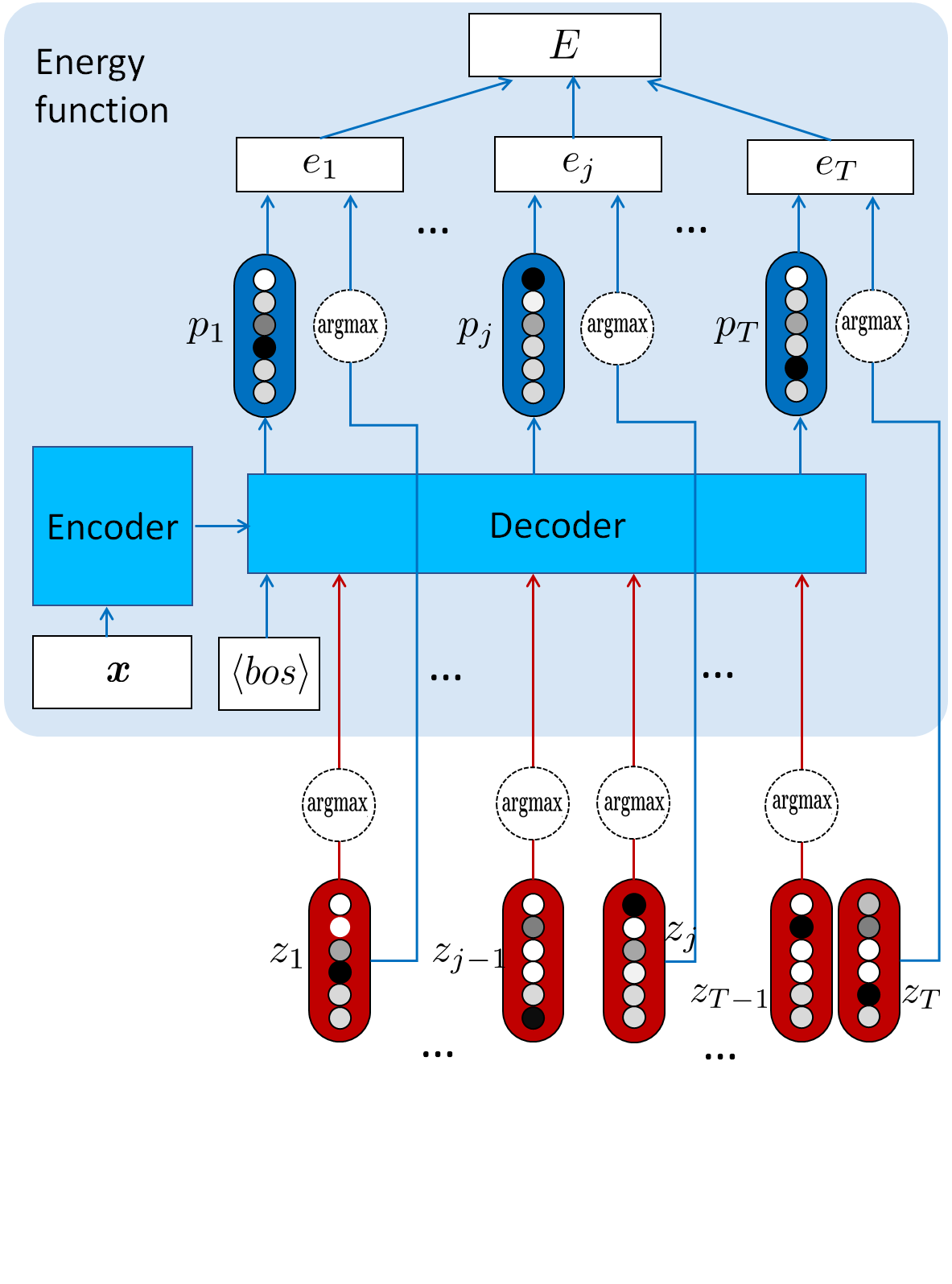}
\caption{The autoregressive models can be used to score a sequence of word distributions with argmax operations.}
\end{figure}

Most neural machine translation (NMT) systems model the conditional distribution $p_{\Theta}(\y \given \x)$ of a target sequence $\y = \langle y_1, y_2,..., y_{T}\rangle$ given a source sequence $\x = \langle x_1, x_2,..., x_{T_s}\rangle$, where each $y_t$ comes from a vocabulary $\mathcal{V}$, $y_T$ is $\langle \mathit{eos}\rangle$, and $y_{0}$ is $\langle \mathit{bos}\rangle$. 
It is common in NMT to define this conditional distribution using an ``autoregressive'' factorization~\citep{sutskever2014sequence,BahdanauCB14,NIPS2017_7181}: 
\begin{align}
    \log p_{\Theta}(\y \given \x) =  \sum_{t=1}^{|\y|} \log p_{\Theta}(y_t \mid \y_{0:t-1}, \x) \nonumber
\end{align}
This model can be viewed as an energy-based model~\citep{lecun-06} by defining the \textbf{energy function} $E_\Theta(\x, \y)= -\log p_{\Theta}(\y \given \x)$. 
Given trained parameters $\Theta$, 
test time inference seeks to find the translation for a given source sentence $\x$ with the lowest energy: $\hat{\y} = \argmin_{\y} \,E_\Theta(\x, \y)$. 

Finding the translation that minimizes the energy involves combinatorial search. 
We train \textbf{inference networks} to perform this search approximately. 
The idea of this approach is to replace the test time combinatorial search typically employed in structured prediction  with the output of a network trained to produce approximately optimal predictions as shown in Section~\ref{sec:benchmark} and Section~\ref{sec:SPENTraining}. 
More formally, 
we define an inference network $\infnet$ which maps an input $\x$ to a translation $\y$ and 
is trained with the goal that 
$\infnet(\x) \approx \argmin_{\y} E_\Theta(\x, \y)$.

Specifically, we train the inference network parameters $\Psi$ as follows (assuming $\Theta$ is pretrained and fixed):
\begin{align}
\widehat{\Psi} = \argmin_{\Psi}\!\!\sum_{\langle\x,\y\rangle \in \mathcal{D}} \!\! E_\Theta(\x, \infnet(\x))\!  
\end{align}
\noindent where $\mathcal{D}$ is a training set of sentence pairs. The network architecture of $\infnet$ can be different from the architectures used in the energy function. In this paper, we combine an autoregressive energy function with a non-autoregressive inference network. By doing so, we seek to combine the effectiveness of the autoregressive energy with the fast inference speed of a non-autoregressive network.

In order to allow for gradient-based optimization of the inference network parameters $\Psi$, we now define a more general family of  
energy functions for NMT. First, we change the representation of the translation $\y$ in the energy, redefining $\y=\langle \y_0,\dots, \y_{|\y|} \rangle$ as a sequence of 
\textbf{\emph{distributions}} over words instead of a sequence of words.

In particular, we consider the generalized energy
\begin{equation}
E_\Theta(\x, \y) = \sum_{t=1}^{|\y|} e_t (\x,\y)
\label{eq:loss}  
\end{equation}
where
\begin{equation}
e_t(\x, \y) = - \y_t^\top \log p_{\Theta} (\cdot \mid \y_{0}, \y_{1}, \dots, \y_{t-1}, \x).
\label{eq:localen}
\end{equation}
We use the $\cdot$ notation in $p_{\Theta} (\cdot \mid \ldots)$ above to indicate that we may need the full distribution over words. Note that by replacing the $\y_t$ with one-hot distributions we recover the original energy.

\begin{figure}[t]
\centering
\includegraphics[width=0.7\textwidth]{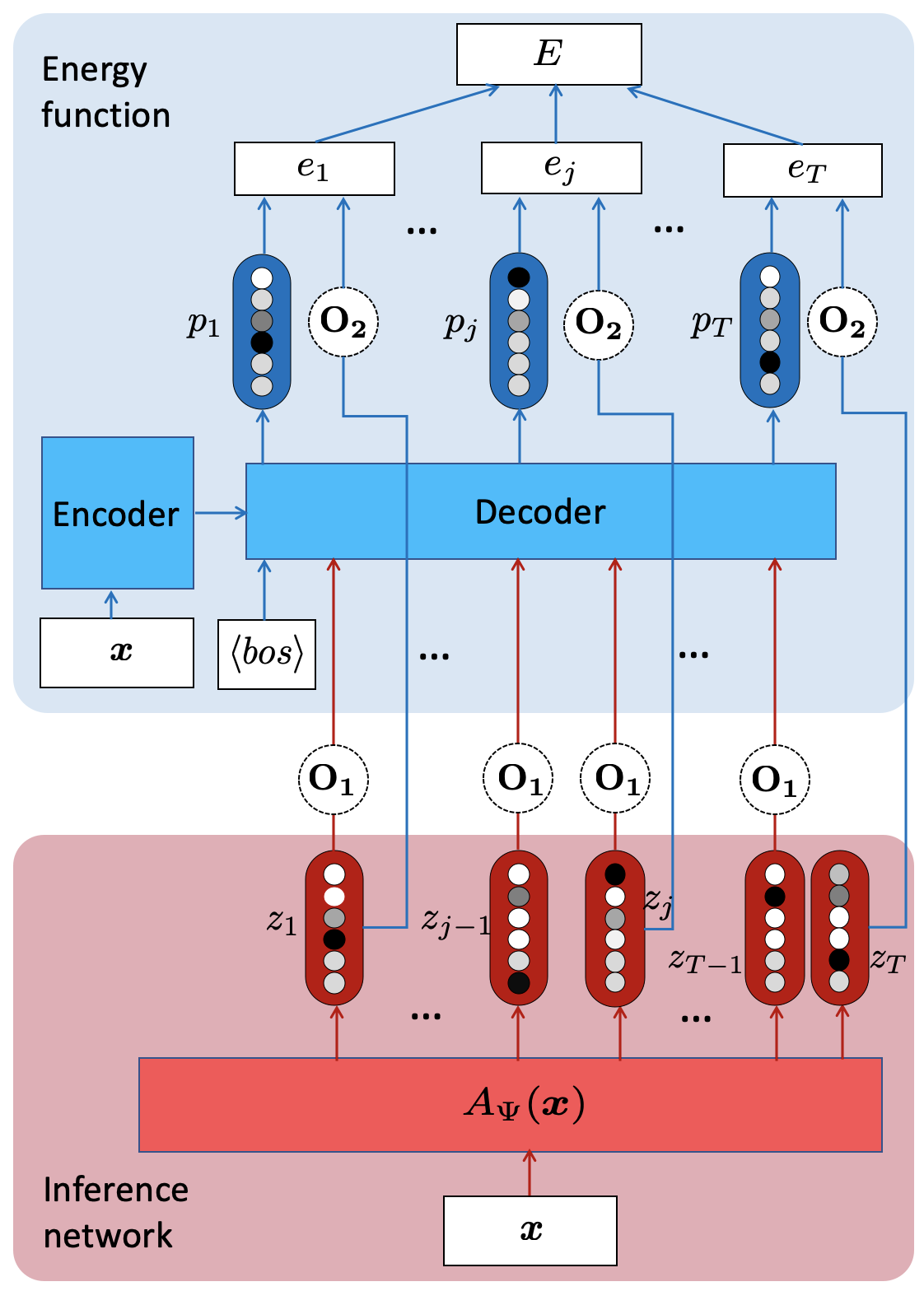}
\caption{The model for learning test-time inference networks for NAT-NMT when the energy function $E_\Theta(\x, \y)$ is a pretrained seq2seq model with attention.
\label{fig:model}}
\end{figure}

In order to train an inference network to minimize this energy, we simply need a network architecture that can produce a sequence of word distributions, which is satisfied by recent non-autoregressive NMT models~\citep{ghazvininejad-etal-2019-mask}. However, because the distributions involved in the original energy are one-hot, it may be advantageous for the inference network too to output distributions that are one-hot or approximately so. We will accordingly view inference networks as producing a sequence of $T$ logit vectors $\z_t \in \mathbb{R}^\mathcal{|V|}$, and we will consider two operators $\oone$ and $\otwo$ that will be used to map these $\z_t$ logits into distributions for use in the energy. Figure~\ref{fig:model} provides an overview of our approach, including this generalized energy function, the inference network, and the two operators $\oone$ and $\otwo$.

\begin{table}[h]
  \begin{center}
    \begin{tabular}{lcc}
      \toprule
      & $\ogen(\z)$ & $\frac{\partial \ogen(\z)}{\partial \z}$ \\
      \midrule
      \textbf{SX} & $\q$ & $ \frac{\partial \q}{\partial \z}$ \\[1ex]
      \textbf{STL} & $\onehot{ (\argmax (\z))}$ & $\boldsymbol{I}$ \\ [1ex]
      \textbf{SG} &  $\onehot{ (\argmax (\tilde{\q}))}$ & $ \frac{\partial \tilde{\q}}{\partial \tilde{\z}}$ \\[1ex]
    \textbf{ST} & $\onehot{ (\argmax (\q))}$ & $ \frac{\partial \q}{\partial \z}$ \\[1ex]
     \textbf{GX} & $\tilde{\q}$  &  $\frac{\partial \tilde{\q}}{\partial \tilde{\z}}$ \\[1ex] 
     \bottomrule
    \end{tabular}
    \caption{Let $\ogen(\z) \, {\in} \, \Delta^{|\mathcal{V}|-1}$ be the result of applying an $\oone$ or $\otwo$ operation to logits $\z$ output by the inference network. Also let $\tilde{\z} \, {=} \, \z \, {+} \, \boldsymbol{g}$, where $\boldsymbol{g}$ is Gumbel noise, $\q \, {=} \, \softmax(\z)$, and $\tilde{\q} \, {=} \, \softmax(\tilde{\z})$. We show the Jacobian (approximation) $\frac{\partial \ogen(\z)}{\partial \z}$ we use when computing  $\frac{\partial \loss}{\partial \z} \, {=} \, \frac{\partial \loss}{\partial \ogen(\z)} \frac{\partial \ogen(\z)}{\partial \z}$, for each $\ogen(\z)$ considered. 
    }
     \label{tab:operator}
  \end{center}
\end{table}
\vspace{2.5cm}

\section{Choices for Operators}
\label{sec:operations}

The choices we consider for $\oone$ and $\otwo$, which we present generically for operator $\ogen$ and logit vector $\z$, are shown in Table~\ref{tab:operator}, and described in more detail below. Some of these $\ogen$ operations are not differentiable, and so the Jacobian matrix $\frac{\partial \ogen(\z)}{\partial \z}$ must be approximated during learning; we show the approximations we use in Table~\ref{tab:operator} as well.

We consider five choices for each $\ogen$:

\begin{itemize}
\item[(a)] \textbf{SX}: $\softmax$. Here $\ogen(\z) \, {=} \, \softmax(\z)$; no Jacobian approximation is necessary.

\item[(b)] \textbf{STL}: straight-through logits. Here $\ogen(\z) \, {=} \, \onehot{(\argmax_i \z})$. $\frac{\partial \ogen(\z)}{\partial \z}$ is approximated by the identity matrix $\boldsymbol{I}$ (see \citet{BengioLC13}).

\item[(c)] \textbf{SG}: straight-through Gumbel-Softmax. Here $\ogen(\z) \, {=} \, \onehot{(\argmax_i \softmax(\z + \boldsymbol{g})})$, where $g_i$ is Gumbel noise. $g_i=-\log(-\log(u_i))$ and $u_i \sim \mathbf{Uniform}(0,1)$. $\frac{\partial \ogen(\z)}{\partial \z}$ is approximated with $\frac{\partial \, \softmax(\z + \boldsymbol{g})}{\partial \z}$ \citep{jang2016categorical}.

\item[(d)] \textbf{ST}: straight-through. This setting is identical to SG with $\boldsymbol{g} \, {=} \, \boldsymbol{0}$ (see ~\citet{BengioLC13}).

\item[(e)] \textbf{GX}: Gumbel-Softmax. Here $\ogen(\z) \, {=} \, \softmax(\z + \boldsymbol{g})$, where again $g_i$ is Gumbel noise; no Jacobian approximation is necessary.
\end{itemize}
\vspace{1.5cm}

\begin{table}[h]
\centering
    \setlength{\tabcolsep}{1pt}
    \begin{subtable}{0.5\columnwidth}
      \centering
      \small
        \begin{tabular}{c  ccccc}
    \toprule
	$\oone \setminus \otwo$ & SX & STL & SG & ST & GX\\ 
    \midrule
    SX & \textbf{55 (20.2)} & 256 (0) & 56 (19.6) & \textbf{55 (20.1)} & 55 (19.6) \\
    STL & 97 (14.8) & 164 (8.2)  & 94 (13.7) & 95 (14.6) & 190 (0)\\
    SG & 82 (15.2) & 206 (0) & 81 (14.7) & 82 (15.0)  & 83 (13.5)\\
    ST & 81 (14.7) & 170 (0) & 81 (14.4) & 80 (14.3) & 83 (13.7)  \\
    GX & \textbf{53 (19.8)} & 201 (0) & 56 (18.3) & 54 (19.6)  & 55 (19.4) \\
    \bottomrule
    \end{tabular}
    \caption{seq2seq AR energy, \\
    BiLSTM inference networks}
    \end{subtable}
    \  \ 
    \begin{subtable}{0.45\columnwidth}
      \centering
      \small
        \begin{tabular}{  ccccc}
    \toprule
	SX & STL & SG & ST & GX \\ 
    \midrule
       \textbf{80 (31.7)} &  133 (27.8) &  81 (31.5) &  \textbf{80 (31.7)} & 81 (31.6)\\
      186 (25.3) & 133 (27.8)  & 95 (20.0) & 97 (30.1) & 180 (26.0)\\
    
      98 (30.1) & 133 (27.8) & 95 (30.1) & 97 (30.0) & 97 (29.8)\\

      98 (30.2) & 133 (27.8) & 95 (30.0) & 97 (30.1) & 97 (30.0)\\
       81 (31.5) & 133 (27.8) &  81 (31.2) &  81 (31.5)  & 81 (31.4) \\
    \bottomrule
    \end{tabular}
    \caption{transformer AR energy, \\ 
    CMLM inference networks} 
    \end{subtable} 
    \caption{Comparison of operator choices in terms of energies (BLEU scores) 
on the IWSLT14 DE-EN dev set with two energy/inference network combinations. 
Oracle lengths are used for decoding. $\oone$ is the operation for feeding inference network outputs into the decoder input slots in the energy. $\otwo$ is the operation for computing the energy on the output. Each row corresponds to the same $\oone$, and each column corresponds to the same $\otwo$.  
}
\label{tab:argmax}
\end{table}

\section{Experimental Setup}
\subsubsection{Datasets} 
We evaluate our methods on two datasets: IWSLT14 German (DE) $\rightarrow$ English (EN) and WMT16 Romanian (RO) $\rightarrow$ English (EN). 
All data are tokenized and then segmented into subword units using byte-pair encoding \citep{sennrich-etal-2016-neural}. We use the data provided by \citet{lee-etal-2018-deterministic} for RO-EN. 

\subsection{Autoregressive Energies}
We consider two architectures for the pretrained autoregressive (AR) energy function. 
The first is an autoregressive sequence-to-sequence (seq2seq) model with attention~\citep{luong-etal-2015-effective}. The encoder is a two-layer BiLSTM with 512 units in each direction, the decoder is a two-layer LSTM with 768 units, and the word embedding size is 512. 
The second is an autoregressive transformer model~\citep{NIPS2017_7181}, where both the encoder and decoder have 6 layers, 8 attention heads per layer, model dimension 512, and hidden dimension 2048.

\subsection{Inference Network Architectures}

We choose two different architectures: a BiLSTM ``tagger'' (a 2-layer BiLSTM followed by a fully-connected layer) and a conditional masked language
model (CMLM; \citealp{ghazvininejad-etal-2019-mask}), 
a transformer with 6 layers per stack, 8 attention heads per layer, model dimension 512, 
and hidden dimension 2048. Both architectures require the target sequence length in advance; methods for handling length are discussed in Sec.~\ref{sec:lengths}. 
For baselines, we train these inference network architectures as non-autoregressive models using the standard per-position cross-entropy loss. 
For faster inference network training, we initialize inference networks with the baselines trained with cross-entropy loss in our experiments. 

\begin{figure}[h]
\centering
\includegraphics[width=0.8\textwidth]{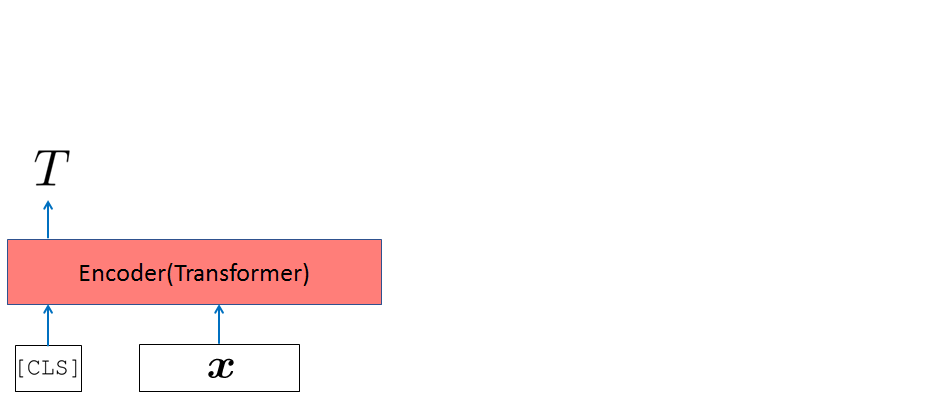}
\caption{The architecture of CMLM. Predicting target sequence length $T$ according to the encoder. }
\end{figure}

\vspace{2.5cm}

\begin{figure}[h]
\centering
\includegraphics[width=0.8\textwidth]{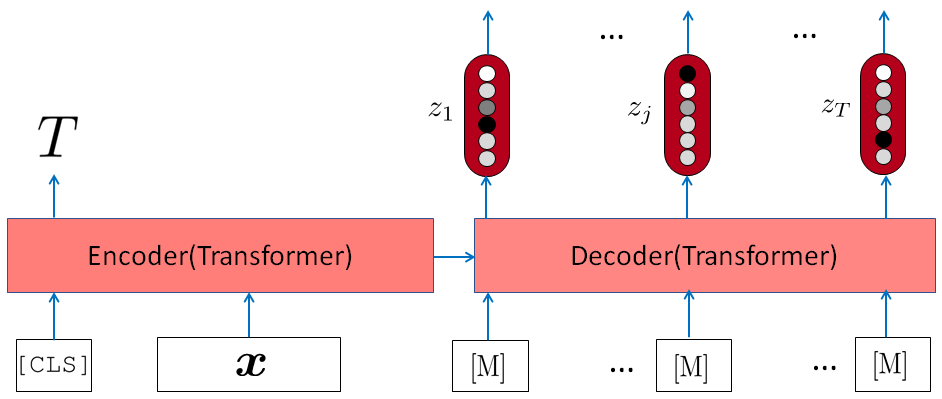}
\caption{The architecture of CMLM. The decoder inputs arethe special masked tokens [M]. }
\end{figure}
\vspace{1.5cm}

The baseline CMLMs use the partial masking strategy
described by \citet{ghazvininejad-etal-2019-mask}. This involves using some masked input tokens and some provided input tokens during training. At test time, multiple iterations (``refinement iterations'') can be used for improved results~\citep{ghazvininejad-etal-2019-mask}. Each iteration uses partially-masked input from the preceding iteration. We consider the use of multiple refinement iterations for both the CMLM baseline and the CMLM inference network. The CMLM inference network is trained with full masking (no partial masking like in the CMLM baseline). However, since the CMLM inference network is initialized using the CMLM baseline, which is trained using partial masking, the CMLM inference network is still compatible with refinement iterations at test time.

\subsection{Hyperparameters}
For inference network training, the batch size is 1024 tokens. We train with the Adam optimizer \citep{kingma2015adam}. We tune the learning rate in $\{5\mathrm{e}\!-\!4, 1\mathrm{e}\!-\!4, 5\mathrm{e}\!-\!5, 1\mathrm{e}\!-\!5, 5\mathrm{e}\!-\!6, 1\mathrm{e}\!-\!6\}$. For regularization, we use L2 weight decay with rate 0.01, and dropout with rate 0.1. 
We train all models for 30 epochs. For the baselines, we train the models with local cross entropy loss and do early stopping based on the BLEU score on the dev set. For the inference network, we train the model to minimize the energy and do early stopping based on the energy on the dev set.

\subsection{Predicting Target Sequence Lengths}
\label{sec:lengths}

Non-autoregressive models  
often need a target sequence length in advance~\citep{lee-etal-2018-deterministic}. 
We report results both with oracle lengths and 
with a simple method of predicting it. 
We follow \citet{ghazvininejad-etal-2019-mask} in predicting the length of the translation using a representation of the source sequence from the encoder. 
The length loss is added to the cross-entropy loss for the target sequence. During decoding, we select the top $k=3$ length candidates with the highest probabilities, 
decode with the different lengths in parallel, and return the translation with the highest average of log probabilities of its tokens. 

\section{Results}

\paragraph{Effect of choices for $\oone$ and $\otwo$.} 

Table~\ref{tab:argmax} compares various choices for the operations $\oone$ and $\otwo$. 
For subsequent experiments, we choose the setting that feeds the whole distribution into the energy function ($\oone$ = SX) and computes the loss with straight-through ($\otwo$ = ST). 
Using Gumbel noise in $\otwo$ has only minimal effect, and rarely helps. Using ST instead also speeds up training by avoiding the noise sampling step.

\paragraph{Training with Distilled Outputs vs. Training with Energy}

In order to compare \name with training on distilled outputs, 
we train BiLSTM models in three ways: ``baseline'' which is trained with the human-written reference translations, ``distill'' which is trained with the distilled outputs (generated using the autoregressive models), and  ``\name'', our method which trains the BiLSTM as an inference network to minimize the pretrained seq2seq autoregressive energy. Oracle lengths are used for decoding.  Table~\ref{tab:cropus} shows test results for both datasets, showing significant gains of \name over the baseline and distill methods. Although the results shown here are lower than the transformer results, the trend is clearly indicated.

\begin{table}[!th]
\setlength{\tabcolsep}{3pt}
\centering
\small
\begin{tabular}{ c | c c | c c}
    \toprule
      &  \multicolumn{2}{c|}{IWSLT14 DE-EN} & \multicolumn{2}{c}{WMT16 RO-EN} \\ 
      & {Energy ($\downarrow$) } &  {BLEU ($\uparrow$)} &  {Energy ($\downarrow$)} &  {BLEU ($\uparrow$)}\\
    \midrule
 baseline  & 153.54 & 8.28 & 175.94 & 9.47 \\
distill  & 112.36  & 14.58 & 205.71 & 5.76\\ 
\name & 51.98 & 19.55 & 64.03 & 21.69 \\ 
\bottomrule
\end{tabular}
\caption{Test results of non-autoregressive models when training with the references (``baseline''), distilled outputs (``distill''), and energy (``\name''). Oracle lengths are used for decoding. Here, \name uses BiLSTM inference networks and pretrained seq2seq AR energies.  \name  outperforms training on both the references and a pseudocorpus. 
}
\label{tab:cropus}
\end{table}

\paragraph{Impact of refinement iterations.}
\citet{ghazvininejad-etal-2019-mask} show improvements with multiple refinement iterations.  Table~\ref{tab:refinements} shows refinement results of CMLM and \name. Both improve with multiple iterations, though the improvement is much larger with CMLM. However, even with 10 iterations, \name is comparable to CMLM on DE-EN and outperforms it on RO-EN.

\begin{table}[h]
\centering
\small
\begin{tabular}{c c  c c c  c }
    \toprule
    & \multicolumn{2}{c}{IWSLT14 DE-EN} & & \multicolumn{2}{c}{WMT16 RO-EN} \\
    \cline{2-3} \cline{5-6}
    \noalign{\smallskip}
    & \multicolumn{2}{c}{\# iterations} & & \multicolumn{2}{c}{\# iterations} \\ 
    \cline{2-3} \cline{5-6}
    \noalign{\smallskip}
    & 1 &  10 & & 1 &  10  \\
    \midrule
 CMLM  & 28.11 & 33.39 & & 28.20 & 33.31\\
 \name  &  31.99 & 33.17 &  & 33.16 & 34.04\\
\bottomrule
\end{tabular}
\caption{Test BLEU scores of non-autoregressive models using no refinement (\# iterations = 1) and using refinement (\# iterations = 10). Note that the \# iterations = 1 results are purely non-autoregressive. \name uses a CMLM as the inference network architecture 
and the transformer AR energy. The length beam size is 5 for CMLM and 3 for \name. 
}
\label{tab:refinements}
\end{table}

\paragraph{Comparison to other NAT models.}
Table~\ref{tab:finalInfnet} shows 1-iteration  results on two datasets. To the best of our knowledge, \name achieves state-of-the-art NAT performance: 31.99 on IWSLT14 DE-EN and 33.16 on WMT16 RO-EN. In addition, \name achieves comparable performance with the autoregressive NMT model.

\begin{table}[h]
\small
\setlength{\tabcolsep}{2pt}
\begin{center}
\begin{tabular}{l l  c   c }
\toprule
 &  & \multicolumn{1}{c}{IWSLT14} & \multicolumn{1}{c}{WMT16} \\
 &  & {DE-EN} & {RO-EN} \\ 
\midrule
 \multicolumn{2}{l}{\textbf{Autoregressive} (Transformer)} \\
\midrule
\multicolumn{1}{l}{} & Greedy Decoding  &  33.00 &  33.33 \\  
\multicolumn{1}{l}{} & Beam Search  & 34.11  & 34.07  \\ 
\midrule
 
\multicolumn{2}{l}{\textbf{Non-autoregressive}} \\
\midrule
& \makecell[l]{Iterative Refinement 
\citep{lee-etal-2018-deterministic}} & -& 25.73{{\makebox[0pt][l]{$^\dagger$}}}\\
& NAT with Fertility \citep{gu2018non} & - & 29.06{{\makebox[0pt][l]{$^\dagger$}}} \\
& CTC~\citep{libovicky-helcl-2018-end} & - & 24.71{{\makebox[0pt][l]{$^\dagger$}}} \\
& FlowSeq~\citep{ma-etal-2019-flowseq}  & 27.55{{\makebox[0pt][l]{$^\dagger$}}} & 30.44{{\makebox[0pt][l]{$^\dagger$}}} \\

\multicolumn{1}{l}{} & \makecell[l]{CMLM 
\citep{ghazvininejad-etal-2019-mask}}  & 28.25 &  28.20{{\makebox[0pt][l]{$^\dagger$}}} \\ 
\multicolumn{1}{l}{} &  \makecell[l]{Bag-of-ngrams-based loss 
\citep{shao2019minimizing}} & - & 29.29{{\makebox[0pt][l]{$^\dagger$}}}   \\
\multicolumn{1}{l}{} & \makecell[l]{AXE CMLM  
\citep{ghazvininejad2020aligned}} & - & 31.54{{\makebox[0pt][l]{$^\dagger$}}}  \\ 
\multicolumn{1}{l}{} & \makecell[l]{Imputer-based model 
\citep{saharia2020nonautoregressive}} & - & 31.7{{\makebox[0pt][l]{$^\dagger$}}}   \\ 
\multicolumn{1}{l}{} & {ENGINE (ours)}  & \textbf{31.99} & \textbf{33.16}\\
\bottomrule
\end{tabular}
\caption{BLEU scores on two datasets for several non-autoregressive methods. The inference network architecture is the CMLM. 
For methods that permit multiple refinement iterations (CMLM, AXE CMLM, ENGINE), one decoding iteration is used (meaning the methods are purely non-autoregressive). 
$^\dagger$Results are from the corresponding papers.  
\label{tab:finalInfnet}
}
\end{center}
\end{table}

\section{Analysis of Translation Results}

\begin{table*}[!h]
\small
\centering
\begin{tabular}{|ll|}
	\hline
	\textbf{Source:} & \\ 
	\multicolumn{2}{|p{1\columnwidth}|}{
    seful onu a solicitat din nou tuturor partilor , inclusiv consiliului de securitate onu divizat sa se unifice si sa sustina negocierile pentru a gasi o solutie politica .}\\
	\textbf{Reference} : &\\ 
	\multicolumn{2}{|p{1\columnwidth}|}{the u.n. chief again urged all parties , including the divided u.n. security council , to unite and support inclusive negotiations to find a political solution .} \\
	\textbf{CMLM} :&\\ 
	\multicolumn{2}{|p{1\columnwidth}|}{the un chief again again urged all parties , including the divided un security council to unify and support negotiations in order to find a political solution .} \\
	\textbf{\name} :  &\\ 
	\multicolumn{2}{|p{1\columnwidth}|}{the un chief has again urged all parties , including the divided un security council to unify and support negotiations in order to find a political solution .} \\
    \hline
    \end{tabular}

\begin{tabular}{|ll|}
	\hline
	\textbf{Source:} & \\ 
	\multicolumn{2}{|p{1\columnwidth}|}{adevarul este ca a rupt o racheta atunci cand a pierdut din cauza ca a acuzat crampe in us , insa nu este primul jucator care rupe o racheta din frustrare fata de el insusi si il cunosc pe thanasi suficient de bine incat sa stiu ca nu s @-@ ar mandri cu asta .
    }\\
	\textbf{Reference} : &\\ 
	\multicolumn{2}{|p{1\columnwidth}|}{he did break a racquet when he lost when he cramped in the us , but he \&apos;s not the first player to break a racquet out of frustration with himself , and i know thanasi well enough to know he wouldn \&apos;t be proud of that .} \\
	\textbf{CMLM} :&\\ 
	\multicolumn{2}{|p{1\columnwidth}|}{the truth is that it has broken a rocket when it lost because accused crcrpe in the us , but it is not the first player to break rocket rocket rocket frustration frustration himself himself and i know thanthanasi enough enough know know he would not be proud of that .} \\
	\textbf{\name} :  &\\ 
	\multicolumn{2}{|p{1\columnwidth}|}{the truth is that it broke a rocket when it lost because he accused crpe in the us , but it is not the first player to break a rocket from frustration with himself and i know thanasi well well enough to know he would not be proud of it .} \\
    \hline
    \end{tabular}
    
    \begin{tabular}{|ll|}
	\hline
	\textbf{Source:} & \\ 
	\multicolumn{2}{|p{1\columnwidth}|}{realizatorii studiului mai transmit ca \&quot; romanii simt nevoie de ceva mai multa aventura in viata lor ( 24 \% ) , urmat de afectiune ( 21 \% ) , bani ( 21 \% ) , siguranta ( 20 \% ) , nou ( 19 \% ) , sex ( 19 \% ) , respect 18 \% , incredere 17 \% , placere 17 \% , conectare 17 \% , cunoastere 16 \% , protectie 14 \% , importanta 14 \% , invatare 12 \% , libertate 11 \% , autocunoastere 10 \% si control 7 \% \&quot; .
    }\\
	\textbf{Reference} : &\\ 
	\multicolumn{2}{|p{1\columnwidth}|}{the study \&apos;s conductors transmit that \&quot; romanians feel the need for a little more adventure in their lives ( 24 \% ) , followed by affection ( 21 \% ) , money ( 21 \% ) , safety ( 20 \% ) , new things ( 19 \% ) , sex ( 19 \% ) respect 18 \% , confidence 17 \% , pleasure 17 \% , connection 17 \% , knowledge 16 \% , protection 14 \% , importance 14 \% , learning 12 \% , freedom 11 \% , self @-@ awareness 10 \% and control 7 \% . \&quot; } \\
	\textbf{CMLM} :&\\ 
	\multicolumn{2}{|p{1\columnwidth}|}{survey survey makers say that \&apos; romanians romanians some something adventadventure ure their lives 24 24 \% ) followed followed by \% \% \% \% \% , ( 21 \% \% ), safety ( \% \% \% ), new19\% \% ), ), 19 \% \% \% ), respect 18 \% \% \% \% \% \% \% \% , , \% \% \% \% \% \% \% , , \% , 14 \% , 12 \% \% } \\
	\textbf{\name} :  & \\ 
	\multicolumn{2}{|p{1\columnwidth}|}{realisation of the survey say that \&apos; romanians feel a slightly more adventure in their lives ( 24 \% ) followed by aff\% ( 21 \% ) , money ( 21 \% ), safety ( 20 \% ) , new 19 \% ) , sex ( 19 \% ) , respect 18 \% , confidence 17 \% , 17 \% , connecting 17 \% , knowledge \% \% , 14 \% , 14 \% , 12 \% \%} \\
    \hline
    \end{tabular}
    \caption{Examples of translation outputs from \name and CMLM on WMT16 RO-EN without refinement iterations.}
    \label{tab:translations}
\end{table*}

In Table~\ref{tab:translations}, we present randomly chosen translation outputs from WMT16 RO-EN. For each Romanian sentence, we show the reference from the dataset, the translation from CMLM, and the translation from \name. We could observe that without the refinement iterations, CMLM could performs well for shorter source sentences. However, it still prefers generating repeated tokens. \name, on the other hand, could generates much better translations with fewer repeated tokens.

\section{Conclusion}
We proposed a new method to train non-autoregressive neural machine translation systems via minimizing pretrained energy functions with inference networks. In the future, we seek to expand upon energy-based translation using our method. 
\newpage

\mychapter{5}{SPEN Training Using Inference Networks}
\label{sec:SPENTraining}
In the previous two chapters, we discussed training inference networks for a pretrained, fixed energy function for sequence labeling and neural machine translation. In this chapter, we now describe our completed work in joint learning of energy functions and inference networks.

This chapter includes some material originally presented in ~\citet{tu-18}.

\section{Introduction}

Deep energy-based models are powerful, but pose challenges for learning and inference. In chapter 2, we show several previous methods. \citet{belanger2016structured} proposed a structured hinge loss:

\begin{align}
\min_{\Theta} \sum_{\langle \x_i, \y_i\rangle\in\mathcal{D}} \left[\max_{\y\in\relyspace(\x)} \left(\cost(\y, \y_i) - E_{\Theta}(\x_i,\y) + E_{\Theta}(\x_i, \y_i)\right)\right]_{+}
\label{eq:ssvm}
\end{align}
\noindent where $\mathcal{D}$ is the set of training pairs, $\relyspace$ is the relaxed output space, $[f]_+ = \max(0,f)$, and $\cost(\y,\y')$ is a structured \textbf{cost} function that returns a nonnegative value indicating the difference between $\y$ and $\y'$. This loss is often referred to as ``margin-rescaled'' structured hinge loss~\citep{m3,tsochantaridis2005large}.



During learning, there is a cost-augmented inference step:

\begin{align}
     \y_{F} = \argmax_{\y\in\relyspace(\x)} \cost(\y, \y_i) - E_{\Theta}(\x_i,\y) + E_{\Theta}(\x_i, \y_i) \nonumber
\end{align}

After learning the energy function, prediction minimizes energy:
\begin{align}
\hat{\y} = \argmin_{\y\in\yspace(\x)}E_{\Theta}(\x, \y) \nonumber
\end{align}
However, solving the above equations requires combinatorial algorithms because $\yspace$ is a discrete structured space. This becomes intractable when $E_{\Theta}$ does not decompose into a sum over small ``parts'' of $\y$. \citet{belanger2016structured} relax this problem by allowing the discrete vector $\y$ to be continuous. 
 For MLC, $\relyspace(\x) = [0,1]^L$.  
They solve the relaxed problem 
by using gradient descent to iteratively optimize the energy with respect to $\y$. In this chappter, We also relax $\y$ but we use a different strategy to approximate inference and learning energy function. 

\vspace{0.2cm}

The following section show how to jointly train SPENs and inference networks. 

\section{Joint Training of SPENs and Inference Networks}

\citet{belanger2016structured} train with a structured large-margin objective, repeated inference is required during learning. However, this loss is expensive to minimize for structured models because of the ``cost-augmented'' inference step ($\max_{\y\in\relyspace(\x)}$). In prior work with SPENs, this step used gradient descent. They note that using gradient descent for this inference step is time-consuming and makes learning less stable. So \citet{End-to-EndSPEN} propose an ``end-to-end'' learning procedure inspired by \citet{AISTATS2012_Domke12}. This approach performs backpropagation through each step of gradient descent. 


We replace this with a \textbf{cost-augmented inference network} $\canet(\x)$.
\begin{align}
    \canet(\x_i) \approx \y_{F} = \argmax_{\y \in \relyspace(\x)}  \cost(\y, \y_i) - E_\Theta(\x_i, \y)  \nonumber
\end{align}

\paragraph{Can Approximate Inference Be Used During Training?} 

The cost-augmented inference network $\canet$ is to approximate output $y$ with high cost and low energy. The approximate inference can potentially include search error. Although the approximate inference method is used, the energy function may incorrectly
assign low energy to some modes. Firstly, $\canet$ can be a powerful deep neural network that has enough capacity. Secondly, if some answers $\y$ with really low energies can not found by the inference method during training. It generally means these answers with low energies will also not be found by the inference method. So We do not need to worry about them. Chapter 8.3 of LeCun's tutorial~\citep{lecun-06} also has some discussion. Approximate inference can be used during training.

The cost-augmented inference network $\canet$ and the inference network $\infnet$ can have the same functional form, but use different parameters $\Phi$ and $\Psi$. 

We write our new optimization problem as:
\begin{align}
\min_{\Theta} \max_{\Phi}\sum_{\langle \x_i, \y_i\rangle\in\mathcal{D}} \left[\cost(\canet(\x_i), \y_i) - E_{\Theta}(\x_i,\canet(\x_i)) + E_{\Theta}(\x_i, \y_i)\right]_{+} 
\end{align}

\begin{figure}[!h]
\centering
\includegraphics[width=0.95\textwidth]{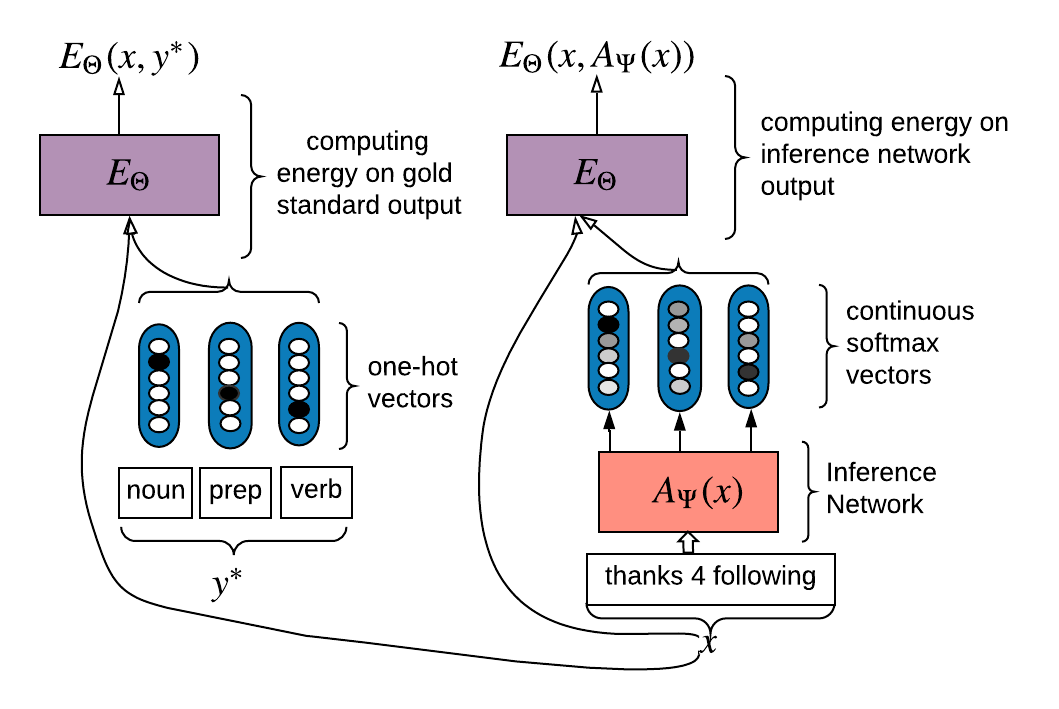}
\caption{
The architectures of inference network $\infnet$ and energy network $E_\Theta$. \label{fig:spen_inf}
}
\end{figure}

Figure \ref{fig:spen_inf} shows the architectures of inference network $\canet$ and energy network $E_\Theta$. 

We treat this optimization problem as a minmax game and find a saddle point for the game. Following \citet{goodfellow2014generative}, we implement this using an iterative numerical approach. We alternatively optimize $\Phi$ and $\Theta$, holding the other fixed. Optimizing $\Phi$ to completion in the inner loop of training is computationally prohibitive and may lead to overfitting. So we alternate between one mini-batch for optimizing $\Phi$ and one for optimizing $\Theta$. We also add $L_2$ regularization terms for $\Theta$ and $\Phi$.

The objective for the cost-augmented inference network is:

\begin{align}
\hat{\Phi} \gets \argmax_{\Phi}[\cost(\canet(\x_i), \y_i) - E_{\Theta}(\x_i,\canet(\x)_i) + E_{\Theta}(\x_i, \y_i)]_{+} \label{eq:up-infnet}
\end{align}

That is, we update $\Phi$ so that $\canet$ yields an output that has low energy and high cost, in order to mimic cost-augmented inference. The energy parameters $\Theta$ are kept fixed. There is an analogy here to the generator in GANs: $\canet$ is trained to produce a high-cost structured output that is also appealing to the current energy function. 

The objective for the energy function is:
\begin{align}
\hat{\Theta} \gets & \argmin_{\Theta}[\bigtriangleup(\mathbf{A}_{\Phi}(\x_i), \y_i) - E_{\Theta}(\x_i,\canet(\x_i)) + E_{\Theta}(\x_i, \y_i)]_{+} + \lambda \| \Theta \|_2^2\label{eq:up-energy}
\end{align}
That is, we update $\Theta$ so as to widen the gap between the cost-augmented and ground truth outputs. There is an analogy here to the discriminator in GANs. The energy function is updated so as to enable it to distinguish ``fake'' outputs produced by $\canet$ from real outputs $\y_i$. Training iterates between updating $\Phi$ and $\Theta$ using the objectives above.

\section{Test-Time Inference}

After training, we want to use an inference network $\infnet$ defined in Eq.~\eqref{eq:infnet}. However, training only gives us a cost-augmented inference network $\canet$. Since $\infnet$ and $\canet$ have the same functional form, we can use $\Phi$ to initialize $\Psi$, then do additional training on $\mathbf{A}_{\Psi}$ as in Eq.~\eqref{eq:finalnet}
where $X$ is the training or validation set. This step helps the resulting inference network to produce outputs with lower energy, as it is no longer affected by the cost function. Since this procedure does not use the output labels of the $\x$'s in $X$, it could also be applied to the test data in a transductive setting.

\section{Variations and Special Cases}
\label{sec:summary-losses}

This approach also permits us to use large-margin structured prediction with slack rescaling~\citep{tsochantaridis2005large}. 
Slack rescaling can yield higher accuracies than margin rescaling, but requires ``cost-scaled'' inference during training which is intractable for many classes of output structures. 

However, we can use our notion of inference networks to circumvent this tractability issue and approximately optimize the slack-rescaled hinge loss, yielding the following optimization problem:

\begin{align}
\min_{\Theta} \max_{\Phi}\sum_{\langle \x_i, \y_i\rangle\in\mathcal{D}} \bigtriangleup(\canet(\x_i), \y_i) [1- E_{\Theta}(\x_i,\canet(\x_i)) + E_{\Theta}(\x_i, \y_i)]_{+} \label{eq:slack-hinge}
\end{align}
Using the same argument as above, we can also break this into alternating optimization of $\Phi$ and $\Theta$.

We can optimize a structured perceptron~\citep{collins2002discriminative} version by using the margin-rescaled hinge loss (Eq.~\eqref{eq:margin-hinge}) and fixing $\cost(\canet(\x_i), \y_i) = 0$. When using this loss, the cost-augmented inference network is actually a test-time inference network, because the cost is always zero, so using this loss may lessen the need to retune the inference network after training. 

When we fix $\cost(\canet(\x_i), \y_i) = 1$, then margin-rescaled hinge is equivalent to slack-rescaled hinge. While using $\cost=1$ is not useful in standard max-margin training with exact $\argmax$ inference (because the cost has no impact on optimization when fixed to a positive constant), it is potentially useful in our setting. 

Consider our SPEN objectives with $\cost = 1$:
\begin{equation}
\left[1 - E_{\Theta}(\x_i,\canet(\x_i)) + E_{\Theta}(\x_i, \y_i)\right]_{+}
\end{equation}
\noindent There will always be a nonzero difference between the two energies because $\canet(\x_i)$ will never exactly equal the discrete vector $\y_i$. 

Since there is no explicit minimization over all discrete vectors $\y$, this case is more similar to a ``contrastive'' hinge loss which seeks to make the energy of the true output lower than the energy of a particular ``negative sample'' by a margin of at least 1.

\section{Improving Training for Inference Networks}
\label{sec:stable}
We found that the alternating nature of the optimization led to difficulties during training. Similar observations have been noted about other alternative optimization settings, especially those underlying generative adversarial networks~\citep{NIPS2016_6125}. 

Below we describe several techniques we found to help stabilize training, which are optional terms added to the objective in Eq.~\eqref{eq:up-infnet}.

\textbf{$L_2$ Regularization:} 
We use $L_2$ regularization, adding the penalty term $\| \Phi \|_2^2$ with coefficient $\lambda_1$. 

\textbf{Entropy Regularization:} 
We add an entropy-based regularizer $\mathrm{loss}_{\mathrm{H}}(\canet(\x))$ defined for the problem under consideration. 
For MLC, the output of $\canet(\x)$ is a vector of scalars in $[0,1]$, one for each label, where the scalar is interpreted as a label probability. The entropy regularizer $\mathrm{loss}_{\mathrm{H}}$ is the sum of the entropies over these label binary distributions. 
 
For sequence labeling, where the length of $\x$ is $N$ and where there are $L$ unique labels, the output of $\canet(\x)$ is a length-$N$ sequence of length-$L$ vectors, each of which represents the distribution over the $L$ labels at that position in $\x$. Then, $\mathrm{loss}_{\mathrm{H}}$ is the sum of entropies of these label distributions across positions in the sequence. 

When tuning the coefficient $\lambda_2$ for this regularizer, we consider both positive and negative values, permitting us to favor either low- or high-entropy distributions as the task prefers.\footnote{For MLC, encouraging lower entropy distributions worked better, while for sequence labeling, higher entropy was better, similar to the effect found by \citet{DBLP:journals/corr/PereyraTCKH17}. Further research is required to gain understanding of the role of entropy regularization in such alternating optimization settings.} 

\textbf{Local Cross Entropy Loss:} 
We add a local (non-structured) cross entropy $\mathrm{loss}_{\mathrm{CE}}(\canet(\x_i), \y_i)$ defined for the problem under consideration. We only experiment with this loss for sequence labeling. 
 
It is the sum of the label cross entropy losses over all positions in the sequence. This loss provides more explicit feedback to the inference network, helping the optimization procedure to find a solution that minimizes the energy function while also correctly classifying individual labels. It can also be viewed as a multi-task loss for the inference network.

\textbf{Regularization Toward Pretrained Inference Network:} 
We add the penalty $\| \Phi-\Phi_{0} \|_2^2$ 
where $\Phi_{0}$ is a pretrained network, e.g., a local classifier trained to independently predict each part of $\y$. 

Each additional term has its own tunable hyperparameter. Finally we obtain: 
\begin{align}
\hat{\Phi} \gets \argmax_{\Phi}\: &\left[\bigtriangleup(\canet(\x_i), \y_i) - E_{\Theta}(\x_i,\canet(\x_i)) + E_{\Theta}(\x_i, \y_i)\right]_{+} - \lambda_1 \| \Phi \|_2^2 \nonumber\\
&+ \lambda_2\mathrm{loss}_{\mathrm{H}}(\canet(\x_i)) - \lambda_3\mathrm{loss}_{\mathrm{CE}}(\canet(\x_i), \y_i) - 
\lambda_4\| \Phi-\Phi_{0} \|_2^2\nonumber
\end{align}

\section{Adversarial Training.}
Our training methods are reminiscent of other alternating optimization problems like that underlying generative adversarial networks (GANs; \citealt{goodfellow2014generative, NIPS2016_6125,ZhaoML16,Arjovsky2017WassersteinG}). GANs are based on a minimax game
and have a value function that one agent (a discriminator $D$) seeks to maximize and another (a generator $G$) seeks to minimize.

Progress in training GANs has come largely from overcoming learning difficulties by modifying loss functions and optimization, and GANs have become more successful and popular as a result. 
Notably, Wasserstein GANs \citep{Arjovsky2017WassersteinG} provided the first convergence measure in GAN training using Wasserstein distance. To compute Wasserstein distance, the discriminator uses weight clipping, which limits network capacity. 
Weight clipping was subsequently replaced with a gradient norm constraint~\citep{NIPS2017_7159}. \citet{miyato2018spectral} proposed a novel weight normalization technique called spectral normalization. These methods may be applicable to the similar optimization problems solved in learning SPENs. 
By their analysis, a log loss discriminator converges to a degenerate uniform solution. When using hinge loss, we can get a non-degenerate discriminator while matching the data distribution~\citep{calibrating,ZhaoML16}. Our formulation is closer to this hinge loss version of the GAN.

\section{Results}
In this section, we compare our approach to previous work on traing SPENs.

\subsection{Multi-Label Classification}

\begin{table}[h]
\begin{center}
\begin{tabular}{l|ccc|c}
& Bibtex & Bookmarks & Delicious & avg. \\
\hline
MLP & 38.9 & 33.8 & 37.8 & 36.8 \\
SPEN (BM16) & \bf 42.2 & 34.4 & \bf 37.5 & 38.0 \\
SPEN (E2E) & 38.1 & 33.9 & 34.4 & 35.5 \\
SPEN (InfNet) & \bf 42.2 & \bf 37.6 & \bf 37.5 & \bf 39.1 \\
\end{tabular}
\end{center}
\caption{ Test F1 when comparing methods on multi-label classification datasets.
\label{table:mlc_result}
}
\end{table}

\paragraph{Energy Functions for Multi-label Classification.}
We describe the SPEN for multi-label classification (MLC) from \citet{belanger2016structured}. Here, $\x$ is a fixed-length feature vector. We assume there are $L$ labels, each of which can be on or off for each input, so $\yspace(\x) = \{0,1\}^L$ for all $\x$. 
The energy function is the sum of two terms: $E_{\Theta}(\x,\y) = E^{\mlocal}(\x,\y) + E^{\mlabel}(\y)$. 
$E^{\mlocal}(\x,\y)$ is the sum of linear models:
\begin{align}
E^{\mlocal}(\x,\y)= \sum_{i=1}^L y_{i}b_{i}^\top F(\boldsymbol{x})
\end{align}
where $b_{i}$ is a parameter vector for label $i$ and $F(\x)$ is a multi-layer perceptron computing a feature representation for the input $\x$. 
$E^{\mlabel}(\y)$ scores $\y$ independent of $\x$:
\begin{align}
E^{\mlabel}(\y) = c_{2}^\top g(C_{1}\y)
\end{align}

where $c_2$ is a parameter vector, $g$ is an elementwise non-linearity function, and $C_1$ is a parameter matrix. 



\begin{table}[h!]
\begin{center}
\begin{tabular}{c|c|c|c|c|c}
\multicolumn{1}{c|}{} & \# labels & \# features & \# train  & \# dev & \# test\\
\hline 
Bibtex  & 159  & 1836 & 4836 & - & 2515
\\ 
Bookmarks & 208  & 2151 & 48000 & 12000 & 27856
\\ 
Delicious & 982  & 501 & 12896 & - & 3185
\\ 
\end{tabular}
\end{center}
\caption{Statistics of the multi-label classification datasets.}
\label{table:datasize}
\end{table}

\textbf{Datasets.} Table~\ref{table:datasize} shows dataset statistics for the multi-label classification datasets. The dataset is available at \url{https://davidbelanger.github.io/icml_mlc_data.tar.gz}, which is provides by ~\citet{belanger2016structured}.

\textbf{Hyperparameter Tuning.} 
We tune $\lambda$ (the $L_2$ regularization strength for $\Theta$) over the set $\{ 0.01, 0.001, 0.0001\}$. The classification threshold $\tau$ is chosen from: \\ $[0, 0.01, 0.02, 0.03, 0.04,0.05,0.1,0.15,0.2,0.25,0.3,0.35,0.4,0.45,0.5,0.55,0.6,0.65,0.7,0.75 ]$ as also done by \citet{belanger2016structured}. We tune the coefficients for the three stabilization terms for the inference network objective over the follow ranges: $L_2$ regularization ($\lambda_1\in\{ 0.01, 0.001, 0.0001\}$), entropy regularization ($\lambda_2 = 1$), and regularization toward the pretrained feature network ($\lambda_4\in\{ 0, 1, 10\}$).

\begin{table}[t]
\begin{center}
\begin{tabular}{l|cc}
 hinge loss & -retuning & +retuning \\ 
\hline
 margin rescaling         & 38.51 & 38.68 \\
 slack rescaling & 38.57 & 38.62 \\
 perceptron (MR, $\cost=0$)  & 38.55 & 38.70  \\
 contrastive ($\cost =1$) & 38.80  & 38.88 \\
\end{tabular}
\end{center}
\caption{ Development F1 for Bookmarks when comparing hinge losses for SPEN (InfNet) and whether to retune the inference network.}
\label{table:final_inference}
\end{table}

\textbf{Comparison of Loss Functions and Impact of Inference Network Retuning.} 
Table~\ref{table:final_inference} shows results comparing the four loss functions from Section~\ref{sec:summary-losses} on the development set for Bookmarks, the largest of the three datasets. We find performance to be highly similar across the losses, with the contrastive loss appearing slightly better than the others.

After training, we ``retune'' the inference network as specified by Eq.~\eqref{eq:finalnet} on the development set for 20 epochs using a smaller learning rate of 0.00001. 

Table~\ref{table:final_inference} shows slightly higher F1 for all losses with retuning. We were surprised to see that the final cost-augmented inference network performs well as a test-time inference network. This suggests that by the end of training, the cost-augmented network may be approaching the $\argmin$
and that there may not be much need for retuning. 

When using $\cost=0$ or 1, retuning leads to the same small gain as when using the margin-rescaled or slack-rescaled losses. Here the gain is presumably from adjusting the inference network for other inputs rather than from converting it from a cost-augmented to a test-time inference network.

\paragraph{Performance Comparison to Prior Work.} Table~\ref{table:mlc_result} shows results comparing to prior work. 
The MLP and ``SPEN (BM16)'' baseline results are taken from \citep{belanger2016structured}. 
We obtained the ``SPEN (E2E)''~\citep{End-to-EndSPEN} results by running the code available from the authors on these datasets. This method constructs a recurrent neural network that performs gradient-based minimization of the energy with respect to $\y$. They noted in their software release that, while this method is more stable, it is prone to overfitting and actually performs worse than the original SPEN. We indeed find this to be the case, as SPEN (E2E) underperforms SPEN (BM16) on all three datasets. 

Our method (``SPEN (InfNet)'') achieves the best average performance across the three datasets. It performs especially well on Bookmarks, which is the largest of the three. 
Our results use the contrastive hinge loss and retune the inference network on the development data after the energy is trained; these decisions were made based on the tuning, but all four hinge losses led to similarly strong results. 

\begin{table}[h]
\centering
\begin{tabular}{l|ccc|ccc}
& \multicolumn{3}{|c}{Training Speed (examples/sec)} & \multicolumn{3}{|c}{Testing Speed (examples/sec)} \\
& Bibtex & Bookmarks & Delicious & Bibtex & Bookmarks & Delicious \\
\hline
MLP & 21670 & 19591 &  26158 & 90706  & 92307 & 113750 \\
SPEN (E2E) & 551 & 559 & 383 & 1420 & 1401 & 832 \\
SPEN (InfNet) &5533 & 5467 & 4667 & 94194 & 88888 & 112148  \\
\end{tabular}
\caption{Training and test-time inference speed comparison (examples/sec). 
\label{table:train_test_time_result}
}
\end{table}

\paragraph{Speed Comparison.} Table~\ref{table:train_test_time_result} compares training and test-time inference speed among the different methods. We only report speeds of methods that we ran.\footnote{The MLP F1 scores above were taken from \citet{belanger2016structured}, but the MLP timing results reported in Table~\ref{table:train_test_time_result} are from our own experimental replication of their results.} 
The SPEN (E2E) times were obtained using code obtained from Belanger and McCallum. We suspect that SPEN (BM16) training would be comparable to or slower than SPEN (E2E).
 
Our method can process examples during training about 10 times as fast as the end-to-end SPEN, and 60-130 times as fast during test-time inference. In fact, at test time, our method is roughly the same speed as the MLP baseline, since our inference networks use the same architecture as the feature networks which form the MLP baseline. 
Compared to the MLP, the training of our method takes significantly more time overall because of joint training of the energy function and inference network, but fortunately the test-time inference is comparable.

\subsection{Sequence Labeling}

\paragraph{Energy Functions for Sequence Labeling.}

For sequence labeling tasks, given an input sequence $\x = \langle x_1, x_2,..., x_{|\x|}\rangle$, we wish to output a discrete sequence. In Equation~\ref{eq:crf-energy}, the energy function only permits discrete $\y$. For the general case that permits relaxing $\y$ to be \textbf{continuous}, we treat each $y_t$ as a vector. It will be one-hot for the ground truth $\y$ and will be a vector of label probabilities for relaxed $\y$'s. Then the general energy function :
\begin{align}
E_{\Theta}(\x, \y) = -\left(\sum_{t} \sum_{i=1}^L y_{t,i} \left(U_i^\top f(\x,t)\right) + \sum_{t}\y_{t-1}^\top W \y_{t}\right) \nonumber
\end{align}
where $y_{t,i}$ is the $i$th entry of the vector $y_t$. In the discrete case, this entry is 1 for a single $i$ and 0 for all others, so this energy reduces to Eq.~\eqref{eq:crf-energy} in that case. In the continuous case, this scalar indicates the probability of the $t$th position being labeled with label $i$. 

For the label pair terms in this general energy function, we use a bilinear product between the vectors $\y_{t-1}$ and $\y_t$ using parameter matrix $W$, which also reduces to the discrete version when they are one-hot vectors.

\paragraph{Experimental Setup}

For Twitter part-of-speech (POS) tagging, we use the annotated data from \citet{gimpel-etal-2011-part} and \citet{owoputi-etal-2013-improved} which contains
$L=25$ POS tags. For training, we combine the 1000-tweet \textsc{Oct27Train} set and the 327-tweet  \textsc{Oct27Dev}  set. For validation, we use the 500-tweet \textsc{Oct27Test} set and for testing we use the 547-tweet \textsc{Daily547} test set. We use 100-dimensional skip-gram embeddings trained on 56 million English tweets with \texttt{word2vec}~\citep{mikolov2013distributed}.\footnote{The pretrained embeddings are the same as those used by \citet{tu-17-long} and are available at \url{http://ttic.uchicago.edu/~lifu/}}

We use a BLSTM to compute the ``input feature vector'' $f(\x, t)$ for each position $t$, using hidden vectors of dimensionality $d=100$. 
We also use BLSTMs for the inference networks. 
The output layer of the inference network is a softmax function, so at every position, the inference network produces a distribution over labels at that position. 
We train inference networks using stochastic gradient descent  (SGD) with momentum and train the energy parameters using Adam. For $\cost$, we use $L_1$ distance. We tune hyperparameters on the validation set; full details of tuning are provided in the appendix. We found that the cross entropy stabilization term worked well for this setting.

We compare to standard BLSTM and CRF baselines. 
We train the BLSTM baseline to minimize per-token log loss; this is often called a ``BLSTM tagger''. We train a CRF baseline using the energy in Eq.~\eqref{eq:crf-energy} with the standard conditional log-likelihood objective using the standard dynamic programming algorithms (forward-backward) to compute gradients during training. Further details are provided in the appendix.

\begin{table}[h]
\begin{center}
\begin{tabular}{l|cc}
  & \multicolumn{2}{|c}{validation accuracy (\%)}  \\
 inference network stabilization terms & -retuning & +retuning \\
\hline
 cross entropy & 89.1 & 89.3 \\
 entropy & 84.2 & 86.8 \\
\end{tabular}
\end{center}
\caption{Comparison of inference network stabilization terms and showing impact of retuning when training SPENs with margin-rescaled hinge (Twitter POS validation accuracies).} 
\label{table:pos_inference_term_test}
\end{table}

\paragraph{Hyperparameter Tuning}

When training inference networks and SPENs for Twitter POS tagging, we use the following hyperparameter tuning. 

We tune the inference network learning rate ($\{0.1, 0.05, 0.02, 0.01, 0.005, 0.001\}$), $L_2$ regularization ($\lambda_1\in\{ 0, 1\mathrm{e}\! -\! 3, 1\mathrm{e}\!-\! 4, 1\mathrm{e}\! -\! 5,1\mathrm{e}\! -\! 6, 1\mathrm{e}\! -\! 7\}$), the entropy regularization term ($\lambda_2 \in \{ 0.1, 0.5,1,2,5,10\}$), the cross entropy regularization term ($\lambda_3\in\{0.1, 0.5,1,2,5,10\}$), and the squared L2 distance ($\lambda_4\in\{0, 0.1, 0.2, 0.5, 1, 2, 10\}$).  
We train the energy functions with Adam with a learning rate of 0.001 and $L_2$ regularization ($\lambda_1\in\{ 0, 1\mathrm{e}\! -\! 3, 1\mathrm{e}\! -\! 4, 1\mathrm{e}\! -\! 5,1\mathrm{e}\! -\! 6, 1\mathrm{e}\! -\! 7\}$). 

Table~\ref{table:pos_inference_term_test} compares the use of the cross entropy and entropy stabilization terms when training inference networks for a SPEN with margin-rescaled hinge. Cross entropy works better than entropy in this setting, though retuning permits the latter  to bridge the gap more than halfway. 

When training CRFs, we use SGD with momentum.

We tune the learning rate (over $\{0.1, 0.05, 0.02, 0.01, 0.005, 0.001\}$) and $L_2$ regularization coefficient (over $\{0, 1\mathrm{e}\! -\! 3, 1\mathrm{e}\! -\! 4, 1\mathrm{e}\! -\! 5, 1\mathrm{e}\! -\! 6, 1\mathrm{e}\! -\! 7\}$). 
For all methods, we use early stopping based on validation accuracy. 

\begin{table}[h]
\begin{center}
\begin{tabular}{l|cc}
		& \multicolumn{2}{|c}{validation accuracy (\%)} \\ 
       SPEN hinge loss & -retuning & +retuning \\
\hline
margin rescaling & 89.1 & 89.3 \\
slack rescaling & 89.4 & 89.6 \\
perceptron (MR, $\cost = 0$) & 89.2 & 89.4 \\
contrastive ($\cost = 1$) & 88.8 & 89.0 \\
\end{tabular}
\end{center}
\caption{Comparison of SPEN hinge losses and showing the impact of retuning (Twitter POS validation accuracies). Inference networks are trained with the cross entropy term.
\label{table:pos_inference_margin_test}
} 
\end{table}

\paragraph{Learned Pairwise Potential Matrix}
\begin{figure}[ht]
\centering
\vspace{-0.5cm}
\includegraphics[width=1.0\textwidth]{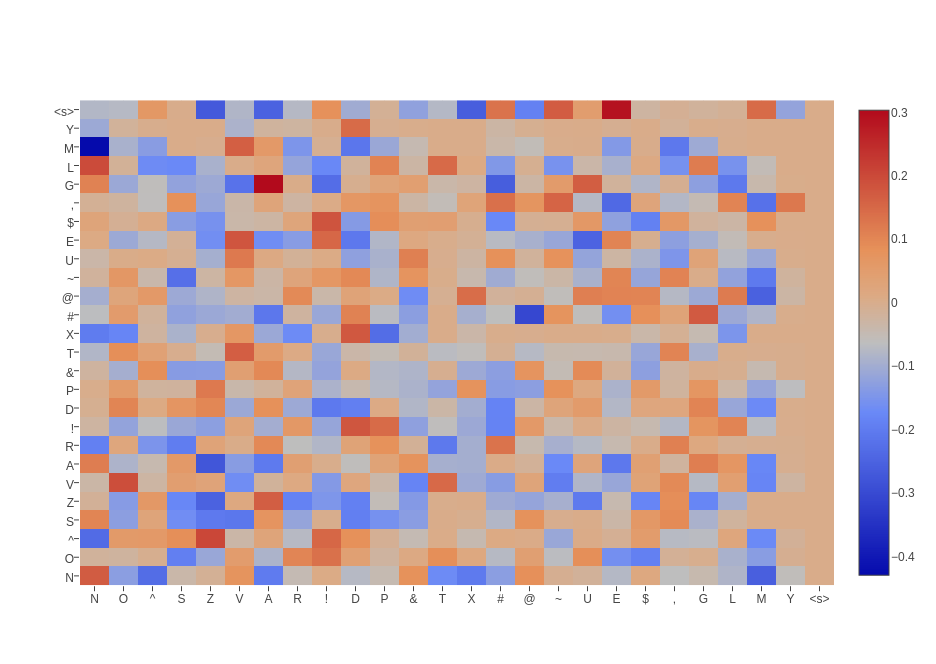}
\vspace{-1.4cm}
\caption{Learned pairwise potential matrix for Twitter POS tagging. 
\label{fig:heatmap}}
\end{figure}

Figure~\ref{fig:heatmap} shows the learned pairwise potential matrix $W$ in Twitter POS tagging. We can see strong correlations between labels in neighborhoods. For example, an adjective (A) is more likely to be followed by a noun (N) than a verb (V) (see row labeled ``A'' in the figure).

\paragraph{Loss Function Comparison.}
Table~\ref{table:pos_inference_margin_test} shows results when comparing SPEN training objectives. 
We see a larger difference among losses here than for MLC tasks. When using the perceptron loss, there is no margin, which leads to overfitting: 89.4 on validation, 88.6 on test (not shown in the table). The contrastive loss, which strives to achieve a margin of 1, does better on test (89.0). 
We also see here that margin rescaling and slack rescaling  both outperform the contrastive hinge, unlike the MLC tasks. We suspect that in the case in which each input/output has a different length, using a cost that captures length is more important.

\begin{table}[h]
\begin{center}
\begin{tabular}{l|cc|cc}
 & validation & test & training speed & testing speed \\
 & accuracy (\%) & accuracy (\%) & (examples/sec) & (examples/sec) \\
\hline
BLSTM  & 88.6 & 88.8 & 385 & 1250 \\
CRF & 89.1 & 89.2 & 250 & 500 \\
SPEN (InfNet) & 89.6 & 89.8 & 125 & 1250 \\
\end{tabular}
\end{center}
\caption{Twitter POS accuracies of BLSTM, CRF, and SPEN (InfNet), using our tuned SPEN configuration (slack-rescaled hinge, inference network trained with cross entropy term). Though slowest to train, the SPEN matches the test-time speed of the BLSTM while achieving the highest accuracies. \label{table:twitter-baselines}}
\end{table}

\paragraph{Comparison to Standard Baselines.} 
Table~\ref{table:twitter-baselines} compares our final tuned SPEN configuration to two standard baselines: a BLSTM tagger and a CRF. The SPEN achieves higher validation and test accuracies with faster test-time inference. 
While our method is slower than the baselines during training, it is faster than the CRF at test time, operating at essentially the same speed as the BLSTM baseline while being more accurate. 

\subsection{Tag Language Model}

The above results only use the pairwise energy. In order to capture long-distance dependencies in an entire sequence of labels, 
we define an additional energy term $E^{\mathrm{TLM}}(\y)$ based on the pretrained TLM. If the argument $\y$ consisted of one-hot vectors, we could simply compute its likelihood. However, to support relaxed $\y$'s, we need to define a more general function: 
\begin{equation}
E^{\mathrm{TLM}}(\y) = - \sum_{t=1}^{|\y|+1} \log(y_t^\top \LM(\langle y_0,...,y_{t-1}\rangle) )
\label{eq:tlm}
\end{equation}

\noindent where $y_0$ is the start-of-sequence symbol, $y_{|\y|+1}$ is the end-of-sequence symbol, and $\LM(\langle y_0,...,y_{t-1}\rangle)$ returns the softmax distribution over tags at position $t$ (under the pretrained tag language model) given the preceding tag vectors. When each $y_t$ is a one-hot vector, this energy reduces to the negative log-likelihood of the tag sequence specified by $\y$. 

\begin{table}[h]
\begin{center}
\begin{tabular}{l|cc}

 \multicolumn{1}{l|}{} & val. accuracy (\%) & test accuracy (\%) \\
\hline
 -TLM 
 & 89.8 & 89.6 \\
 +TLM 
 & 89.9 & 90.2 \\
\end{tabular}
\caption{ Twitter POS validation/test accuracies when adding tag language model (TLM) energy term to a SPEN trained with margin-rescaled hinge.
\label{table:tlm-results}
}
\end{center}
\end{table}

We define the new joint energy as the sum of the energy function in Eq.~\eqref{eq:crf} and the TLM energy function in Eq.~\eqref{eq:tlm}. During learning, we keep the TLM parameters fixed to their pretrained values, but we tune the weight of the TLM energy (over the set $\{0.1, 0.2, 0.5\}$) in the joint energy. 
We train SPENs with the new joint energy using the margin-rescaled hinge, training the inference network with the cross entropy term. 

\paragraph{Setup}

To compute the TLM energy term, we first automatically tag unlabeled tweets, then train an LSTM language model on the automatic tag sequences. When doing so, we define the input tag embeddings to be $L$-dimensional one-hot vectors specifying the tags in the training sequences. This is nonstandard compared to standard language modeling. In standard language modeling, we train on observed sequences and compute likelihoods of other fully-observed sequences. However, in our case, we train on tag sequences but we want to use the same model on sequences of tag \emph{distributions} produced by an inference network. We train the TLM on sequences of one-hot vectors and then use it to compute likelihoods of sequences of tag distributions. 

To obtain training data for training the tag language model, we run the Twitter POS tagger from \citet{owoputi-etal-2013-improved} on a dataset of 303K randomly-sampled English tweets. We train the tag language model on 300K tweets and use the remaining 3K for tuning hyperparameters and early stopping. 
We train an LSTM language model on the tag sequences using stochastic gradient descent with momentum and early stopping on the validation set. We used a dropout rate of 0.5 for the LSTM hidden layer. We tune the learning rate ($\{ 0.1,0.2, 0.5, 1.0\}$), the number of LSTM layers ($\{ 1,2\}$), and the hidden layer size ($\{ 50,100,200\}$).

\paragraph{Results}

Table~\ref{table:tlm-results} shows results.\footnote{The baseline results differ slightly from earlier results because we found that we could achieve higher accuracies in SPEN training by avoiding using pretrained feature network parameters for the inference network.}
Adding the TLM energy leads to a gain of 0.6 on the test set. 
Other settings showed more variance; 
when using slack-rescaled hinge, we found a small drop on test, 
while when simply training inference networks for a fixed, pretrained joint energy with tuned mixture coefficient, 
we found a gain of 0.3 on test when adding the TLM energy. 
We investigated the improvements and found some to involve corrections that seemingly stem from handling non-local dependencies better.

\begin{table}[h]
\begin{center}
\begin{tabular}{c|p{8.7cm}|c|c}
 \multicolumn{2}{c}{} & \multicolumn{2}{|c}{predicted tags} \\
\# & tweet (target word in bold) & -TLM & +TLM \\
\hline
1 & 
... that's a t-17 , technically . does \textbf{that} count as top-25 ?
 & determiner & pronoun \\
 \hline
2 & ... lol you know im down \textbf{like} 4 flats on a cadillac ... lol ...
& adjective & preposition \\
3 & 
... them who he is : 
he wants her to \textbf{like} him for his pers ... & preposition & verb \\
\hline
4 & I wonder when Nic Cage is going to \textbf{film} " Another Something Something Las Vegas " . & noun & verb \\
5 & Cut my hair , \textbf{gag} and bore me & noun & verb \\
\hline
6 & ... they had their fun , we \textbf{hd} ours ! ;) lmaooo & proper noun & verb \\
7 & " Logic will get you from A to \textbf{B} . Imagination will take you everywhere . " - Albert Einstein . & verb & noun \\
8 & 
lmao I'm not a sheep who listens to it \textbf{cos} everyone else does ... 
& verb & preposition \\
9 & 
Noo its not cuss you have swag \textbf{andd} you wont look dumb ! ... 
& noun & coord. conj. \\
\end{tabular}
\end{center}
\caption{Examples of improvements in Twitter POS tagging when using tag language model (TLM). In all of these examples, the predicted tag when using the TLM matches the gold standard.
\label{table:tag-qual}
}
\end{table}

Table~\ref{table:tag-qual} shows examples in which our SPEN that includes the TLM appears to be using broader context when making tagging decisions. These are examples from the test set labeled by two models: the SPEN without the TLM (which achieves 89.6\% accuracy, as shown in Table~\ref{table:tlm-results}) and the SPEN with the TLM (which reaches 90.2\% accuracy). In example 1, the token ``that'' is predicted to be a determiner based on local context, but is correctly labeled a pronoun when using the TLM. This example is difficult because of the noun/verb tag ambiguity of the next word (``count'') and its impact on the tag for ``that''. 
Examples 2 and 3 show two corrections for the token ``like'', which is a highly ambiguous word in Twitter POS tagging. The broader context makes it much clearer which tag is intended. 

The next two examples (4 and 5) are cases of noun/verb ambiguity that are resolvable with larger context. 
The last four examples show improvements for nonstandard word forms. The shortened form of ``had'' (example 6) is difficult to tag due to its collision with ``HD'' (high-definition), but the model with the TLM is able to tag it correctly. In example 7, the ambiguous token ``b'' is frequently used as a short form of ``be'' on Twitter, and since it comes after ``to'' in this context, the verb interpretation is encouraged. However, the broader context makes it clear that it is not a verb and the TLM-enriched model tags it correctly. The words in the last two examples are nonstandard word forms that were not observed in the training data, which is likely the reason for their erroneous predictions. When using the TLM, we can better handle these rare forms based on the broader context. These results suggest that our method of training inference networks can be used to add rich  features to structured prediction, though we leave a thorough exploration of global energies to future work.

\section{CONCLUSIONS}
We presented ways to jointly train structured energy functions and inference networks using large-margin objectives. The energy function captures arbitrary dependencies among the labels, while theinference networks learns to capture the properties of the energy in an efficient manner, yielding fasttest-time inference. Future work includes exploring the space of network architectures for inferencenetworks to balance accuracy and efficiency, experimenting with additional global terms in structuredenergy functions, and exploring richer structured output spaces such as trees and sentences.
\newpage

\mychapter{6}{Joint 
Parameterizations for Inference Networks}
\label{sec:jointobjective}
In the previous chapter, we develop an efficient framework for energy-based models by training ``inference networks'' to approximate structured inference instead of using gradient descent. However, their alternating optimization approach suffers from instabilities during training, requiring additional loss terms and careful hyperparameter tuning. In this paper, we contribute several strategies to stabilize and improve this joint training of energy functions and inference networks for structured prediction. We design a compound objective to jointly train both cost-augmented and test-time inference networks along with the energy function. We propose joint parameterizations for the inference networks that encourage them to capture complementary functionality during learning. We empirically validate our strategies on two sequence labeling tasks, showing easier paths to strong performance than prior work, as well as further improvements with global energy terms.

This chapter includes some material originally presented in ~\citet{tu-etal-2020-improving}.

\section{Previous Pipeline}
In the previous chapter, we jointly train the cost-augmented inference network and energy network, then do fine-tuning of the cost-augmented inference network to make it more like a test-time inference network. In our previous work, there are two steps in order to get the test-time inference network $\infnet(\x)$.

\paragraph{Step 1:} 
\begin{align}
    \hat{\Theta}, \hat{\Phi} = \min_{\Theta} \max_{\Phi} \sum_{\langle \x_i, \y_i\rangle\in\mathcal{D}}  [ \cost(\canet(\x_i), \y_i)  - E_{\Theta}(\x_i,\canet(\x_i))  + E_{\Theta}(\x_i, \y_i) ]_{+} \nonumber
\end{align}

Update $\Phi$ to yield output with low energy and high cost 
    
\paragraph{Step 2:}: 
\begin{align}
    \hat{\Psi} = \argmin_{\Psi} E_\Theta(\x, \infnet(\x)) \nonumber
\end{align}
 where $\infnet$ is initialized by trained $\canet$.

\section{An Objective for Joint Learning of Inference Networks}
In this section, we propose a different loss that separates the two inference networks and trains them jointly:
\begin{align}
&\min_{\Theta} \frac{\lambda}{n} 
\sum_{i=1}^{n} \left[ \max_{\y}( - E_{\Theta}(\x_i,\y) + E_{\Theta}(\x_i, \y_i))\right]_{+} \!\!+ 
\frac{1}{n} \!\sum_{i=1}^{n} \!\left[\max_{\y}  (\cost(\y, \y_i) \!-\! E_{\Theta}(\x_i,\y) 
\!+\! E_{\Theta}(\x_i, \y_i))\!\right]_{+} \nonumber
\end{align}

\noindent The above objective contains two different inference problems, which are also the two inference problems that must be solved in structured max-margin learning, whether during training or during test-time inference. Eq.~(\ref{eq:inf}) shows the test-time inference problem. The other one is cost-augmented inference, defined as follows:
\begin{align}
\argmin_{\y'\in\yspace(\x)} (E_\Theta(\x, \y)-\cost(\y', \y))\label{eq:cost1}
\end{align}

This inference problem involves finding an output with low energy but high cost relative to the gold standard output. Thus, it is not well-aligned with the test-time inference problem. 
In Chapter 5, we used the same inference network for solving both problems, which led them to have to perform fine-tuning at test-time with a different objective. We avoid this issue by instead jointly training two inference networks, one for cost-augmented inference and the other for test-time inference:

\begin{align}
&\min_{\Theta} \max_{\Phi, \Psi}\sum_{\langle \x_i, \y_i\rangle\in\mathcal{D}} \nonumber \\ & \underbrace{[\cost(\canet(\x), \y_i) \!-\! E_{\Theta}(\x_i,\canet(\x)) + E_{\Theta}(\x_i, \y_i)]_{+}}_{\text{margin-rescaled loss}} 
+ \lambda \underbrace{\left[- E_{\Theta}(\x_i,\infnet(\x_i)) + E_{\Theta}(\x_i, \y_i)\right]_{+}}_{\text{perceptron loss}} \label{eq:margin-hinge} 
\end{align}

We treat this optimization problem as a minmax game and find a saddle point for the game similar to Chapter 5 and \citet{goodfellow2014generative}. We alternatively optimize $\Theta$, $\Phi$ and $\Psi$.   

We drop the zero truncation ($\max(0,.)$) when updating the inference network parameters to improve stability during training. This also lets us remove the terms that do not have inference networks.

When we remove the truncation at 0, the objective for the inference network parameters is:
\begin{align}
\begin{split}
\hat{\Psi}, \hat{\Phi} \gets \argmax_{\Psi,\Phi}  \cost(\canet(\x), \y_i) \!-\! E_{\Theta}(\x_i,\canet(\x))
- \lambda E_{\Theta}(\x_i,\infnet(\x_i))
\end{split}\nonumber
\end{align}
\noindent The objective for the energy function is:
\begin{align}
\small
\begin{split}
\hat{\Theta} \gets \argmin_{\Theta} \big[\cost(\canet(\x), \y_i) \!-\! E_{\Theta}(\x_i,\canet(\x)) + E_{\Theta}(\x_i, \y_i)\big]_{+}  + \lambda \big[- E_{\Theta}(\x_i,\infnet(\x_i)) + E_{\Theta}(\x_i, \y_i)\big]_{+} \nonumber
\end{split}
\end{align}

\begin{figure*}[!h]
\centering
\includegraphics[width=0.95\textwidth]{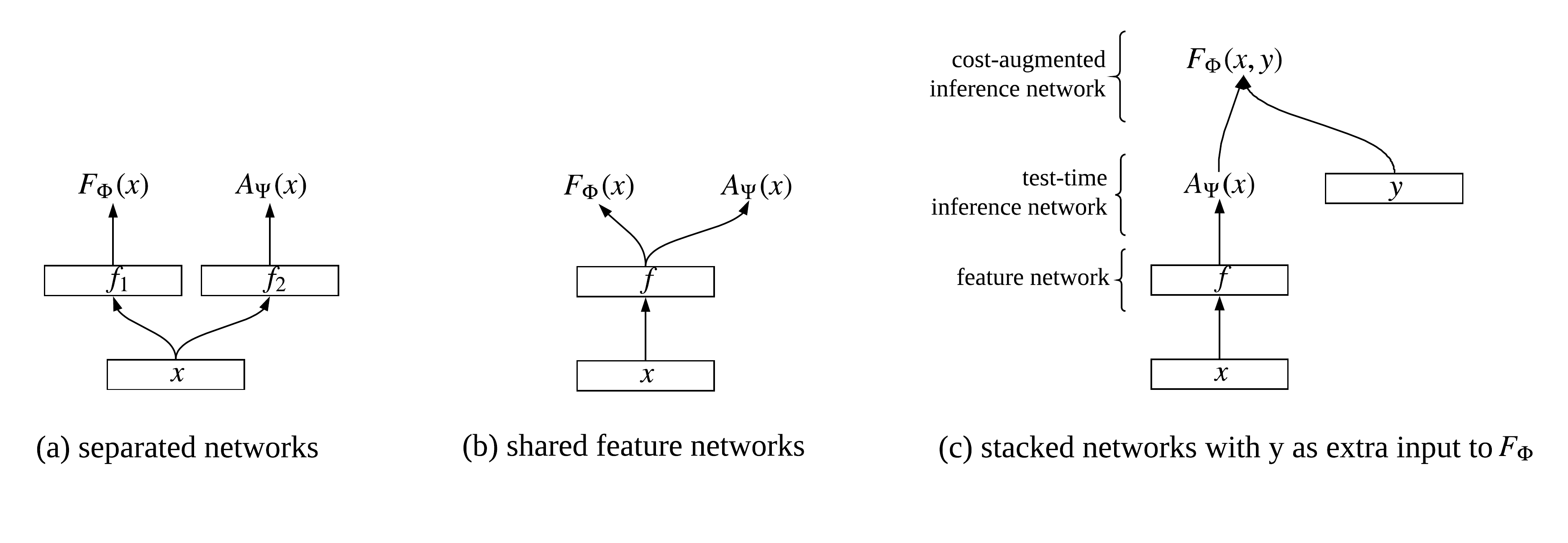}
\vspace{-0.6cm}
\caption{
Parameterizations for cost-augmented inference network $\canet$ and test-time inference network $\infnet$. \label{fig:infnets}
}
\end{figure*}

The new objective jointly trains the energy function $E_{\Theta}$, cost-augmented inference network $\canet$, and test-time inference network $\infnet$. 
This objective offers us several options for defining joint parameterizations of the two inference networks. 

We consider three options which are visualized in Figure~\ref{fig:infnets} and described below:
\begin{itemize}
  \item (a) Separated: $\canet$ and $\infnet$ are two independent networks with their own architectures and parameters as shown in Figure~\ref{fig:infnets}(a).
  \item (b) Shared: $\canet$ and $\infnet$ share the feature network as shown in Figure \ref{fig:infnets}(b). We consider this option because both $\canet$ and $\infnet$ are trained to produce output labels with low energy. However $\canet$ also needs to produce output labels with high cost $\cost$ (i.e., far away from the ground truth). 
  \item (c) Stacked: Here, the cost-augmented network is a function of the output of the test-time inference network and the gold standard output $\y$ is included as an additional input to the cost-augmented network. That is, $\canet=f(\infnet(\x),\y)$ where $f$ is a parameterized function. This is depicted in Figure~\ref{fig:infnets}(c). 
  Note that we block the gradient at $\infnet$ when updating $\Psi$. 

\end{itemize}
For the third option, we will consider multiple choices for the function $f$. 
One choice is to use an affine transform on the concatenation of the inference network and the ground truth label: 
\begin{align}
\canet(\x, \y)_i = \softmax(W[\infnet(\x)_i;\y_i]+b) \nonumber
\end{align}
where semicolon (;) denotes vertical concatenation, $L$ is the label set size, $\y_i \in \mathbb{R}^L$ (position $i$ of $\y$) is a one-hot vector, $\infnet(\x)_i$ and $\canet(\x)_i$ are position $i$ of $\infnet$ and $\canet$, and 
$W$ is a $2L$ by $L$ parameter matrix.
Another choice of $f$ is a BiLSTM: 
\begin{align}
\canet(\x,\y)_i = \BLSTM([\infnet(\x);\y]) \nonumber
\end{align}
\noindent We could have $\y$ as input to the other architectures, but we limit our search to these three options. One motivation for these parameterizations is to reduce the total number of parameters in the procedure. Generally, the number of parameters is expected to decrease when moving from option (a) to (b), and when moving from (b) to (c). We will compare the three options empirically in our experiments, in terms of both accuracy and number of parameters. 

Another motivation, specifically for the third option, is to distinguish the two inference networks in terms of their learned functionality. With all three parameterizations, the cost-augmented network will be trained to produce an output that differs from the ground truth, due to the presence of the $\cost(\canet(\x), \y_i)$ term. However, in Chapter 5, we found that the trained cost-augmented network was barely affected by fine-tuning for the test-time inference objective. This suggests that the cost-augmented network was mostly acting as a test-time inference network by the time of convergence. With the third parameterization above, however, we explicitly provide the ground truth output $\y$ to the cost-augmented network, permitting it to learn to change the predictions of the test-time network in appropriate ways to improve the energy function. We will explore this effect quantitatively and qualitatively below in our experiments.

\section{Training Stability and Effectiveness}\label{sec:techniques}

We now discuss several methods that simplify and stabilize training SPENs with inference networks. When describing them, we will illustrate their impact by showing training trajectories for the Twitter part-of-speech tagging task. 
\begin{figure}[t]
  \centering
  \begin{subfigure}[b]{0.35\linewidth}
    \includegraphics[width=\linewidth]{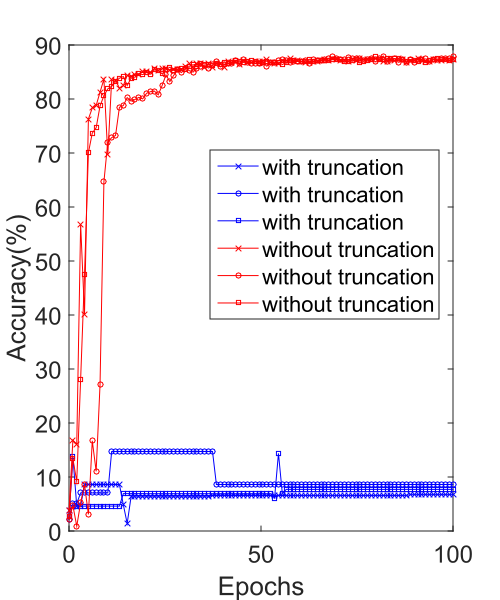}
    \caption{Truncating at 0 (without CE).}
  \end{subfigure}\quad\quad\quad \quad\quad
    \begin{subfigure}[b]{0.35\linewidth}
    \includegraphics[width=\linewidth]{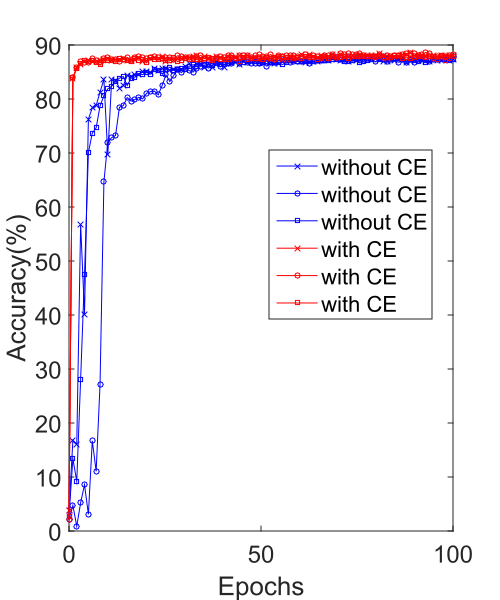}
    \caption{Adding CE loss (without truncation).}
  \end{subfigure}
  \caption{Part-of-speech tagging training trajectories. The three curves in each setting correspond to different random seeds. 
  (a) Without the local CE loss, training fails when using zero truncation. (b) The CE loss reduces the number of epochs for training. 
  In the previous work, we always use zero truncation and CE during training. 
  \label{fig:trunc}
  }

\end{figure}

\subsection{Removing Zero Truncation} 
\label{sec:truncation}

\citet{tu-18} used the following objective for the cost-augmented inference network (maximizing it with respect to $\Phi$): $l_0 =$
\begin{align}
[\cost(\canet(\x), \y) - E_{\Theta}(\x,\canet(\x)) + E_{\Theta}(\x, \y)]_{+} \nonumber
\end{align}
\noindent where $[h]_+ = \max(0,h)$. 
However,  
there are two potential reasons why $l_0$ will equal zero and trigger no gradient update. First, $E_\Theta$ (the energy function, corresponding to the  discriminator in a GAN) may already be well-trained, and it can easily separate the gold standard output from the cost-augmented inference network output. Second, the cost-augmented inference network (corresponding to the generator in a GAN) could be so poorly trained that the energy of its output is very large, leading the margin constraints to be satisfied and $l_0$ to be zero. 

In standard margin-rescaled max-margin learning in structured prediction~\citep{m3,ssvm}, the cost-augmented inference step is performed exactly (or approximately with reasonable guarantee of effectiveness), ensuring that when $l_0$ is zero, the energy parameters are well trained. However, in our case, $l_0$ may be zero simply because the cost-augmented inference network is undertrained, which will be the case early in training. 
Then, when using zero truncation, the gradient of the inference network parameters will be 0. This is likely why \citet{tu-18} found it important to add several stabilization terms to the $l_0$ objective.  
We find that by instead removing the truncation, learning stabilizes and becomes less dependent on these additional terms. Note that we retain the truncation at zero when updating the energy parameters $\Theta$. 

As shown in Figure~\ref{fig:trunc}(a), without any stabilization terms and with truncation, the inference network will barely move from its starting point and learning fails overall. 
However, without truncation, the inference network can work well even without any stabilization terms.

\begin{figure}[th]
  \centering
  \begin{subfigure}[b]{0.35\linewidth}
    \includegraphics[width=\linewidth]{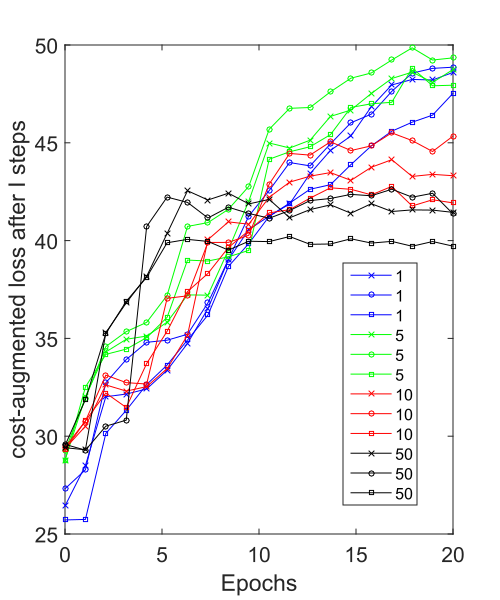}
    \caption{cost-augmented loss $l_1$}
  \end{subfigure}
  \quad\quad\quad
    \begin{subfigure}[b]{0.35\linewidth}
    \includegraphics[width=\linewidth]{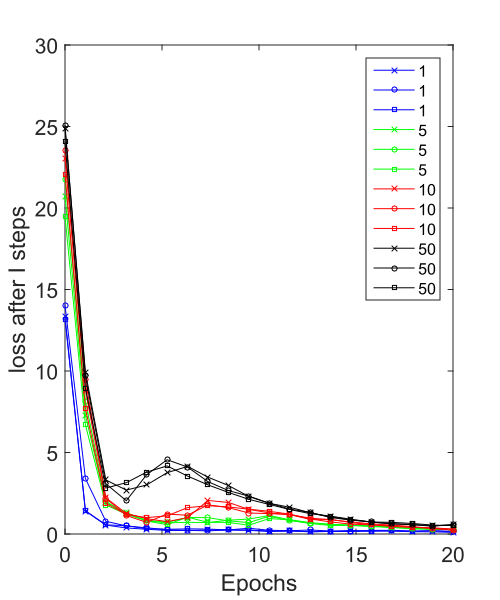}
    \caption{margin-rescaled loss $l_0$}
  \end{subfigure}
  \\
  \begin{subfigure}[b]{0.35\linewidth}
    \includegraphics[width=\linewidth]{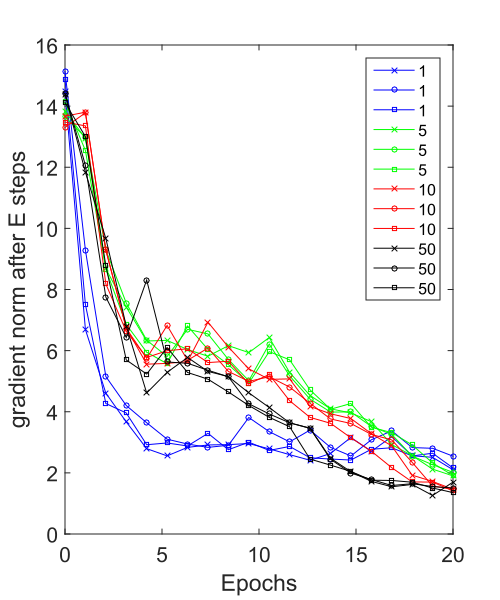}
    \caption{gradient norm of $\Theta$}
  \end{subfigure}
  \quad\quad\quad
  \begin{subfigure}[b]{0.35\linewidth}
    \includegraphics[width=\linewidth]{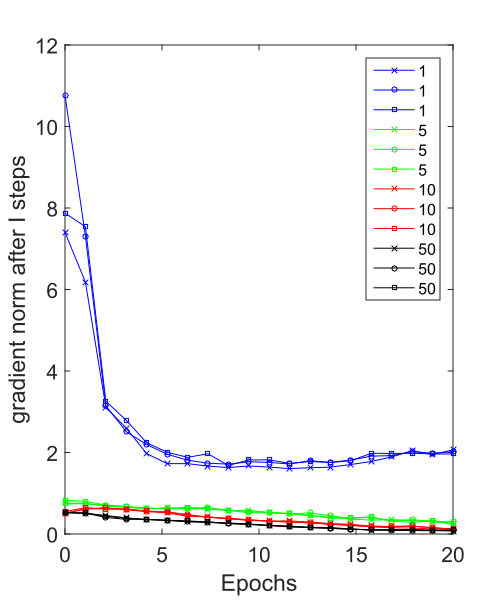}
    \caption{gradient norm of $\Psi$}
  \end{subfigure}
  \caption{
  POS training trajectories with different numbers of I steps. The three curves in each setting correspond to different random seeds. 
  (a)  cost-augmented loss after I steps; (b) margin-rescaled hinge loss after I steps;  
  (c) gradient norm of energy function parameters after E steps; (d) gradient norm of test-time inference network parameters after I steps. 
  \label{fig:setting_mg}}
\end{figure}

\subsection{Local Cross Entropy (CE) Loss}
\label{sec:ce}
\citet{tu-18} proposed adding a local cross entropy (CE) loss, which is the sum of the label cross entropy losses over all positions in the sequence, 
to stabilize inference network training. 
We similarly find this term to help speed up convergence and improve accuracy. 
Figure~\ref{fig:trunc}(b) shows faster convergence to high accuracy when adding the local CE term. See Section~\ref{sec: objective_results} for more details. 

\subsection{Multiple Inference Network Update Steps}
When training SPENs with inference networks, the inference network parameters are nested within the energy function.
We found that the gradient components of the inference network parameters consequently have smaller absolute values than those of the energy function parameters. 
So, we alternate between $k\geq 1$ steps of optimizing the inference network parameters (``I steps'') and one step of optimizing the energy function parameters (``E steps''). We find this strategy especially helpful when using complex inference network architectures. 

To analyze, we compute the 
cost-augmented loss $l_1 = \cost(\canet(\x), \y) - E_{\Theta}(\x,\canet(\x))$ 
and the margin-rescaled hinge loss $l_0 = [\cost(\canet(\x), \y) - E_{\Theta}(\x,\canet(\x)) + E_{\Theta}(\x, \y)]_{+}$ 
averaged over all training pairs $(\x, \y)$ after each set of I steps. 
The I steps update $\Psi$ and $\Phi$ to maximize these  losses. Meanwhile the E steps update $\Theta$ to minimize these losses. 
Figs.~\ref{fig:setting_mg}(a) and (b) show $l_1$ and $l_0$  
during training for different numbers ($k$) of I steps for every one E step. 
Fig.~\ref{fig:setting_mg}(c) shows the norm of the energy parameters after the E steps, 
and 
Fig.~\ref{fig:setting_mg}(d) shows the norm of $\frac{\partial E_{\Theta}(\x, \infnet)}{\partial \Psi}$ after the I steps. 

With $k=1$, the setting used by \citet{tu-18}, the inference network lags behind the energy, making the energy parameter updates very small, as shown by the small norms in Fig.~\ref{fig:setting_mg}(c). The inference network  gradient norm (Fig.~\ref{fig:setting_mg}(d)) remains high, indicating underfitting. 
However, increasing $k$ too much also harms learning, as evidenced by the ``plateau'' effect in the $l_1$ curves for $k=50$; this indicates that the energy function is lagging behind the  inference network. 
Using $k=5$ leads to more of a balance between $l_1$ and $l_0$ and gradient norms that are mostly decreasing during training. We treat $k$ as a hyperparameter that is tuned in our experiments. 

There is a potential connection between our use of multiple I steps and a similar procedure used in GANs \citep{goodfellow2014generative}. In the GAN objective, the discriminator $D$ is updated in the inner loop, and they alternate between multiple update steps for $D$ and one update step for $G$. 
In this section, we similarly found benefit from multiple steps of inner loop optimization for every step of the outer loop. However, the analogy is limited, since GAN training involves sampling noise vectors and using them to generate data, while there are no noise vectors or explicitly-generated samples in our framework.

\section{Energies for Sequence Labeling}
\label{sec:tlm-training}
For our sequence labeling experiments in this paper, the input $\x$ is a length-$T$ sequence of tokens, and the output $\y$ is a sequence of labels of length $T$. 
We use $\y_t$ to denote the output label at position $t$, where $\y_t$ is a vector of length $L$ (the number of labels in the label set) and where $y_{t,j}$ is the $j$th entry of the vector $\y_t$. In the original output space $\yspace(\x)$, $y_{t,j}$ is 1 for a single $j$ and 0 for all others. In the relaxed output space $\relyspace(\x)$, $y_{t,j}$  can be interpreted as the probability of the $t$th position being labeled with label $j$. 
We then use the following energy for sequence labeling~\citep{tu-18}: 
\begin{align}
\label{eqn:energy-sequence-labeling}
E_{\Theta}(\x, \y) &= -\Bigg(\sum_{t=1}^T  \sum_{j=1}^L y_{t,j}\! \left(U_j^\top b(\x,t)\right) \nonumber \\
& + \sum_{t=1}^T \y_{t-1}^\top W \y_{t}\Bigg) 
\end{align}
where $U_{j}\in\mathbb{R}^d$ is a parameter vector for label $j$  
and the parameter matrix $W\in\mathbb{R}^{L \times L}$ contains label-pair parameters. 
Also, $b(\x,t)\in\mathbb{R}^d$ denotes the ``input feature vector'' for position $t$. 
We define $b$ to be the $d$-dimensional BiLSTM~\citep{Hochreiter:1997:LSM:1246443.1246450} hidden vector at $t$.
The full set of energy parameters $\Theta$ includes the $U_j$ vectors, $W$, and the parameters of the BiLSTM.

\paragraph{Global Energies for Sequence Labeling.}
In addition to new training strategies, we also experiment with several global energy terms for sequence labeling. 
Eq.~(\ref{eqn:energy-sequence-labeling}) shows the base energy, and to capture long-distance  dependencies, we include global energy (GE) terms in the form of Eq.~(\ref{eqn:basic-tlm}). 

We use $h$ to denote an LSTM tag language model (TLM) that takes a sequence of labels as input and returns a distribution over next labels. 
We define $\overline{\y}_t = {h}(\y_0,\dots,\y_{t-1})$ to be the distribution given the preceding label vectors (under a LSTM language model). Then, the energy term is:

\begin{align}
E^{\mathrm{TLM}}(\y) = - \sum_{t=1}^{T+1} \log \left(\y_t^\top \overline{\y}_t\right)
\label{eqn:basic-tlm}
\end{align}

\noindent 
where $\y_0$ is the start-of-sequence symbol and  $\y_{T+1}$ is the end-of-sequence symbol. 
This energy returns the negative log-likelihood under the TLM of the candidate output $\y$. 
\citet{tu-18} pretrained their $h$ on a large,  automatically-tagged corpus and fixed its parameters when optimizing $\Theta$. 
Our approach has one critical difference. We instead \textit{do not} pretrain $h$, and its parameters are learned when optimizing $\Theta$. We show that even without pretraining, our global energy terms are still able to capture useful additional information.

We also propose new global energy terms. Define $\overline{\y}_t = {h}(\y_0,\dots,\y_{t-1})$ where $h$ is an LSTM TLM that takes a sequence of labels as input and returns a distribution over next labels. 
First, we add a TLM in the backward direction (denoted $\overline{\y}'_t$ analogously to the forward TLM). 
Second, we include words as additional inputs to forward and backward TLMs.  
We define $\widetilde{\y}_t = {g}(\x_0,..., \x_{t-1}, \y_0,...,\y_{t-1})$ where $g$ is a forward LSTM TLM.
We define the backward version similarly (denoted $\widetilde{\y}'_t$). The global energy is therefore
\begin{align}
E^{\mathrm{GE}}(\y) &= - \sum_{t=1}^{T+1} \log (\y_t^\top \overline{\y}_t) +  \log (\y_t^\top \overline{\y}'_t) + \gamma\big(\log (\y_t^\top \widetilde{\y}_t)  + \log (\y_t^\top \widetilde{\y}'_t)\big) 
\label{eqn:stronger-tlm}
\end{align}
Here $\gamma$ is a hyperparameter that is tuned. We experiment with three settings for the global energy: GE(a): forward TLM as in \citet{tu-18}; GE(b): forward and backward TLMs  ($\gamma = 0$); GE(c): all four TLMs in Eq.~(\ref{eqn:stronger-tlm}).

\section{Experimental Setup}
\label{sec:expsetup}

We consider two sequence labeling tasks: Twitter part-of-speech (POS) tagging~\citep{gimpel-etal-2011-part} and named entity recognition (NER;~\citealp{tjong-kim-sang-de-meulder-2003-introduction}).

\paragraph{Twitter Part-of-Speech (POS) Tagging.}

We use the 
Twitter POS data from \citet{gimpel-etal-2011-part} and \citet{owoputi-etal-2013-improved} which contain 25 tags. We use 100-dimensional skip-gram \citep{mikolov2013distributed} embeddings from \citet{tu-17-long}. 
Like \citet{tu-18}, we use a BiLSTM to compute the input feature vector for each position, using hidden size 100. We also use BiLSTMs for the inference networks. The output of the inference network is a softmax function, so the inference network will produce a distribution over labels at each position. The $\Delta$ is L1 distance. We train the inference network using stochastic gradient descent (SGD) with momentum and train the energy parameters using Adam~\citep{adam}. We also explore training the inference network using Adam when not using the local CE loss.\footnote{We find that Adam works better than SGD when training the inference network without the local cross entropy term.} In experiments with the local CE term, its weight is set to 1.

\paragraph{Named Entity Recognition (NER).}

We use the CoNLL 2003 English dataset \citep{tjong-kim-sang-de-meulder-2003-introduction}. We use the BIOES tagging scheme, following previous work \citep{Ratinov:2009},  resulting in 17 NER labels. We use 100-dimensional pretrained GloVe embeddings \citep{pennington-etal-2014-glove}. 
The task is evaluated using F1 score computed with the \texttt{conlleval} script.   
The architectures for the feature networks in the energy function and inference networks are all BiLSTMs. The architectures for tag language models are LSTMs. 
We use a dropout \texttt{keep-prob} of 0.7 for all LSTM cells. The hidden size for all LSTMs is 128. We use Adam~\citep{adam} and do early stopping on the development set. We use a learning rate of $5\cdot 10^{-4}$. Similar to above, the weight for the CE term is set to 1. 

We consider three NER modeling configurations. \textbf{NER} uses only words as input and pretrained, fixed GloVe embeddings. \textbf{NER+} uses words, the case of the first letter, POS tags, and chunk labels, as well as pretrained GloVe embeddings with fine-tuning. \textbf{NER++} includes everything in \textbf{NER+} as well as character-based word representations obtained using a convolutional network over the character sequence in each word.  Unless otherwise indicated, our SPENs use the energy in Eq.~(\ref{eqn:energy-sequence-labeling}).

\section{Results and Analysis}\label{sec:results}

\label{sec: objective_results}

\begin{table}[h!]
\setlength{\tabcolsep}{5pt}
\begin{center}
\begin{tabular}{lcc|c||c||c|}
\cline{4-6}
&zero&& \multicolumn{1}{c||}{POS}
& \multicolumn{1}{c||}{NER}
& \multicolumn{1}{c|}{NER+}
\\ \cline{4-6}
 &trunc.&CE & acc (\%) & F1 (\%) & F1 (\%)\\ 
\hline
\multicolumn{1}{|c}{} & yes & no & 13.9  &  3.91  & 3.91\\
\multicolumn{1}{|c}{margin-} & no & no & 87.9 & 85.1 & 88.6\\
\multicolumn{1}{|c}{rescaled} & yes & yes & 89.4{{\makebox[0pt][l]{*}}} &  85.2{{\makebox[0pt][l]{*}}} &  89.5{{\makebox[0pt][l]{*}}} \\
\multicolumn{1}{|c}{} & no & yes & 89.4 & 85.2 & 89.5\\
\hline

\multicolumn{1}{|c}{\multirow{2}{*}{perceptron}} & no & no & 88.2  & 84.0 & 88.1\\
\multicolumn{1}{|c}{} & no & yes & 88.6 & 84.7 & 89.0 \\
\hline
\end{tabular}
\end{center}
\caption{Test set results for Twitter POS tagging and NER of several SPEN configurations. Results with * correspond to the setting of Section~\ref{sec:SPENTraining}.\label{table:results2}} 
\end{table}

\begin{table*}[th]
\setlength{\tabcolsep}{4pt}
\begin{center}
\begin{tabular}{lc|c|c|c|c||c|c|c|c||c|}
\cline{3-11}
&& \multicolumn{4}{c||}{POS}
& \multicolumn{4}{c||}{NER}
& \multicolumn{1}{c|}{NER+}
\\ 
 && \multicolumn{1}{c}{acc (\%)} & \multicolumn{1}{c}{$|T|$} &    \multicolumn{1}{c}{$|I|$}  &\multicolumn{1}{c||}{speed} & \multicolumn{1}{c}{F1 (\%)} & \multicolumn{1}{c}{$|T|$} &  \multicolumn{1}{c}{$|I|$} & \multicolumn{1}{c||}{speed}  & F1 (\%) \\
\hline
\multicolumn{2}{|l|}{BiLSTM}   & 88.8 & 166K  &166K & -- & 84.9 & 239K & 239K & -- &  89.3 \\
\hline
\\ 
\multicolumn{9}{l}{\textbf{SPENs with inference networks in Section~\ref{sec:SPENTraining}:}} \\
\hline
\multicolumn{2}{|l|}{margin-rescaled} & 89.4 & 333K  & 166K & -- & 85.2 & 479K &239K & -- &89.5 \\

\multicolumn{2}{|l|}{perceptron} & 88.6 & 333K &166K & -- & 84.4 & 479K & 239K & -- & 89.0 \\
\hline
\\
\multicolumn{11}{l}{\textbf{SPENs with inference networks, compound objective, CE, no zero truncation (this paper):}} \\
\hline
\multicolumn{2}{|l|}{separated} & 89.7 & 500K & 166K & 66 &  85.0 & 719K & 239K & 32 & 89.8 \\
\multicolumn{2}{|l|}{shared} & 89.8 & 339K & 166K & 78 & 85.6 & 485K & 239K & 38 & 90.1 \\
\multicolumn{2}{|l|}{\bf{stacked}} & \bf{89.8} & \bf{335K} & \bf{166K} &  \bf{92} & \bf{85.6} & \bf{481K} & \bf{239K} & \bf{46} & \bf{90.1} \\
\hline
\end{tabular}
\end{center}
\caption{Test set results for Twitter POS tagging and NER. 
$|T|$ is the number of trained parameters; $|I|$ is the number of parameters needed during the inference procedure. 
Training speeds (examples/second) are shown for joint parameterizations to compare them in terms of efficiency. Best setting (highest performance with fewest parameters and fastest training) is in boldface. 
\label{table:results1}}
\end{table*}

\paragraph{Effect of Removing Truncation.}
Table~\ref{table:results2} shows results for the margin-rescaled and perceptron losses when considering the removal of zero truncation and its interaction with the use of the local CE term. 
Training fails for both tasks when using zero truncation without the CE term. Removing truncation makes learning succeed and leads to effective models even without using CE. 
However, when using the local CE term, truncation has little effect on performance. The importance of CE in Section~\ref{sec:SPENTraining} is likely due to the fact that truncation was being used. 

\vspace{-0.2cm}
\paragraph{Effect of Local CE.}
The local cross entropy (CE) term is useful for both tasks, though it appears more helpful for tagging. This may be because POS tagging is a more local task. 
Regardless, for both tasks, the inclusion of the CE term speeds convergence and improves training stability. 
For example, on NER, using the CE term reduces the number of epochs chosen by early stopping from $\sim$100 to $\sim$25. On Twitter POS Tagging, using the CE term reduces the number of epochs chosen by early stopping from $\sim$150 to $\sim$60.

\paragraph{Effect of Compound Objective and Joint Parameterizations.}
The compound objective is the sum of the margin-rescaled and perceptron losses, and outperforms them both (see Table \ref{table:results1}). 
Across all tasks, the shared and stacked parameterizations are more accurate than the previous objectives. For the separated parameterization, the performance drops slightly for NER, likely due to the larger number of parameters. 
The shared and stacked options have fewer parameters to train than the separated option, and the stacked version processes examples at the fastest rate during training.

\begin{table}[t]
\centering
\begin{tabular}{lc|c||c|}
\cline{3-4}
& & \multicolumn{1}{c||}{POS}
& \multicolumn{1}{c|}{NER}
\\ \cline{3-4}
   &&  $  \infnet - \canet  $ &  $ \infnet - \canet  $ \\
\hline
\multicolumn{2}{|l|}{margin-rescaled} & 0.2 & 0\\
\hline
\multicolumn{1}{|l}{} & separated & 2.2 & 0.4\\
\multicolumn{1}{|l}{compound} & shared & 1.9  & 0.5\\
\multicolumn{1}{|l}{} & stacked & \bf{2.6}  & \bf{1.7}\\
\hline
\end{tabular}
\\
\medskip
\begin{tabular}{cc}
test-time ($\infnet$)  & cost-augmented ($\canet$) \\
\hline
 common noun  & proper noun \\
 proper noun & common noun   \\
 common noun  & adjective  \\
 proper noun  & proper noun + possessive\\
 adverb & adjective \\
 preposition & adverb \\
 adverb & preposition \\
 verb &  common noun\\
 adjective & verb \\
\end{tabular}
\caption{Top: differences in accuracy/F1 between test-time inference networks $\infnet$ and cost-augmented networks $\canet$ (on development sets). 
The ``margin-rescaled'' row uses a SPEN with the local CE term and without zero truncation, where $\infnet$ is obtained by fine-tuning $\canet$ as done by \citet{tu-18}. Bottom: most frequent output differences between $\infnet$ and $\canet$ on the development set. 
\label{table:phi-vs-psi}
}
\end{table}

The top part of Table~\ref{table:phi-vs-psi} shows how the performance of the test-time inference network $\infnet$ and the cost-augmented inference network $\canet$ vary when using the new compound objective. 
The differences between $\canet$ and $\infnet$ are larger than in the baseline configuration, showing that the two are learning complementary functionality. 
With the stacked parameterization, the cost-augmented network $\canet$ receives as an additional input the gold standard label sequence, which leads to the largest differences as the cost-augmented network can explicitly favor incorrect labels.\footnote{We also tried a BiLSTM in the final layer of the stacked parameterization but results were similar to the simpler affine architecture, so we only report results for the latter.}

The bottom part of Table~\ref{table:phi-vs-psi} shows qualitative differences between the two inference networks. On the POS development set, we count the differences between the predictions of $\infnet$ and $\canet$ when 
$\infnet$ makes the correct prediction.\footnote{We used the stacked parameterization.} 
$\canet$ tends to output tags that are highly confusable with those output by $\infnet$. For example, it often outputs proper noun when the gold standard is common noun or vice versa. It also captures the 
ambiguities among adverbs, adjectives, and prepositions.

\paragraph{Global Energies.}
The results are shown in Table \ref{table:tlm_result}.
Adding the backward (b) and word-augmented TLMs (c) improves over using only the forward TLM from \citet{tu-18}. 
With the global energies, our performance is comparable to several strong results (90.94 of \citealp{lample-etal-2016-neural} and 91.37 of \citealp{ma-hovy-2016-end}). However, it is still lower than the state of the art~\citep{akbik-etal-2018-contextual,devlin-etal-2019-bert}, likely due to the lack of 
contextualized embeddings. 
In the next Section~\ref{sec:ArbitrayOrder}, we proposed and evaluated several other high-order energy terms for sequence labeling using this framework.

\begin{table}[th!]
\centering
\small
\begin{tabular}{|m{3cm}|c|c|c|}
\cline{2-4}
\multicolumn{1}{l|}{}  & NER & NER+ & NER++\\ 
\hline
margin-rescaled & 85.2 & 89.5 & 90.2\\
\hline
{compound, stacked, CE, no truncation} & 85.6 & 90.1 & 90.8\\ 
\hline
+ global energy GE(a) & 85.8 & 90.2 & 90.7\\
+ global energy GE(b) & 85.9 & 90.2 & 90.8\\
+ global energy GE(c) &  \bf{86.3} & \bf{90.4} & \bf{91.0}\\
\hline
\end{tabular}
\caption{NER test F1 scores with global energy terms. 
}
\label{table:tlm_result}
\end{table}

\section{Constituency Parsing Experiments}

We linearize the constituency parsing outputs, similar to \citet{tran-etal-2018-parsing}. We use the following equation plus global energy in the form of Eq.~(8) as the energy function: 
\begin{align}
E_{\Theta}(\x, \y) &= -\Bigg(\sum_{t=1}^T  \sum_{j=1}^L y_{t,j}\! \left(U_j^\top b(\x,t)\right) \nonumber \\
& + \sum_{t=1}^T \y_{t-1}^\top W \y_{t}\Bigg) \nonumber 
\end{align}
Here, $b$ has a seq2seq-with-attention architecture identical to \citet{tran-etal-2018-parsing}. In particular, here is the list of implementation decisions. 
\begin{itemize}
    \item We can write $b=g\circ f$ where $f$ (which we call the ``feature network'') takes in an input sentence, passes it through the encoder, and passes the encoder output to the decoder feature layer to obtain hidden states; $g$ takes in the hidden states and passes them into the rest of the layers in the decoder. In our experiments, the cost-augmented inference network $\canet$, test-time inference network $\infnet$, and $b$ of the energy function above share the same feature network (defined as $f$ above). 
    \item The feature network ($f$) component of $b$ is pretrained using the feed-forward local cross-entropy objective. The cost-augmented inference network $\canet$ and the test-time inference network $\infnet$ are both pretrained using the feed-forward local cross-entropy objective. 
\end{itemize}

The seq2seq baseline achieves 82.80 F1 on the development set in our replication of \citet{tran-etal-2018-parsing}. Using a SPEN with our stacked parameterization, we obtain 83.22 F1.

\section{Conclusions}

We contributed several strategies to stabilize and improve joint training of SPENs and inference networks. Our use of
joint parameterizations mitigates the need for inference network fine-tuning, leads to complementarity in the learned inference networks, 
and yields improved performance overall. These developments offer promise for SPENs to be more easily applied to a broad 
range of NLP tasks. 
Future work will explore other structured prediction tasks, such as parsing and generation. 
We have taken initial steps in this direction, considering constituency parsing with the 
sequence-to-sequence model of \citet{tran-etal-2018-parsing}. 
Preliminary experiments are positive,\footnote{On NXT Switchboard \citep{calhoun2010nxt}, the baseline achieves 82.80 F1 on the development set and the SPEN (stacked parameterization) achieves 83.22. More details are in the appendix.} 
but significant challenges remain, specifically in defining appropriate inference network architectures to enable efficient learning. 
\newpage

\mychapter{7}{Exploration of Arbitrary-Order Sequence Labeling}
\label{sec:ArbitrayOrder}

A major challenge with CRFs is the complexity
of training and inference, which are
quadratic in the number of output labels for first order models and grow exponentially when higher order dependencies are considered. This explains why the most common type of CRF used in practice is a first order model, also referred to as a ``linear chain'' CRF. 

In the previous chapter, we propose a framework that can Jointly train of energy functions and inference networks. In this section, we leverage the frameworks to explore high-order energy functions for sequence labeling. Naively instantiating high-order energy terms can lead to a very large number of parameters to learn, so we instead develop concise neural parameterizations for high-order terms. In particular, we draw from vectorized Kronecker products, convolutional networks, recurrent networks, and self-attention.

This chapter includes some material originally presented in ~\citet{tu-etal-2020-exploration}.

\section{Introduction}

Conditional random fields (CRFs; \citealp{Lafferty:2001:CRF}) have been shown to perform well in various sequence labeling tasks. Recent work uses rich neural network architectures to define the ``unary'' potentials, i.e., terms that only consider a single position's label at a time~\citep{journals/jmlr/CollobertWBKKK11,lample-etal-2016-neural,ma-hovy-2016-end,strubell-etal-2018-linguistically}. However, ``binary'' potentials, which consider pairs of adjacent labels, are usually quite simple and may consist solely of a parameter or parameter vector for each unique label transition. Models with unary and binary potentials are generally referred to as ``first order'' models. 

A major challenge with CRFs is the complexity
of training and inference, which are
quadratic in the number of output labels for first order models and grow exponentially when higher order dependencies are considered. This explains why the most common type of CRF used in practice is a first order model, also referred to as a ``linear chain'' CRF. 

One promising alternative to CRFs is structured prediction energy networks (SPENs; \citealp{belanger2016structured}), which use deep neural networks to parameterize arbitrary potential functions for structured prediction. While SPENs also pose challenges for learning and inference, in the previous chapters, we proposed a way to train SPENs jointly with ``inference networks'', neural networks trained to approximate structured $\argmax$ inference. 

In this paper, we leverage the frameworks of SPENs and inference networks to explore high-order energy functions for sequence labeling. Naively instantiating high-order energy terms can lead to a very large number of parameters to learn, so we instead develop concise neural parameterizations for high-order terms. In particular, we draw from vectorized Kronecker products, convolutional networks, recurrent networks, and self-attention. 
We also consider ``skip-chain'' connections~\citep{sutton04skip} with various skip distances and ways of reducing their total parameter count for increased learnability. 

Our experimental results on four sequence labeling tasks show that a range of high-order energy functions can yield performance improvements. While the optimal energy function varies by task, we find strong performance from skip-chain terms with short skip distances, convolutional networks with filters that consider label trigrams, and recurrent networks and self-attention networks that consider large subsequences of labels.

We also demonstrate that modeling high-order dependencies can lead to significant performance improvements in the setting of noisy training and test sets. 
Visualizations of the high-order energies show various methods capture intuitive structured dependencies among output labels. 

Throughout, we use inference networks that share the same architecture as unstructured classifiers for sequence labeling, so test time inference speeds are unchanged between local models and our method. 
Enlarging the inference network architecture by adding one layer leads consistently to better results, rivaling or improving over a BiLSTM-CRF baseline, 
suggesting that training efficient inference networks with high-order energy terms can make up for errors arising from approximate inference.
While we focus on sequence labeling in this paper, our results show the potential of developing high-order structured models for other NLP tasks in the future.

\section{Energy Functions}

\label{sec:energy}
Considering sequence labeling tasks, the input $\x$ is a length-$T$ sequence of tokens where $x_t$ denotes the token at position $t$. The output $\y$ is a sequence of labels also of length $T$. 
We use $\y_t$ to denote the output label at position $t$, where $\y_t$ is a vector of length $L$ (the number of labels in the label set) and where $y_{t,j}$ is the $j$th entry of the vector $\y_t$. In the original output space $\yspace(\x)$, $y_{t,j}$ is 1 for a single $j$ and 0 for all others. In the relaxed output space $\relyspace(\x)$, $y_{t,j}$  can be interpreted as the probability of the $t$th position being labeled with label $j$. 
We use the following energy:
\begin{align}
E_{\Theta}(\x, \y) = 
-\Bigg(\sum_{t=1}^T  \sum_{j=1}^L y_{t,j}\! \left(U_j^\top b(\x,t)\right) 
+  E_{W}( \y) \Bigg) 
\label{eqn:energy}
\end{align}
where $U_{j}\in\mathbb{R}^d$ is a parameter vector for label $j$  
and $E_{W}( \y)$ is a structured energy term parameterized by parameters $W$. In a linear chain CRF, $W$ is a transition matrix for scoring two adjacent labels. Different instantiations of $E_{W}$ will be detailed in the sections below. 
Also, $b(\x,t)\in\mathbb{R}^d$ denotes the ``input feature vector'' for position $t$. 
We define it to be the $d$-dimensional BiLSTM~\citep{Hochreiter:1997:LSM:1246443.1246450} hidden vector at $t$.
The full set of energy parameters $\Theta$ includes the $U_j$ vectors, $W$, and the parameters of the BiLSTM. 

 Table~\ref{tab:complexity} shows the training and test-time inference requirements 
of our method compared to previous methods. For different formulations of the energy function, the inference network architecture is the same (e.g., BiLSTM). So the inference complexity is the same as the standard neural approaches that do not use structured prediction, which is linear in the label set size. However, even for the first order model (linear-chain CRF), the time complexity is quadratic in the label set size. The time complexity of higher-order CRFs grows exponentially with the order. 

\begin{table*}[!h]
    \setlength{\tabcolsep}{6pt}
    \centering
    \begin{tabular}{lc|c|c||c|c|}
\cline{3-6}
&& \multicolumn{2}{c||}{Training}
& \multicolumn{2}{c|}{Inference}
\\
&& \multicolumn{1}{c}{Time} & \multicolumn{1}{c||}{Number of Parameters}  & \multicolumn{1}{c}{Time} & \multicolumn{1}{c|}{Number of Parameters} \\ 
\hline
\multicolumn{2}{|c|}{BiLSTM} &  $\mathcal{O}(T*L)$ & $\mathcal{O}(|\Psi|)$  & $\mathcal{O}(T*L)$ & $\mathcal{O}(|\Psi|)$ \\
\multicolumn{2}{|c|}{CRF} & $\mathcal{O}(T*L^2)$   & $\mathcal{O}(|\Theta|)$  & $\mathcal{O}(T*L^2)$ & $\mathcal{O}(|\Theta|)$ \\
\hline \hline
\multicolumn{2}{|l|}{Energy-Based Inference Networks} & $\mathcal{O}(T*L)$  & $\mathcal{O}(|\Psi| + |\Phi| + |\Theta|)$  & $\mathcal{O}(T*L)$ & $\mathcal{O}(|\Psi|)$\\
\hline
    \end{tabular}
    \caption{Time complexity and number of parameters of different methods during training and inference, where $T$ is the sequence length, $L$ is the label set size, $\Theta$ are the parameters of energy function, and $\Phi, \Psi$ are the parameters of two energy-based inference networks. For arbitrary-order energy functions or different parameterizations, the size of $\Theta$ can be different.
    }
    \label{tab:complexity}
\end{table*}

\begin{figure}[h]
\centering
\includegraphics[width=1.0\textwidth]{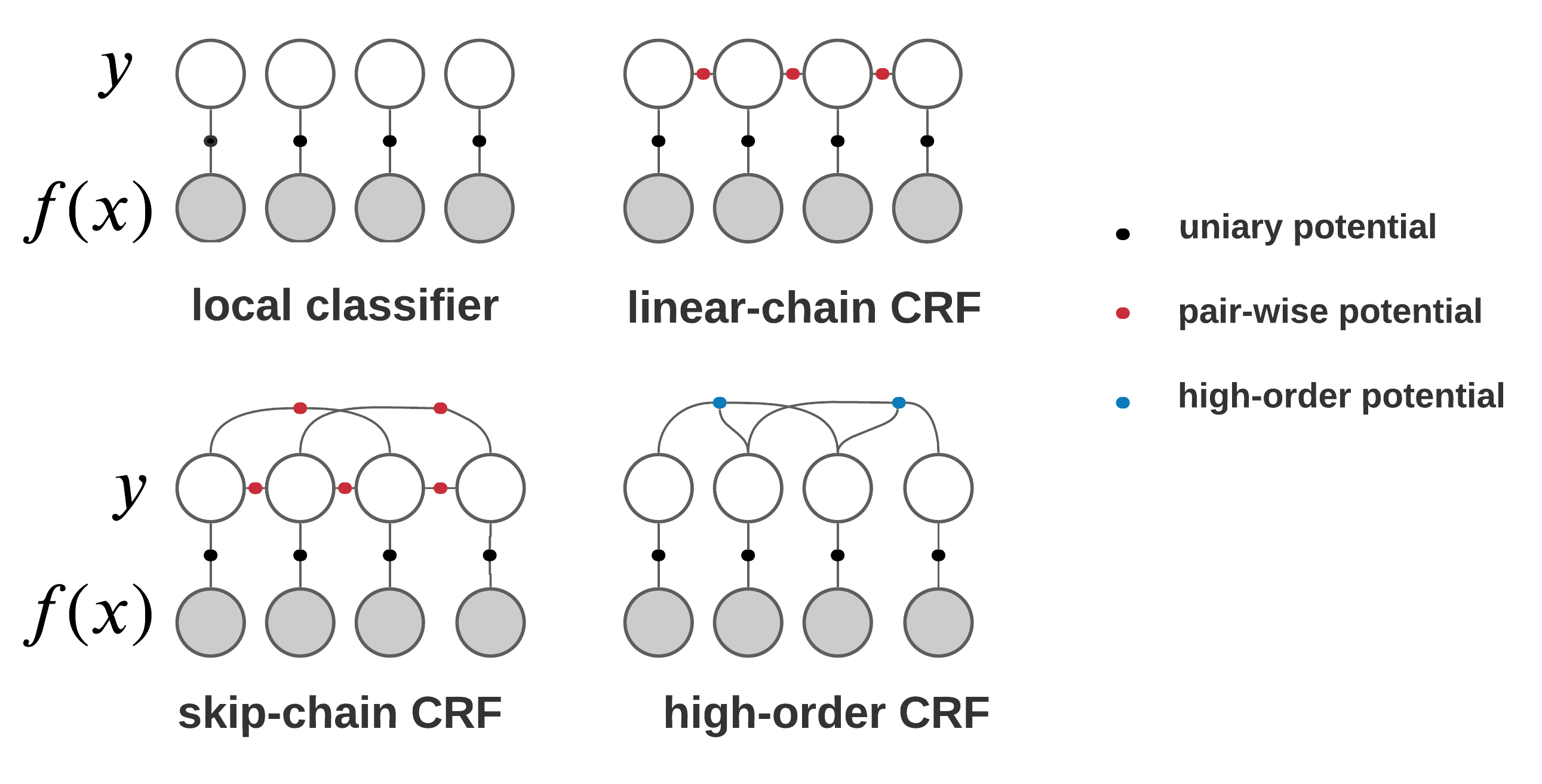}
\caption{Visualization of the models with different orders. }
\end{figure}

\subsection{Linear Chain Energies} 

Our first choice for a structured energy term is relaxed linear chain energy defined for sequence labeling by \citet{tu-18}: 
\begin{align}
E_{W}( \y) = \sum_{t=1}^T \y_{t-1}^\top W \y_{t}  \nonumber
\end{align}
Where $W_i\in\mathbb{R}^{L \times L}$ is the transition matrix, which is used to score the pair of adjacent labels. If this linear chain energy is the only structured energy term in use, exact inference can be performed efficiently using the Viterbi algorithm. 

\subsection{Skip-Chain Energies}
\label{sec:skipchain}
We also consider an energy inspired by ``skip-chain'' conditional random fields~\citep{sutton04skip}. In addition to consecutive labels, this energy also considers pairs of labels appearing in a given window size $M+1$:
\begin{align}
E_{W}( \y) = \sum_{t=1}^T \sum_{i=1}^M \y_{t-i}^\top W_i \y_{t}  \nonumber
\end{align}
where each $W_i\in\mathbb{R}^{L \times L}$ and the max window size $M$ is a hyperparameter. While linear chain energies allow efficient exact inference, using skip-chain energies causes exact inference to require time exponential in the size of $M$.

\subsection{High-Order Energies}
\label{sec:hoEnergy}
We also consider $M$th-order energy terms. We use the function $F$ to score the $M+1$ consecutive labels $\y_{t-M}, \dots, \y_{t}$, then sum over positions:  
\begin{align}
E_{W}( \y) = \sum_{t=M}^T F(\y_{t-M},\dots,\y_t) 
\label{eqn:high-order}
\end{align}
\noindent We consider several different ways to define the function $F$, detailed below.
\paragraph{Vectorized Kronecker Product (VKP):} 
A naive way to parameterize a high-order energy term would involve using a parameter tensor $W \in\mathbb{R}^{L^{M+1}}$ with an entry for each possible label sequence of length $M+1$. To avoid this exponentially-large number of parameters, we define a more efficient parameterization as follows. 
We first define a label embedding lookup table $\in \mathbb{R}^{L \times n_l}$ and denote the embedding for label $j$ by $e_j$. We consider $M=2$ as an example. Then, for a tensor $W \in\mathbb{R}^{L\times L\times L}$, its value $W_{i,j,k}$ at indices $(i,j,k)$ is calculated as 
$$
\pvect^\top \mathrm{LayerNorm}([e_i;e_j;e_k]+\MLP([e_i;e_j;e_k]))$$
\noindent where $\pvect \in \mathbb{R}^{(M+1) n_l}$ is a parameter vector and $`;'$ denotes vector concatenation. $\MLP$ expects and returns vectors of dimension ${(M+1) \times n_l}$ and is parameterized as a multilayer perceptron. Then, the energy is computed:
\begin{align}
F(\y_{t-M},\dots,\y_t) = \product(\y_{t-M}, \dots, \y_{t-1})W \y_t
\nonumber 
\end{align}
\noindent where $W$ is reshaped as $\in\mathbb{R}^{L^M \times L}$. 
The operator $\product$ is somewhat similar to the Kronecker product of the $k$ vectors $\vv_{1}, \dots, \vv_{k}$\footnote{There are some work~\citep{lei-etal-2014-low,NIPS2014_5323, yu-etal-2016-embedding} that use Kronecker product for higher order feature combinations with low-rank tensors. Here we use this form to express the computation when scoring the consecutive labels. }. However it will return a vector, not a tensor: 
\begin{align}
    & \product(\vv_{1}, \dots, \vv_{k}) = \nonumber \\
    &\begin{cases}
     \vv_{1} & k=1 \\
     \mathbf{vec}(\vv_{1} \vv_{2}^\top) & k=2 \\
     \mathbf{vec}(\product(\vv_1,\dots, \vv_{k-1}) \vv_{k}^\top) & k > 2
    \end{cases} \nonumber
\end{align}
\noindent Where $\mathbf{vec}$ is the operation that vectorizes a tensor into a (column) vector.

\paragraph{CNN:} Convolutional neural networks (CNN)  are frequently used in NLP to extract features based on words or characters~\citep{journals/jmlr/CollobertWBKKK11,kim-2014-convolutional}. We apply CNN filters over the sequence of $M+1$ consecutive labels. 
The $F$ function is computed as follows:
\begin{align}
&F(\y_{t-M},\dots,\y_t) = \sum_n f_{n}(\y_{t-M},\dots,\y_t) \nonumber \\
&f_{n}(\y_{t-M},\dots,\y_t) = g(W_n [\y_{t-M};
...;\y_{t} ] +\mathbf{b}_n)  \nonumber
\end{align}
\noindent where $g$ is a ReLU nonlinearity and the vector $W_n \in\mathbb{R}^{L(M+1)}$ and scalar $b_n \in\mathbb{R}$ are the parameters for filter $n$. 
The filter size of all filters is the same as the window size, namely, $M+1$. The $F$ function sums over all CNN filters. When viewing this high-order energy as a CNN, we can think of the summation in Eq.~\ref{eqn:high-order} as corresponding to sum pooling over time of the feature map outputs.

\paragraph{Tag Language Model (TLM):} \citet{tu-18} defined an energy term based on a pretrained ``tag language model'', which computes the probability of an entire sequence of labels. We also use a TLM, scoring a sequence of $M+1$ consecutive labels in a way similar to \citet{tu-18}; however, the parameters of the TLM are trained in our setting:
\begin{align}
& F(\y_{t-M},\dots,\y_t) =  \nonumber \\
& - \! \sum_{t'=t-M+1}^{t} \! \y_{t'}^\top \log(\LM(\langle \y_{t-M},...,\y_{t'-1}\rangle) ) \nonumber
\end{align}
\noindent where $\LM(\langle \y_{t-M},...,y_{t'-1}\rangle)$ returns the softmax distribution over tags at position $t'$ (under the  tag language model) given the preceding tag vectors. When each $y_{t'}$ is a one-hot vector, this energy reduces to the negative log-likelihood of the tag sequence specified by $\y_{t-M},\dots, \y_t$.

\paragraph{Self-Attention (S-Att):} We adopt the multi-head self-attention formulation from \citet{NIPS2017_7181}. Given a matrix of the $M+1$ consecutive labels $Q=K=V= [\y_{t-M};\dots;\y_{t} ] \in \mathbb{R}^{(M+1) \times L}$: 
\begin{align}
    H = \attention(Q, K, V) \nonumber \\
    F(\y_{t-M},\dots,\y_t) = \sum H \nonumber
\end{align}
\noindent where $\attention$ is the general attention mechanism: the weighted sum of the value vectors $V$ using query vectors $Q$ and key vectors $K$ \citep{NIPS2017_7181}. The energy on the $M+1$ consecutive labels is defined as the sum of entries in the feature map $H \in \mathbb{R}^{L\times (M+1)}$ after the self-attention transformation.

\subsection{Fully-Connected Energies}
We can simulate a ``fully-connected'' energy function by setting a very large value for $M$ in the skip-chain energy (Section \ref{sec:skipchain}). For efficiency and learnability, we
use a low-rank parameterization for the many translation matrices $W_i$ that will result from increasing $M$. We first define a matrix $S\in \mathbb{R}^{L \times d}$ that all $W_i$ will use. Each $i$ has a learned parameter matrix $D_i \in \mathbb{R}^{L \times d} $ and together $S$ and $D_i$ are used to compute $W_i$: 
\begin{align}
    W_i = S D_i^\top \nonumber
\end{align}
\noindent where $d$ is a tunable hyperparameter that affects the number of learnable parameters.

\section{Related Work}

Linear chain CRFs \citep{Lafferty:2001:CRF},  which consider dependencies between at most two adjacent labels or segments, are commonly used in practice \citep{NIPS2004_2648,lample-etal-2016-neural,ma-hovy-2016-end}. 

There have been several efforts in developing efficient algorithms for handling higher-order CRFs. 
\citet{QianJZHW09} developed an efficient decoding algorithm under the assumption that all  high-order features have non-negative weights. Some work has shown that high-order CRFs can be handled relatively efficiently if particular patterns of sparsity are assumed~\citep{NIPS2009_3815,JMLR:v15:cuong14a}. \citet{mueller-etal-2013-efficient} proposed an approximate CRF using coarse-to-fine decoding and early updating. 
Loopy belief propagation~\citep{MurphyWJ99} has been used for  approximate inference in high-order CRFs, such as skip-chain CRFs~\citep{sutton04skip}, which form the inspiration for one category of energy function in this paper. .

CRFs are typically trained by maximizing conditional log-likelihood. Even assuming that the graph structure underlying the CRF admits tractable inference, it is still time-consuming to compute the partition function. Margin-based methods have been proposed~\citep{m3,ssvm} to avoid the summation over all possible outputs. Similar losses are used when training SPENs~\citep{belanger2016structured,End-to-EndSPEN}, including in this paper. .  
The energy-based inference network learning framework has been used for multi-label classification \citep{tu-18}, non-autoregressive machine translation \citep{tu-etal-2020-engine}, and previously for sequence labeling \citep{tu-gimpel-2019-benchmarking}.

Moving beyond CRFs and sequence labeling, there has been a great deal of work in the NLP community in designing non-local features,
often combined with the development of approximate algorithms to incorporate them during inference. 
These include $n$-best reranking \citep{och-etal-2004-smorgasbord}, 
beam search \citep{Lowerre:1976:HSR:907741}, 
loopy belief propagation \citep{sutton04skip,smith-eisner-2008-dependency}, 
Gibbs sampling \citep{finkel-etal-2005-incorporating}, 
stacked learning \citep{DBLP:conf/ijcai/CohenC05,krishnan-manning-2006-effective}, 
sequential Monte Carlo algorithms~\citep{yang-eisenstein-2013-log}, 
dynamic programming approximations like cube pruning~\citep{chiang-2007-hierarchical,huang-chiang-2007-forest},
dual decomposition \citep{rush-etal-2010-dual,martins-etal-2011-dual}, 
and methods based on black-box optimization like integer linear programming~\citep{roth-yih-2004-linear}. These methods are often developed or applied with particular types of non-local energy terms in mind. By contrast, here we find that the framework of SPEN learning with inference networks can support a wide range of high-order energies for sequence labeling.

\section{Experimental Setup}
We perform experiments on four tasks: Twitter part-of-speech tagging (POS), named entity recognition (NER), CCG supertagging (CCG), and semantic role labeling (SRL).

\subsection{Datasets}
\paragraph{POS.}
We use the annotated data from \citet{gimpel-etal-2011-part} and \citet{owoputi-etal-2013-improved} which contains
25 POS tags.
We use the 100-dimensional skip-gram embeddings from \citet{tu-17-long} which were trained on a dataset of 56 million English tweets using  \texttt{word2vec}~\citep{mikolov2013distributed}. The evaluation metric is tagging accuracy.

\paragraph{NER.}
We use the CoNLL 2003 English data \citep{tjong-kim-sang-de-meulder-2003-introduction}. 
We use the BIOES tagging scheme, so there are 17 labels. 
We use 100-dimensional pretrained GloVe~\citep{pennington-etal-2014-glove} embeddings.  
The task is evaluated with micro-averaged F1 score.  

\paragraph{CCG.}
We use the standard splits from CCGbank~\citep{hockenmaier-steedman-2002-acquiring}.
We only keep sentences with length less than 50 in the original training data during training. 
We use only the 400 most frequent labels.
The training data contains 1,284 unique labels, but  because the label distribution has a long tail,  we use only the 400 most frequent labels, replacing the others by a special tag $*$. The percentages of $*$ in train/development/test are 0.25/0.23/0.23$\%$.  When the gold standard tag is $*$, the prediction is always evaluated as incorrect. We use the same GloVe embeddings as in NER. . 
The task is evaluated with per-token accuracy.

\paragraph{SRL.}
We use the standard split from CoNLL 2005 \citep{carreras-marquez-2005-introduction}. The gold predicates are provided as part of the input. We use the official evaluation script from the CoNLL 2005 shared task for  evaluation. 
We again use the same GloVe embeddings as in NER. 
To form the inputs to our models, an embedding of a binary feature indicating whether the word is the given predicate is concatenated to the word embedding.\footnote{Our SRL baseline  is most similar to \citet{zhou-xu-2015-end}, though there are some differences. 
We use GloVe embeddings while they train word embeddings on Wikipedia. We both use the same predicate context features.} 

\section{Training}
\paragraph{Local Classifiers.}
We consider local baselines that use a BiLSTM trained with the local loss $\ltok$. For POS, NER and CCG, we use a 1-layer BiLSTM with hidden size 100, and the word embeddings are fixed during training. For SRL, we use a 4-layer BiLSTM with hidden size 300 and the word embeddings are fine-tuned. 

\paragraph{BiLSTM-CRF.}
We also train BiLSTM-CRF models with the standard conditional log-likelihood objective. A 1-layer BiLSTM with hidden size 100 is used for extracting input features. The CRF part uses a linear chain energy with a single tag transition parameter matrix. 
We do early
stopping based on development sets. The usual dynamic programming algorithms are used for training and inference, e.g., the Viterbi algorithm is used for inference. The same pretrained word embeddings as for the local classifiers are used.

\paragraph{Inference Networks.}
When defining architectures for the inference networks, we use the same architectures as the local classifiers. However, the objective of the inference networks is different. $\lambda=1$ and $\tau=1$ are used for training. We do early
stopping based on the development set.

\paragraph{Energy Terms.}
The unary terms are parameterized using a one-layer BiLSTM with hidden size 100. 
For the structured energy terms, the $\product$ operation uses $n_l=20$, the number of CNN filters is 50, and the tag language model is a 1-layer LSTM with hidden size 100. For the fully-connected energy, $d=20$ for the approximation of the transition matrix and $M=20$ for the approximation of the fully-connected energies.

\paragraph{Hyperparameters.}
For the inference network training, the batch size is 100. We update the energy function parameters using the Adam optimizer~\citep{adam} with learning rate 0.001. For POS, NER, and CCG, we train the inference networks parameter with stochastic gradient descent with momentum as the optimizer. The learning rate is 0.005 and the momentum is 0.9. For SRL, we train the inference networks using Adam with learning rate 0.001.

\section{Results}

\paragraph{Parameterizations for High-Order Energies.}

\begin{table}[t]
    \centering
    \begin{tabular}{ll|c|c|c|}
\cline{3-5}
&  & POS & NER & CCG\\ \hline
\multicolumn{2}{|l|}{Linear Chain}  & 89.5 & 90.6 & 92.8 \\
\hline \hline
\multicolumn{1}{|l|}{\multirow{3}{*}{VKP}} & $M=2$ & \textbf{89.9} & 91.1 & \textbf{93.1}\\
\multicolumn{1}{|l|}{} & $M=3$ & 89.8  & \textbf{91.2} & 92.9\\
\multicolumn{1}{|l|}{} & $M=4$ & 89.5  & 90.8 & 92.8\\
\hline \hline
\multicolumn{1}{|l|}{} & $M=1$ & 89.7 & 91.1 & 93.0 \\
\multicolumn{1}{|l|}{\multirow{2}{*}{CNN}} & $M=2$ & \textbf{90.0}  & \textbf{91.3} & \textbf{93.0} \\
\multicolumn{1}{|l|}{} & $M=3$ & 89.9  & 91.2 & 92.9\\
\multicolumn{1}{|l|}{} & $M=4$ & 89.7   & 91.0 & \textbf{93.0} \\
\hline \hline
\multicolumn{1}{|l|}{} & $M=2$ & 89.7 &  90.8 & 92.4\\
\multicolumn{1}{|l|}{\multirow{2}{*}{TLM}} & $M=3$ & 89.8  & 91.0 & 92.7\\
\multicolumn{1}{|l|}{} & $M=4$ & 89.8  & 91.3 & 92.7\\
\multicolumn{1}{|l|}{} & all &  \textbf{90.0}  & \textbf{91.4} & \textbf{92.9}\\
\hline \hline
\multicolumn{1}{|l|}{} & $M=2$ & 89.7 & 90.7 & 92.6  \\
\multicolumn{1}{|l|}{} & $M=4$ & 89.8 & 90.8 & 92.8\\
\multicolumn{1}{|l|}{\multirow{1}{*}{S-Att}} & $M=6$ & \textbf{89.9}  & 90.9 & 92.8 \\
\multicolumn{1}{|l|}{} & $M=8$ &  \textbf{89.9} & \textbf{91.0} & 93.0\\
\multicolumn{1}{|l|}{} & all & 89.7 & 90.8 & \textbf{93.1} \\
\hline
    \end{tabular}
    \caption{Development results for different parameterizations of high-order energies when increasing the window size $M$ of  consecutive labels, where ``all'' denotes the whole relaxed label sequence. The inference network architecture is a one-layer BiLSTM. We ran $t$-tests for the mean performance (over five runs) of our proposed energies (the settings in bold) and the linear-chain energy. 
    All differences are significant at $p<0.001$ for NER and $p<0.005$ for other tasks.
    }
    \label{tab:HighOrder}
\end{table}

We first compare several choices for energy functions within our inference network learning framework. In Section~\ref{sec:hoEnergy}, we considered several ways to define the high-order energy function $F$. We compare performance of the parameterizations on three tasks: POS, NER, and CCG.
The results are shown in Table~\ref{tab:HighOrder}. 

For $\product$ high-order energies, there are small differences between $2$nd and $3$rd order models, however, $4$th order models are consistently worse.  
The CNN high-order energy is best when $M$=2 for the three tasks. Increasing $M$ does not consistently help. 
The tag language model (TLM) works best when scoring the entire label sequence. In the following experiment with TLM energies, we always use it with this ``all'' setting. 
Self-attention (S-Att) also shows better performance with larger $M$. However, the results for NER are not as high overall as for other energy terms. 

Overall, there is no clear winner among the four types of parameterizations, indicating that a variety of high-order energy terms can work well on these tasks, once appropriate window sizes are chosen. We do note differences among tasks: NER benefits more from larger window sizes than POS.

\paragraph{Comparing Structured Energy Terms.}

Above we compared parameterizations of the high-order energy terms. In Table~\ref{ta:main_result}, we compare instantiations of the structured energy term $E_{W}(\y)$: linear-chain energies, skip-chain energies, high-order energies, and fully-connected energies.\footnote{$M$ values are tuned based on dev sets. Tuned $M$ values for POS/NER/CCG/SRL: Skip-Chain: 3/4/3/3; VKP: 2/3/2/2; CNN: 2/2/2/2; TLM: whole sequence; S-Att: 8/8/8/8.}  
We also compare to local classifiers (BiLSTM). The models with structured energies typically improve over the local classifiers, even with just the linear chain energy. 

The richer energy terms tend to perform better than linear chain, at least for most tasks and energies. 
The skip-chain energies benefit from relatively large $M$ values, i.e., 3 or 4 depending on the task. These tend to be larger than the optimal VKP $M$ values. 
We note that S-Att high-order energies work well on SRL. This points to the  benefits of self-attention on SRL, which has been found in recent work \citep{tan2018,strubell-etal-2018-linguistically}.

Both the skip-chain and high-order energy models achieve substantial improvements over the linear chain CRF, notably a gain of 0.8 F1 for NER. The fully-connected energy is not as strong as the others, possibly due to the energies from label pairs spanning a long range. These long-range energies do not appear helpful for these tasks. 

\begin{table}[t]
\setlength{\tabcolsep}{5pt}
\centering
\begin{tabular}{|ll|c|c|c|cc|}
\cline{3-7}
\multicolumn{2}{l|}{}  & \multicolumn{1}{c|}{\multirow{2}{*}{POS}} & \multicolumn{1}{c|}{\multirow{2}{*}{NER}} & \multicolumn{1}{c|}{\multirow{2}{*}{CCG}} & \multicolumn{2}{c|}{SRL}\\ 
\multicolumn{2}{l|}{}   &  &  &  & WSJ & Brown \\
\hline
\multicolumn{2}{|l|}{BiLSTM} &  88.7 &  85.3  & 92.8 &  81.8 & 71.8 \\
\hline \hline
\multicolumn{2}{|l|}{Linear Chain}  & 89.7 & 85.9 & 93.0 & 81.7 & 72.0\\
\hline \hline
\multicolumn{2}{|l|}{Skip-Chain} & 90.0 & \textbf{86.7}  & \textbf{93.3} & 82.1 & \textbf{72.4}\\
\hline \hline
 & VKP & \textbf{90.1}  & \textbf{86.7} & \textbf{93.3} &  81.8 & 72.0\\
 High-& CNN & \textbf{90.1}  & 86.5 & 93.2 & 81.9  & 72.2 \\
 Order & TLM &  90.0 & 86.6 & 93.0 & 81.8  & 72.1\\
 & S-Att & \textbf{90.1}  & 86.5 & \textbf{93.3} & \textbf{82.2}  & 72.2 \\
\hline \hline
\multicolumn{2}{|l|}{Fully-Connected} & 89.8 & 86.3 & 92.9 & 81.4 & 71.4\\
\hline
\end{tabular}
\caption{Test results on all tasks for local classifiers (BiLSTM) and different structured energy functions. 
POS/CCG use accuracy while NER/SRL use F1. 
The architecture of inference networks is one-layer BiLSTM.  
More results are shown in the appendix. 
}
\label{ta:main_result}
\end{table}

\paragraph{Comparison using Deeper Inference Networks.}

\begin{table}[t]
\centering
\begin{tabular}{|ll|c|c|c|}
\cline{3-5}
\multicolumn{2}{l|}{} & \multicolumn{1}{c|}{POS} & \multicolumn{1}{c|}{NER}  & \multicolumn{1}{c|}{CCG} \\ \hline
\multicolumn{2}{|l|}{2-layer BiLSTM} & 88.8  & 86.0 & 93.4 \\
\hline \hline
\multicolumn{2}{|l|}{BiLSTM-CRF}  & 89.2  & 87.3 & 93.1  \\
\hline \hline
\multicolumn{2}{|l|}{Linear Chain}  & 90.0 & 86.6 &  93.7 \\
\hline \hline
\multicolumn{2}{|l|}{Skip-Chain} & \textbf{90.2} &  \textbf{87.5} & \textbf{93.8} \\
\hline \hline
& VKP  & \textbf{90.2}  & 87.2 &  \textbf{93.8} \\
High-& CNN  & \textbf{90.2}  & 87.3 &  93.6 \\
Order & TLM  & 90.1 & 87.1 &  93.6 \\
& S-Att & 90.0 & 87.3 &  93.7 \\
\hline \hline
\multicolumn{2}{|l|}{Fully-Connected} & 90.0 & 87.2 &  93.3 \\
\hline
    \end{tabular}
    \caption{Test results when inference networks have 2 layers (so the local classifier baseline also has 2 layers).}
    \label{tab:2BiLSTM}
\end{table}

Table~\ref{tab:2BiLSTM} compares methods when using 2-layer BiLSTMs as inference networks.\footnote{$M$ values are retuned based on dev sets when using 2-layer inference networks. Tuned $M$ values for POS/NER/CCG: Skip-Chain: 3/4/3; VKP: 2/3/2; CNN: 2/2/2; TLM: whole sequence; S-Att: 8/8/8.} 
The deeper inference networks reach higher performance across all tasks compared to 1-layer  inference networks. 

We observe that inference networks trained with skip-chain energies and high-order energies achieve better results than BiLSTM-CRF on the three datasets (the Viterbi algorithm is used for exact inference for BiLSTM-CRF). This indicates that adding richer energy terms can make up for approximate inference during training and inference. Moreover, a 2-layer BiLSTM is much cheaper computationally than Viterbi, especially for tasks with large label sets. 

\section{Results on Noisy Datasets}

\begin{table}[t]
    \centering
    \begin{tabular}{|l|c|c|c|}
\cline{2-4}
\multicolumn{1}{l|}{} & \multicolumn{1}{c|}{$\alpha$=0.1} & \multicolumn{1}{c|}{$\alpha$=0.2} & \multicolumn{1}{c|}{$\alpha$=0.3}  \\ \hline
BiLSTM &  75.0  & 67.2 & 58.8 \\
\hline \hline
Linear Chain & 75.2  & 67.4 &  59.1\\
\hline \hline
Skip-Chain ($M$=4) &  75.5  & 67.9 &  59.5\\
\hline \hline
VKP ($M$=3) &  75.3  & 67.7 &  59.3\\
CNN ($M$=0)  & 75.7 & 67.9 & 59.4 \\  
CNN ($M$=2)  & 76.3  & 68.6  & 60.2\\
CNN ($M$=4)  & \textbf{76.7}  & \textbf{69.8} &  \textbf{60.4}\\
TLM  & 76.0  & 67.8 &  59.9\\
S-Att ($M$=8)  & 75.6  & 67.6 &  59.7\\\hline
    \end{tabular}
    \caption{UnkTest setting for NER: 
    words in the test set are replaced by the unknown word symbol with probability $\alpha$. For CNN energies (the settings in bold) and linear-chain energy, they differ significantly with $p<0.001$.} 
    \label{tab:unktest}
\end{table}

\begin{table}[t]
    \centering
    \begin{tabular}{|l|c|c|c|}
\cline{2-4}
\cline{2-4}
\multicolumn{1}{l|}{} & \multicolumn{1}{c|}{$\alpha$=0.1} & \multicolumn{1}{c|}{$\alpha$=0.2}  & \multicolumn{1}{c|}{$\alpha$=0.3} \\ \hline
BiLSTM &  80.1 &  76.0 & 70.6\\
\hline \hline
Linear Chain &  80.4 & 76.3  & 70.9\\
\hline \hline
Skip-Chain ($M$=4) &  81.2 &  76.7 &  71.2\\
\hline \hline
VKP ($M$=3) &  81.4  & 76.8 &  71.4\\
CNN ($M$=0) & 81.1 & 76.7 &  71.5\\
CNN ($M$=2) & 81.8 & 77.0 &  \textbf{71.8}\\
CNN ($M$=4)  & \textbf{82.0} & \textbf{77.1} &  71.7\\
TLM &  80.9 &  76.3 &  71.1\\
S-Att ($M$=8) & 81.4  & 76.9 &  71.4\\\hline
    \end{tabular}
    \caption{UnkTrain setting for NER: training on noisy text, evaluating on noisy test sets. Words are replaced by the unknown word symbol with probability $\alpha$. For CNN energies (the settings in bold) and linear-chain energy, they differ significantly with $p<0.001$. }
    \label{tab:unktrain}
\end{table}

We now consider the impact of our structured energy terms in noisy data settings. Our motivation for these experiments stems from the assumption that structured energies will be more helpful when there is a weaker relationship between the observations and the labels. One way to achieve this is by introducing noise into the observations. 

So, we create new datasets: for any given sentence, we randomly replace a token $x$ with an unknown word symbol ``UNK'' with probability $\alpha$. 
From previous results, we see that NER shows more benefit from structured energies, so we focus on NER 
and consider two 
settings: 
    \textbf{UnkTest}: train on clean text, evaluate on noisy text; and 
  \textbf{UnkTrain}: train on noisy text, evaluate on noisy text.
  
Table~\ref{tab:unktest} shows results for UnkTest. 
CNN 
energies are best among all structured energy terms, including the different parameterizations. 
Increasing $M$ improves F1, showing that high-order information helps the model recover from the high degree of noise. 
Table~\ref{tab:unktrain} shows results for UnkTrain. 
The CNN high-order energies again yield large gains: roughly 2 points compared to the local classifier and 1.8 compared to the linear chain energy.

\section{Incorporating BERT} 
Researchers have recently been applying large-scale pretrained transformers like BERT~\citep{devlin-etal-2019-bert} to many tasks, including sequence labeling. 
To explore the impact of high-order energies on BERT-like models, we now consider experiments that use BERT$_{\text{BASE}}$ in various ways. We use two baselines: (1) BERT finetuned for NER using a local loss, and (2) a CRF using BERT features (``BERT-CRF''). 
Within our framework, we also experiment with using BERT in both the energy function and inference network architecture. 
That is, the ``input feature vector'' in Equation~\ref{eqn:energy} is replaced by the features from BERT. The energy and inference networks are trained with the objective in Section~\ref{sec:jointobjective}. For the training of energy function and inference networks, we use Adam with learning rate $5\mathrm{e}\!-\!5$, a batch size of 32, and L2 weight decay of $1\mathrm{e}\!-\!5$. The results are shown in Table~\ref{tab:berttest}.\footnote{Various high-order energies were explored. We found the skip-chain energy ($M$=3) to achieve the best performance (96.28) on the dev set, so we use it when reporting the test results.}

There is a slight improvement when moving from BERT trained with the local loss to using BERT within the CRF (92.13 to 92.34). There is little difference (92.13 vs.~92.14) between the locally-trained BERT model and when using the linear-chain energy function within our framework. However, when using the higher-order energies, the difference is larger (92.13 to 92.46).

\begin{table}[!h]
    \centering
    \begin{tabular}{lc|c|}
\multicolumn{3}{l}{\textbf{Baselines:}}\\ 
\hline
\multicolumn{2}{|l|}{BERT (local loss)} &  92.13  \\

\multicolumn{2}{|l|}{BERT-CRF} &   92.34 \\
\hline 
\multicolumn{3}{l}{\textbf{Energy-based inference networks:}} \\
\hline
\multicolumn{2}{|l|}{Linear Chain }   & 92.14  \\
\multicolumn{2}{|l|}{Skip-Chain ($M$=3)} & 92.46  \\
\hline
    \end{tabular}
    \caption{Test results for NER when using BERT. When using energy-based inference networks (our framework), BERT is used in both the energy function and as the inference network architecture.} 
    \label{tab:berttest}
\end{table}

\section{Analysis of Learned Energies} 

In this section, we visualize our learned energy  functions for NER to see what structural dependencies among labels have been captured. 

\begin{figure}[!h]
\small
\begin{center}
\includegraphics[width=0.8\linewidth]{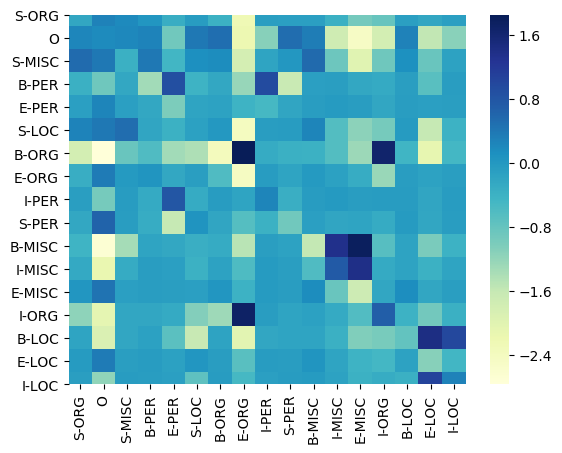}
\\(a) Skip-chain energy matrix $W_1$.
\includegraphics[width=0.8\linewidth]{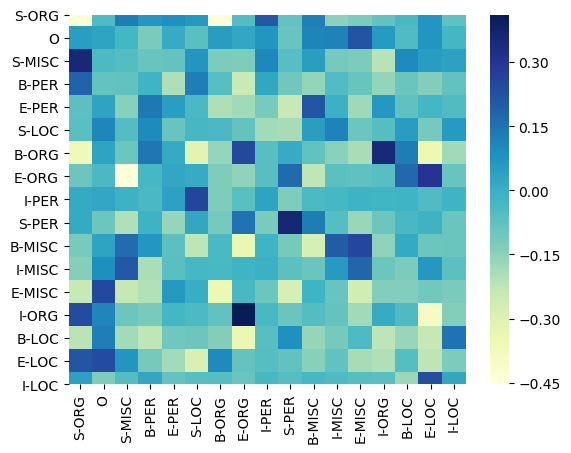}
\\(b) Skip-chain energy matrix $W_3$.
\end{center}
\caption{Learned pairwise potential matrices $W_1$ and $W_3$ for NER with skip-chain energy. The rows correspond to earlier labels and the columns correspond to subsequent labels.
}
\label{fig:skip_chain}
\end{figure}

\begin{figure}[!h]
\small
\begin{center}
\includegraphics[width=0.8\linewidth]{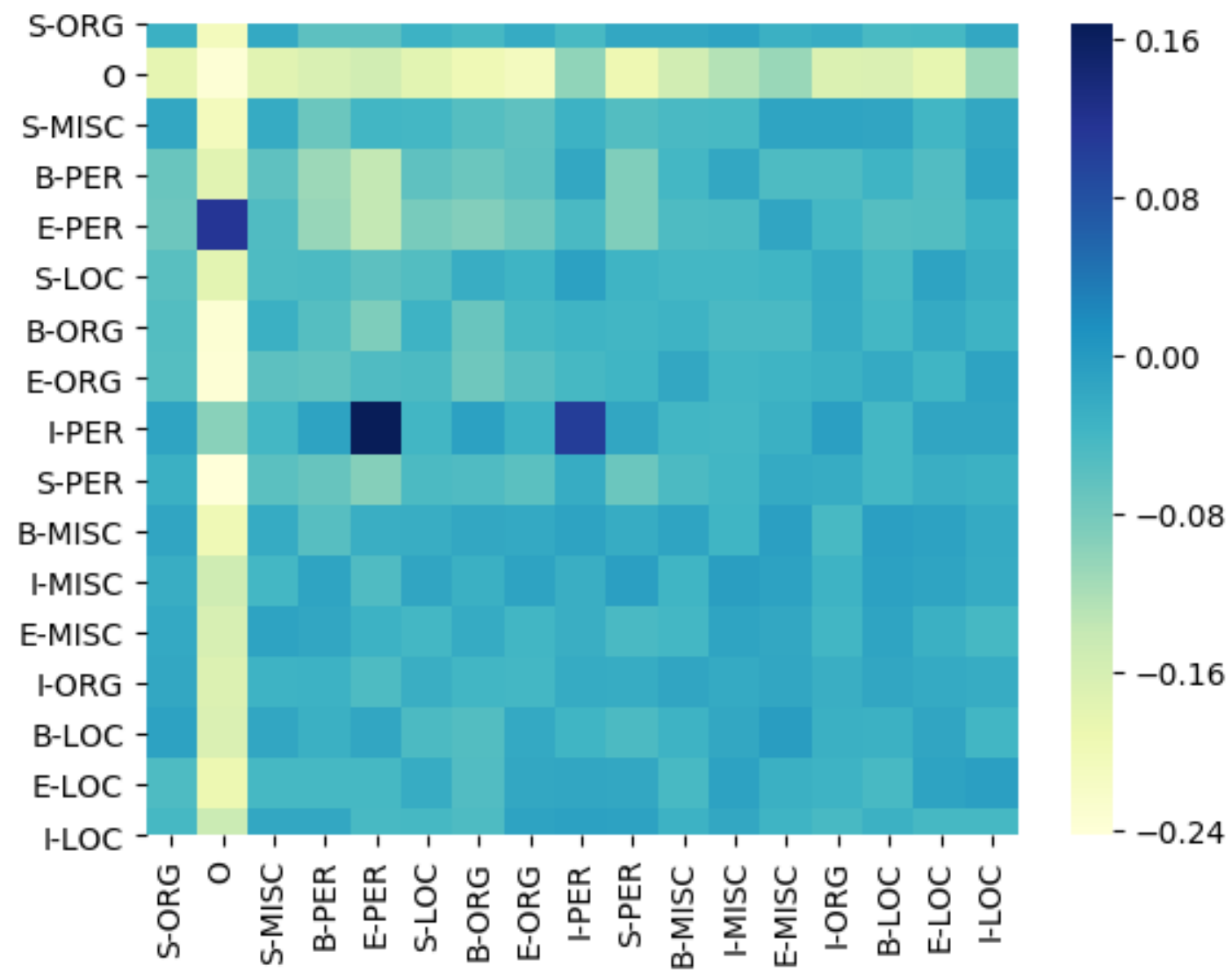}
\end{center}
\caption{Learned 2nd-order VKP energy matrix beginning with B-PER in NER dataset.}
\label{fig:kp}
\end{figure}

Figure~\ref{fig:skip_chain} visualizes two matrices in the skip-chain energy with $M=3$. We can see strong associations among labels in neighborhoods from $W_1$. For example, B-ORG and I-ORG are more likely to be followed by E-ORG. The $W_3$ matrix shows a strong association between I-ORG and E-ORG, which implies that the length of organization names is often long in the dataset. 

\begin{table}[!h]
\centering
\begin{tabular}{llll}
filter 26 & B-MISC & I-MISC & E-MISC \\
filter 12 & B-LOC & I-LOC & E-LOC \\
filter 15 & B-PER & I-PER & I-PER \\
filter 5 & B-MISC &  E-MISC & O \\
filter 6 & O & B-LOC & I-LOC \\
filter 16 & S-LOC & B-ORG & I-ORG \\
filter 44 & B-PER & I-PER & I-PER \\
filter 3 & B-MISC &  I-MISC & E-MISC \\
filter 2 & I-LOC &  E-LOC & O \\
filter 45 & O & B-LOC & E-LOC \\
\end{tabular}
\caption{Top 10 CNN filters with high inner product with 3 consecutive labels for NER. 
}\label{ta:cnnVis}
\end{table}

For the VKP energy with $M$=3, Figure~\ref{fig:kp} shows the learned matrix when the first label is B-PER, showing that B-PER is likely to be  followed by ``I-PER E-PER'', ``E-PER O'', or ``I-PER I-PER''.

\begin{figure}[!h]
\begin{center}
\includegraphics[width=\linewidth]{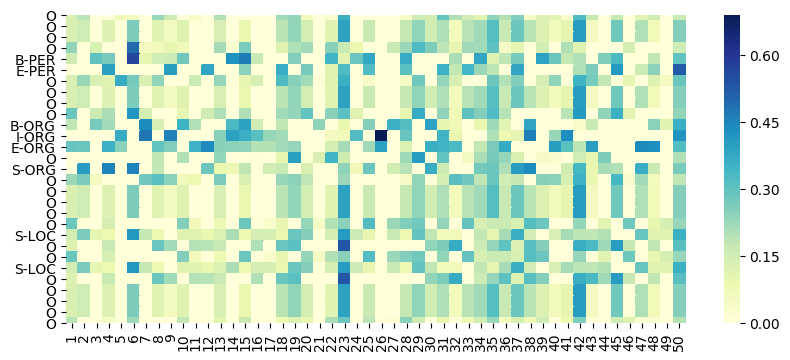}
\end{center}
\caption{Visualization of the scores of 50 CNN filters on a sampled label sequence. We can observe that filters learn the sparse set of label trigrams with strong local dependency. 
}
\label{fig:cnn}
\end{figure}

In order to visualize the learned CNN filters, we calculate the inner product between the filter weights and consecutive labels. For each filter, we select the sequence of consecutive labels with the highest inner product. Table~\ref{ta:cnnVis} shows the 10 filters with the highest inner product and the corresponding label trigram. 
All filters give high scores for structured label sequences with a strong local dependency, such as ``B-MISC {} I-MISC {} E-MISC" and ``B-LOC {}  I-LOC {} E-LOC", etc. 
Figure~\ref{fig:cnn} shows these inner product scores of 50 CNN filters on a sampled NER label sequence. We can observe that filters learn the sparse set of label trigrams with strong local dependency.

\section{Conclusion}
We explore arbitrary-order models with different neural parameterizations on sequence labeling tasks via energy-based inference networks. This approach achieve substantial improvement using high-order energy terms, especially in noisy data conditions, while having same decoding speed as simple local classifiers. 
\newpage

\mychapter{9}{Conclusion and Future Work}
We conclude this thesis by summarizing our key contribution and discussing some directions of future research.

\section{Summary of Contributions}

In this thesis, we made the following contributions:
\begin{itemize}
    \item We summarize the history of energy-based models and several commonly used learning and inference methods. Especially, what are the main benefit and difficulties of energy-based models (Chapter 1 and Chapter 2)? What is the connection of previous models (Chapter 2)? We also show several wildly used energy-based models in structured application in NLP (Chapter 2). This can be useful material for people who are interested in energy-based models. 

    \item For the structure tasks, the inference problem is very challenging due to the exponential large label space. Previously, the Viterbi algorithm and gradient descent were used for inference if considering structured components of complex NLP tasks. We develop a new decoding method called ``\textbf{energy-based inference network}'' which outputs structured continuous values. In our method, the time complexity for the inference is linear with the label set size. In Chapter 3, we shows ``energy-based inference network'' achieves a better speed/accuracy/search error trade off than gradient descent, while also being faster than exact inference at similar accuracy levels.

    \item  We have worked on several NLP tasks, including multi-label classification, part-of-speech tagging, named entity recognition, semantic role labeling, and non-autoregressive machine translation. We train a non-autoregressive machine translation model to minimize the energy defined by a pretrained autoregressive model, which achieves state-of-the-art non-autoregressive results on the IWSLT 2014 DE-EN and WMT 2016 RO-EN datasets, approaching the performance of autoregressive models. This indicates that the methods can be very possibly applied to a larger set of applications, especially more text-based generation tasks. 
    
    \item We also design a margin-based method for training energy-based models such as linear-chain CRF or high-order CRF. According to the visualization of the energy and performance improvements, we demonstrate We empirically demonstrate that this approach achieves substantial improvement using a variety of high-order energy terms on four sequence labeling tasks while having the same decoding speed as simple, local classifiers. We also find high-order energies to help in noisy data conditions.
\end{itemize}

\section{Future Work}

In this section, we propose several future directions.
\subsection{Exploring Energy Terms}
We use the linear-chain CRF energy, Tag Language model and high-order energy terms for sequence labeling task. It is worth to explore some other energy terms to capture complex label dependency. These terms can be used for sequence labeling or text generation tasks.

\paragraph{Language Coherence Terms} The way to improve the language coherence, we could use an additional energy term, the log-likelihood of y under the pretrained language models. The standard LSTM language model or the masked language model(e.g., BERT~\citep{devlin-etal-2019-bert}, RoBERTa~\citep{roberta}).
The pretrained language models are the vital resources to exploit large monolingual corpora for NMT in our framework.

Another approach for the repetition is modeling coverage of the source sentence~\citep{tu-etal-2016-modeling,mi-etal-2016-coverage}. And \citet{holtzman-etal-2018-learning} designed an energy term specifically targeting the prevention of repetition in the output.

\paragraph{Relating Attention to Alignment}
Since the learned attention function may diverge from alignment patterns between languages, several researchers have experimented with adding inductive biases to the attention function~\citep{cohn-etal-2016-incorporating,feng-etal-2016-improving}. This is often motivated by known characteristics about the alignment between the source and target language, particularly those related to monotonicity, distortion, and fertility. It is worth to try similar terms with ~\citep{cohn-etal-2016-incorporating,feng-etal-2016-improving}.

\paragraph{Local Cross Entropy Term} The standard log-likelihood scoring function that is used by nearly all NMT systems. However, it is still not explored how to incorporate the standard cross entropy term with the other proposed energy terms. According to sequence labeling experiments results in Chapter 5, chapter 6, and chapter 7, the local cross entropy loss could contribute the performance of the inference networks. The weight for the term can be carefully tuned for higher performance. In Chapter 3, we use the weight annealing scheme. 

However, the local cross entropy term has some limitations: it does not assign partial credit to the hypotheses if the word order of hypotheses is different from the reference and it could penalize semantically correct hypotheses if they differ lexically from the reference. 

\paragraph{BLEU}
Recently, there are several work that directly optimize the evaluation metrics such as BLEU to improve the translation systems. The only issue is that we need to consider how to do backpropagation through the non-differentiable term. With the similar approximate BLEU from ~\citet{tromble-etal-2008-lattice}, we could directly optimized the BLEU score for translation task or other generation tasks.

\paragraph{Beyond BLEU} ~\citet{wieting-etal-2019-beyond} proposes a new metric based on semantic similarity in order to get partial credit and reduces the penalties on semantically correct hypotheses. This term could potentially lead the inference networks search better hypotheses with semantically similar hypotheses. The embedding model to evaluate similarity allows the range of possible scores to be continuous. The inference networks could get the gradient directly from the term.
\begin{align}
    SIM(r, h) = \mathrm{cos}(g(r), g(h))
\end{align}
\noindent where $r$ is the reference and $h$ are the generate hypothesis. $g$ is the encoder for a token sequence. 
Furthermore, one variation of the metric could be based on the semantic similarity between the source sentence and the hypotheses. This term potentially fine-tuning the hypotheses that have different semantic meaning from source sentence.

\subsection{Learning Methods for Energy-based Models}

In our work , we use margin-based training metric for the energy function training. The objective for the energy function is:
\begin{align}
\Theta \gets & \argmin_{\Theta}\big[\cost(\canet(\x), \y_i) \!-\! E_{\Theta}(\x_i,\canet(\x)) + E_{\Theta}(\x_i, \y_i)\big]_{+} \nonumber
\end{align}

or training two inference networks $\canet$ and $\infnet$ jointly,

\begin{align}
\small
\begin{split}
\Theta \gets \argmin_{\Theta} \big[\cost(\canet(\x), \y_i) \!-\! E_{\Theta}(\x_i,\canet(\x)) + E_{\Theta}(\x_i, \y_i)\big]_{+}  + \lambda \big[- E_{\Theta}(\x_i,\infnet(\x_i)) + E_{\Theta}(\x_i, \y_i)\big]_{+} \nonumber
\end{split}
\end{align}

The other interesting approach for energy training is noise-contrastive estimation~\citep{pmlr-v9-gutmann10a,ouSLT2018,ouICASSP2018,bakhtin2020energybased}  (NCE). 
NCE is proposed for learning unnormalized statistical models. It use logistic regression to discriminate between the data samples drawn from the data distribution and noise samples drawn from a noise distribution. They assume that the learned models are “self-normalized”.

It would be interesting to see some analysis on two different approaches. Or as we know, NCE need a predefined well-formed noise distribution. So it is hard to  inject ``domain
knowledge`` of text understanding. We can add ``negative examples'' even the noise distribution form is unknown. In addition, inference networks can model more complex noise distribution so that a better energy model can be learned.

\bibliographystyle{abbrvnat}
\bibliography{anthology,refs}

\end{document}